\documentclass[10pt,twocolumn,letterpaper]{article}

\usepackage[pagenumbers]{cvpr}

\newif\ifarxiv
\arxivtrue

\usepackage{latex/style/mymath}
\usepackage{latex/style/eccvabbrv}
\usepackage{derivative}
\usepackage{tikz}
\usepackage{tikz-3dplot}
\usetikzlibrary{decorations.pathreplacing}

\usepackage{amsmath}
\usepackage{aligned-overset}
\usepackage{cancel}
\usepackage{multicol}
\usepackage{upgreek}
\usepackage{xcolor}
\usepackage{pgfplots}\pgfplotsset{compat=1.18}

\usepackage{mathrsfs}
\usepackage{upgreek}
\DeclareSymbolFont{euler}{U}{eur}{m}{n}
\DeclareMathSymbol{\calOmega}{\mathalpha}{euler}{"0A}

\newcommand{\timingGpu}{B200}
\newcommand{\timingHeight}{2\,848}
\newcommand{\timingIterations}{25}

\newcommand{\timingMeanS}{703.5}

\newcommand{\timingObservations}{2\,162\,688}
\newcommand{\timingParametersCamera}{1\,443}
\newcommand{\timingParametersLighting}{9}
\newcommand{\timingParametersPoses}{792}
\newcommand{\timingParametersTarget}{29}
\newcommand{\timingParametersTotal}{2\,273}
\newcommand{\timingPerIterationS}{26.1}
\newcommand{\timingPhotographs}{132}
\newcommand{\timingRuns}{8}
\newcommand{\timingSamples}{16\,384}
\newcommand{\timingStdS}{3.7}
\newcommand{\timingThreads}{8}

\newcommand{\timingWidth}{4\,256}

\definecolor{cvprblue}{rgb}{0.21,0.49,0.74}
\usepackage[pagebackref,breaklinks,colorlinks,allcolors=cvprblue]{hyperref}

\title{Real Cameras are Radiometric}

\author{Ties Harder \quad Felix Seegr\"aber \quad David Nakath \quad Hendrik Sauer \quad Kevin K\"oser \quad James Pritts\\
Kiel University, Germany\\
}

\begin{document}
\maketitle
\begin{abstract}
	We propose the radiometric camera, a model that lifts any central
	geometric camera model to a radiometric one.  The lifted model
	keeps the projection of the geometric model and adds a finite
	aperture, a free-form focus surface and a generic vignetting
	function.  We calibrate it by inverse rendering directly on the
	relative irradiance of the sensor, from the same images of
	standard calibration targets used in geometric calibration.
	Geometric calibration absorbs the bias of its feature detections.
	The proposed radiometric calibration uses no features, and our
	radiometric cameras have lower reprojection errors on held-out
	test sets.  The focus surface extends defocus modeling from the
	thin lens to any central camera, and we use it in a first
	multi-view radiometric reconstruction, which recovers the in-focus
	reflectance of a texture from defocused images.  The
	vignetting is estimated jointly and accurately reproduces a flat
	field measured through a diffuser.  Because every standard camera
	model lifts to a radiometric one, the radiometric camera naturally
	and directly extends generative scene models beyond multi-view
	geometry to multi-view radiometric ones, where phenomena such as
	lighting and attenuation can be accurately estimated.\footnote{\url{https://realcamerasareradiometric.github.io}}
\end{abstract}

\section{Introduction}
\label{sec:introduction}
Reconstruction pipelines such as COLMAP \cite{schoenberger2016sfm}
describe the camera with a few central projection models:
Brown--Conrady \cite{Brown1966DecenteringDO} for perspective lenses,
Kannala--Brandt \cite{kannala-brandt} and Double Sphere
\cite{double-sphere} for fisheyes, and generic spline models
\cite{10000params,spline-camera} where a parametric model falls short.
These are the cameras of multiple-view geometry. They model where a
scene point is imaged, but not how much light arrives there. All of
them assume a pinhole aperture, \ie that a single ray connects a scene
point with the sensor. With a pinhole aperture every scene point is in
focus and no light is lost toward the image border, so these models
cannot represent defocus or vignetting.

\suppressfloats[t]% keeps the figure out of the left column, so it goes to the top of the right one
\begin{figure}[t]
        \centering
        \providecommand{\rcfigdir}{fig}%
\begingroup
\ifdefined\rcpanelwd\else\newlength\rcpanelwd\fi\ifdefined\rcpanelht\else\newlength\rcpanelht\fi\ifdefined\rcteaserwd\else\newlength\rcteaserwd\fi
\setlength{\rcteaserwd}{\linewidth}\setlength{\rcpanelwd}{\dimexpr(\linewidth-4pt)/2\relax}\setlength{\rcpanelht}{0.54348\rcpanelwd}%
\footnotesize\setlength{\tabcolsep}{0pt}\renewcommand{\arraystretch}{1}%
\begin{tabular}{@{}l@{}}
  {\def\rcpanelwidth{\rcpanelwd}%
   \providecommand{\rcfigdir}{fig}%
\providecommand{\rcpanelwidth}{\linewidth}%
\begin{tikzpicture}[baseline=(current axis.center)]
\begin{axis}[
  scale only axis, width=\rcpanelwidth, height=0.54348*\rcpanelwidth,
  grid=none, log ticks with fixed point, tick label style={font=\scriptsize}, label style={font=\small}, legend cell align=left, legend style={font=\scriptsize, draw=none, fill=none},
  xmin=0, xmax=460, ymin=0, ymax=250, enlargelimits=false,
  xtick=\empty, ytick=\empty, axis on top, axis line style={draw=none}]
\addplot graphics[xmin=0, xmax=460, ymin=0, ymax=250] {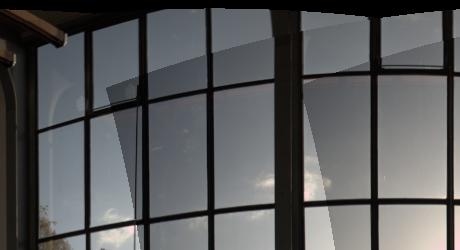};
\end{axis}
\end{tikzpicture}%
\makebox[0pt][r]{\makebox[\rcpanelwd][c]{\raisebox{\dimexpr0.5\rcpanelht-1.5pt\relax}[0pt][0pt]{\tikz[baseline=(k.north)]\node[fill=white,inner sep=1.2pt,font=\scriptsize](k){no vignetting correction};}}}\hspace{4pt}\providecommand{\rcfigdir}{fig}%
\providecommand{\rcpanelwidth}{\linewidth}%
\begin{tikzpicture}[baseline=(current axis.center)]
\begin{axis}[
  scale only axis, width=\rcpanelwidth, height=0.54348*\rcpanelwidth,
  grid=none, log ticks with fixed point, tick label style={font=\scriptsize}, label style={font=\small}, legend cell align=left, legend style={font=\scriptsize, draw=none, fill=none},
  xmin=0, xmax=460, ymin=0, ymax=250, enlargelimits=false,
  xtick=\empty, ytick=\empty, axis on top, axis line style={draw=none}]
\addplot graphics[xmin=0, xmax=460, ymin=0, ymax=250] {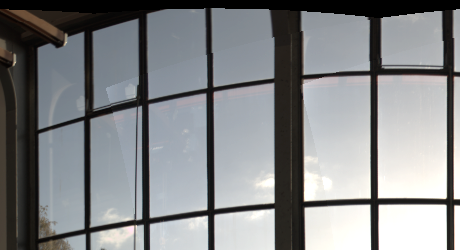};
\end{axis}
\end{tikzpicture}%
\makebox[0pt][r]{\makebox[\rcpanelwd][c]{\raisebox{\dimexpr0.5\rcpanelht-1.5pt\relax}[0pt][0pt]{\tikz[baseline=(k.north)]\node[fill=white,inner sep=1.2pt,font=\scriptsize](k){our vignetting correction};}}}} \\\noalign{\vskip 7pt}
  {\def\rcpanelwidth{\rcteaserwd}\providecommand{\rcfigdir}{fig}%
\providecommand{\rcpanelwidth}{\linewidth}%
\begin{tikzpicture}[baseline=(current axis.center)]
\begin{axis}[
  scale only axis, width=\rcpanelwidth, height=0.13078*\rcpanelwidth,
  grid=none, log ticks with fixed point, tick label style={font=\scriptsize}, label style={font=\small}, legend cell align=left, legend style={font=\scriptsize, draw=none, fill=none},
  xmin=0, xmax=627, ymin=0, ymax=82, enlargelimits=false,
  xtick=\empty, ytick=\empty, axis on top, axis line style={draw=none}]
\addplot graphics[xmin=0, xmax=627, ymin=0, ymax=82] {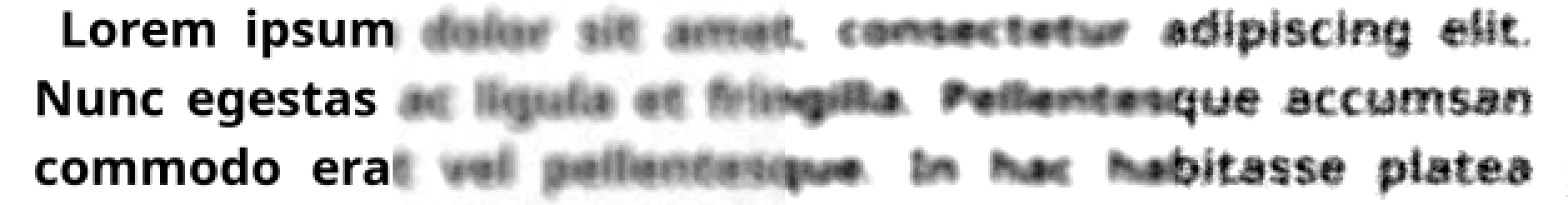};
\end{axis}
\end{tikzpicture}%
} \\\noalign{\vskip 2.5pt}
  \makebox[157\dimexpr\rcteaserwd/627\relax][c]{\scriptsize ground truth}\makebox[157\dimexpr\rcteaserwd/627\relax][c]{\scriptsize training photograph}\makebox[156\dimexpr\rcteaserwd/627\relax][c]{\scriptsize pinhole}\makebox[157\dimexpr\rcteaserwd/627\relax][c]{\scriptsize ours} \\
\end{tabular}%
\endgroup

        \caption{\protect\input{fig/teaser-caption}}
        \label{fig:teaser}
\end{figure}

In this paper, we make two major contributions, the \textit{radiometric camera} and its \textit{calibration}. The \textit{first} is a model that lifts any central geometric camera model to a radiometric
one. The lifted model keeps
the projection of the geometric model and replaces the pinhole
aperture by a finite aperture with a free-form focus surface and a
generic vignetting function. Setting the aperture to a pinhole
recovers the geometric model. To our knowledge, it is the first model
that generalizes the central camera models in common use with the major
radiometric components of a lens, its aperture, focus and vignetting,
\textit{and} that is calibrated in all of them jointly from images of a
standard calibration target (\cref{tab:related}).

The radiometric camera unlocks more complex multi-view radiometric
problems, in which several views jointly constrain radiometric
unknowns such as the radiant intensity distribution and the pose of a
light source
\cite{weber2001practical,park2014nonisotropic,visentini2015simultaneous,santo2018pinmotion}
or the attenuation of water
\cite{bryson2016truecolor,skinner2017automatic,akkaynak2019seathru,nakath2021insitu}.
We aim to solve such problems generatively, \ie by a maximum-likelihood estimate of the scene,
which requires an image-formation model that predicts the measurements
of the sensor from the scene. The camera is the last stage of this
model. The irradiance at a sensor point is gathered over the cone of rays that the
aperture admits, and where the lens is out of focus, these rays come
from different scene points. Earlier radiometric extensions of
geometric cameras relate each sensor value to the radiance of a single
scene point, or add corrections to the image
\cite{Kim-PAMI-2008,rehder2017direct,Hannemose_2019}. The difficulty in extending a geometric camera to
radiometry is that the quantities to be added depend on exactly the components that the
pinhole aperture removed. The \textit{radiometric} camera
restores them, so that the irradiance becomes the integral over the whole
aperture. We address its \textit{calibration} by inverse rendering on the RAW photosites, whose values are
proportional to the irradiance that they receive, from the same
images of standard calibration targets as a geometric
calibration. The constraints are not the incidences of detected
features with their projections, but the relative radiant flux of the
photosites, which the same target under the same lighting must explain
in every view.

However, differentiating a camera
with a finite aperture gives rise to boundary terms, from the
aperture, the discontinuous edges of the calibration pattern and from
visibility, which bias the gradients of standard differentiable
renderers \cite{Zhang:2020:PSDR}. Moreover, the lighting, the reflectance and the shape of
the target have to be estimated simultaneously without being confused
with the camera's defocus and vignetting. We give a single formulation
that resolves all of these difficulties. Neither a parameter-dependent
discontinuity nor a parameter-dependent integration domain remains in
it, the focus surface is parameterized so that its gradient does not
vanish, the lighting varies over the target so that it separates from
the vignetting, the gauge freedoms between the camera,
the lighting and the target are fixed exactly, the shape of the
target is a bending of the board with few parameters, and the linear
structure of the model in the lighting and the vignetting gives a
closed-form start. We use the calibration to validate the components
of the model in three respects. \textit{First}, they benefit the geometric
accuracy. We show that the feature detections of widely used
calibration tools are biased, and that the radiometric calibration,
which uses none, lowers the reprojection error on a held-out
high-precision target. \textit{Second}, their estimated parameters represent
the physical phenomena. The focus surface is learned as a plane for a wide-angle lens and as
a hemisphere for a fisheye, and the vignetting function reproduces a
flat field measured through a diffuser. \textit{Third}, the calibrated camera
solves a first multi-view radiometric problem: it recovers the
unknown in-focus reflectance of a texture from defocused images.

\begin{table}[t]
  \centering
  \begingroup
  \footnotesize\setlength{\tabcolsep}{2pt}%
  \newcommand{\y}{\checkmark}\newcommand{\p}{(\checkmark)}\newcommand{\n}{--}%
  \begin{tabular*}{\linewidth}{@{}l@{\extracolsep{\fill}}ccccc@{}}
    \toprule
     & \textbf{Proj.} & \textbf{Apert.} & \textbf{Focus} & \textbf{Vign.} & \textbf{Calib.} \\
    \midrule
    Thin lens \cite{pharr2023physically}, Mitsuba \cite{Mitsuba3} & \p & \y & \p & \p & \n \\
    Lens-design ray tracing \cite{kolb1995realistic} & \p & \y & \p & \p & \n \\
    Tang and Kutulakos \cite{tang2013aberrated} & \p & \y & \p & \p & \n \\
    \addlinespace
    Kang and Weiss \cite{kang2000calibrate} & \p & \n & \n & \p & \p \\
    Kim and Pollefeys \cite{Kim-PAMI-2008} & \n & \n & \n & \y & \p \\
    Neural lens model \cite{xian2023neurallens} & \y & \n & \n & \y & \p \\
    Rehder \etal \cite{rehder2017direct} & \p & \n & \n & \n & \p \\
    Hannemose \etal \cite{Hannemose_2019} & \p & \n & \n & \n & \p \\
    Lens blur fields \cite{lin2023blurfields} & \n & \p & \n & \n & \p \\
    \addlinespace
    \textbf{Radiometric camera (ours)} & \y & \y & \y & \y & \y \\
    \bottomrule
  \end{tabular*}
  \endgroup
  \caption{Prior camera models (top) and calibration methods (middle) against the radiometric camera: any central projection, a finite aperture, a free-form focus surface, a generic vignetting function, and the joint calibration of all of them from images. \checkmark: general; (\checkmark): restricted, \eg to the perspective projection, a plane of focus, the $\cos^4$ falloff or a subset of the parameters; --: not modeled. \Cref{sec:related-work} gives the details.}
  \label{tab:related}
\end{table}
\section{Related Work}
\label{sec:related-work}

\paragraph{Lens models.} The thin lens
\cite{Hecht+2023,pharr2023physically} gives the perspective camera a
finite aperture, so that a scene point off the plane of focus is
imaged to a circle of confusion. It is tied to the perspective projection and has
no natural counterpart in classic distortion models \cite{kannala-brandt, double-sphere, Brown1966DecenteringDO, Conrady1919DecentredL}. The plane of focus and the vignetting that emerge from this model are too simple to capture the actual phenomena. The
surface of best focus of a real lens is curved, which is known in
optics as field curvature \cite{Kingslake2010}. Tang and Kutulakos
\cite{tang2013aberrated} derive the blur of a rotationally symmetric
lens as an integral over its aperture and show that field curvature
acts as a defocus that depends on the position in the image. This is
what a focus surface describes, and it justifies our design, which
replaces their fixed, symmetric dependence by a free-form one. Their
analysis is theoretical, and the aberrations for which the rays of a
sensor position no longer meet in a point are not part of our model. Models that trace the
rays through the lens elements \cite{kolb1995realistic} need the
optical design of the lens, which manufacturers rarely publish. The
neural lens model of Xian \etal \cite{xian2023neurallens} replaces the
distortion of a pinhole camera by an invertible network and models the
vignetting as a gain per pixel of the image, without an aperture. The thin lens is also the only finite aperture of the
differentiable renderer Mitsuba
\cite{NimierDavidVicini2019Mitsuba2,Mitsuba3}. Its cameras are
perspective and without lens distortion, and the aperture radius and
the focus distance of its thin lens are not differentiable parameters,
so neither the projection of a general central model nor its aperture
and focus can be calibrated with it. Our calibration keeps every
camera parameter differentiable: neither a parameter-dependent
discontinuity nor a parameter-dependent integration domain remains in
its formulation.

\paragraph{Radiometric calibration and image corrections.} Radiometric calibration methods
estimate the vignetting, often together with the response function,
from the pixel values of overlapping images
\cite{Kim-PAMI-2008,single-image-vignetting,direct-sparse-odometry}.
In \cite{Kim-PAMI-2008,xian2023neurallens} the vignetting is an empirical function of the image position, and
each pixel value is related to the radiance of a single scene point.
Reconstruction methods likewise add corrections around a geometric
camera, such as per-image gains \cite{brown2007automatic}, appearance
embeddings \cite{martinbrualla2021nerfw} or blur kernels
\cite{ma2022deblurnerf}. These compensate for the unmodeled radiometry in the image, where
defocus, which depends on the depth of the scene point, cannot be
corrected consistently. The radiometric camera models it inside the
camera, as an integral over its aperture.

\paragraph{Calibration from pixel values.} Kang and Weiss
\cite{kang2000calibrate} obtain a rough focal length of a perspective
camera from the $\cos^4$ falloff in a single image of a uniform
surface. Rehder \etal \cite{rehder2017direct} formulate a
maximum-likelihood calibration on the image intensities by rendering
the target. They assume an evenly lit target and no vignetting, and
take a single radiance sample per pixel, to which they attribute that
their static calibration is no more precise than one from features;
defocus, vignetting and uneven lighting are named as future work.
Hannemose \etal \cite{Hannemose_2019} also render the target densely,
with a pinhole camera with Brown--Conrady distortion, but their
calibration is photometric and not radiometric: the rendering blurs
the texture with a Gaussian whose width is a free parameter of each
interest point, and it models neither the lighting nor the intensity
scale of the image. A separate line calibrates the
blur of a lens as a learned point spread function, apart from the
projection of the camera \cite{lin2023blurfields}. In the radiometric
camera, the defocus follows from the aperture and the focus surface of
the same model that projects the scene, for any central projection,
and it is calibrated together with the vignetting, the lighting and
the target.

\relax

\section{Radiometric Camera Model}
\label{sec:radiometric-camera-model}
\label{sec:image-formation}
Let $\mathcal{I}\subseteq\mathbb{R}^2$ be the sensor plane. A pixel
$\vbu$ with the area $\mathcal{I}_{\vbu}\subseteq\mathcal{I}$ measures
the radiant flux~\cite{pharr2023physically}
\begin{equation}
\label{eq:per-channel-rad-flux}
    \Phi(\vbu) \coloneqq \int_{\mathcal{I}_{\vbu}}\int_{\calOmega(\vbp)} L(\vbx(\vbp,\vbomega), \vbomega_o(\vbp,\vbomega)) \cos\theta(\vbomega)\odif{\vbomega}\odif{\vbp},
\end{equation}
where $\calOmega(\vbp)$ is the solid angle under which the sensor
point $\vbp$ sees the aperture, $\vbx(\vbp,\vbomega)$ the scene point
visible from $\vbp$ along the direction $\vbomega$,
$L(\vbx,\vbomega_o)$ the radiance that leaves $\vbx$ in the direction
$\vbomega_o$ towards the lens, and $\theta$ the angle of incidence on
the sensor. The inner integral is the irradiance $E(\vbp)$. All
quantities are per color channel, and we omit the channel index
(\cref{sec:supp-image-formation} gives the spectral formulation). The
radiance $L$ follows from the lighting and the reflectance of the
scene (\cref{eq:scene-rad}). The lens model determines
$\calOmega(\vbp)$, $\vbx(\vbp,\vbomega)$ and $\vbomega_o(\vbp,\vbomega)$. With a pinhole aperture,
$\calOmega(\vbp)$ is a single direction, so every scene point is in
focus and there is no vignetting (\cref{sec:supp-pinhole-radiometry}).

\subsection{Projection}
Let $\vbpi^{-1}: \mathcal{I}\rightarrow\mathcal{S}^2$ be the
back-projection of a central geometric camera model with the center
of projection $\vbo$. It maps a sensor point $\vbp$ to the unit
direction $\hat{\vbd}(\vbp)\coloneqq\vbpi^{-1}(\vbp)$ of its ray. For
a model that is given by its forward projection, we invert the
projection numerically per sample. The radiometric camera keeps
$\vbpi^{-1}$ as it is.

\subsection{Aperture and Focus Surface}
\label{sec:defocus}
\begin{figure}[t]
        \centering
        \usetikzlibrary{intersections}%
\resizebox{\linewidth}{!}{%
\begin{tikzpicture}[x=1cm, y=1cm, >=stealth, font=\footnotesize,
    ray/.style={line width=0.45pt, blue!70!black},
    chief/.style={line width=0.75pt, black},
    inside/.style={line width=0.3pt, densely dashed, black!55},
    axis/.style={line width=0.3pt, densely dotted, black!60},
    dim/.style={line width=0.3pt, black!60}]
  \newcommand{\rccamera}{%
    \draw[axis] (-0.25,0) -- (1.75,0);
    \draw[line width=1pt] (0,-1.0) -- (0,1.0);
    \node[anchor=south] at (0,1.0) {$\mathcal{I}$};
    \draw[line width=1.8pt] (1.0,0.45) -- (1.0,1.0) (1.0,-0.45) -- (1.0,-1.0);
    \node[anchor=south] at (1.0,1.0) {$\mathcal{A}$};
    \coordinate (p) at (0,0.6); \coordinate (o) at (1.0,0);
    \coordinate (a1) at (1.0,0.45); \coordinate (a2) at (1.0,-0.45);
    \draw[inside] (p) -- (a1) (p) -- (a2) (p) -- (o);
    \fill (p) circle (1.2pt) node[anchor=east, inner sep=2pt] {$\vbp$};
    \fill (o) circle (1.2pt);
    \fill[blue!70!black] (a2) circle (1.2pt);}
  \begin{scope}
    \rccamera
    \node[anchor=north east, inner sep=1.5pt] at (o) {$\vbo$};
    \node[anchor=north east, inner sep=1.5pt] at (a2) {$\vba$};
    \coordinate (xf) at (2.75,-1.05);
    \draw[line width=0.9pt, red!75!black] plot[smooth, tension=0.7] coordinates {(2.62,1.25) (2.78,0.1) (2.75,-1.05) (2.60,-1.85)};
    \node[red!75!black, anchor=south] at (2.62,1.25) {$\mathcal{F}$};
    \path[name path=sceneA] plot[smooth, tension=0.8] coordinates {(3.95,1.25) (3.75,0.4) (3.92,-0.5) (3.74,-1.3) (3.9,-2.05)};
    \path[overlay, name path=rA0] (o) -- ($(o)!2.2!(xf)$);
    \path[overlay, name path=rA1] (a1) -- ($(a1)!2.2!(xf)$);
    \path[overlay, name path=rA2] (a2) -- ($(a2)!2.2!(xf)$);
    \path[name intersections={of=sceneA and rA0, by=s0}];
    \path[name intersections={of=sceneA and rA1, by=s1}];
    \path[name intersections={of=sceneA and rA2, by=s2}];
    \draw[chief] (o) -- (s0);
    \draw[chief, ->] (o) -- ($(o)!0.42!(xf)$);
    \draw[dim] (1.92,0.12) -- ($(o)!0.30!(xf)$);
    \node[anchor=south west, inner sep=1pt] at (1.80,0.10) {$\hat{\vbd}(\vbp)$};
    \draw[ray] (a1) -- (s1) (a2) -- (s2);
    \draw[ray, ->] (a2) -- ($(a2)!0.5!(xf)$);
    \node[blue!70!black, anchor=north, inner sep=2pt] at (1.75,-0.82) {$\vbpi^{-1}(\vbp,\vba)$};
    \draw[line width=1pt] plot[smooth, tension=0.8] coordinates {(3.95,1.25) (3.75,0.4) (3.92,-0.5) (3.74,-1.3) (3.9,-2.05)};
    \node[anchor=south] at (3.9,1.25) {scene};
    \draw[line width=2.2pt, blue!70!black, line cap=round] (s1) -- (s0) -- (s2);
    \draw[dim] (3.30,-2.02) -- ($(s0)+(-0.07,-0.05)$);
    \node[blue!70!black, anchor=north, inner sep=1.5pt] at (3.05,-1.98) {region seen by $\vbp$};
    \fill[red!75!black] (xf) circle (1.7pt);
    \node[anchor=south west, inner sep=1pt] at ($(xf)+(0.10,0.16)$) {$\vbx_f(\vbp)$};
    \node[anchor=north] at (1.9,-2.45) {lifted perspective};
  \end{scope}
  \begin{scope}[xshift=4.75cm]
    \rccamera
    \draw[axis] (o) -- ++(1.15,-0.69);  % the continuation of the line from p: the direction of a perspective model
    \coordinate (xf) at ($(o)+(-62:1.45)$);
    \draw[line width=0.9pt, red!75!black] ($(o)+(62:1.45)$) arc[start angle=62, end angle=-92, radius=1.45];
    \node[red!75!black, anchor=south west, inner sep=1pt] at ($(o)+(58:1.45)$) {$\mathcal{F}$};
    \path[name path=sceneB] plot[smooth, tension=0.8] coordinates {($(o)+(42:2.25)$) ($(o)+(12:2.1)$) ($(o)+(-18:2.25)$) ($(o)+(-48:2.1)$) ($(o)+(-76:2.25)$) ($(o)+(-100:2.15)$)};
    \path[overlay, name path=rB0] (o) -- ($(o)!2.4!(xf)$);
    \path[overlay, name path=rB1] (a1) -- ($(a1)!2.4!(xf)$);
    \path[overlay, name path=rB2] (a2) -- ($(a2)!2.4!(xf)$);
    \path[name intersections={of=sceneB and rB0, by=t0}];
    \path[name intersections={of=sceneB and rB1, by=t1}];
    \path[name intersections={of=sceneB and rB2, by=t2}];
    \draw[chief] (o) -- (t0);
    \draw[chief, ->] (o) -- ($(o)!0.5!(xf)$);
    \draw[dim] (1.72,0.12) -- ($(o)!0.32!(xf)$);
    \node[anchor=south west, inner sep=1pt] at (1.60,0.10) {$\hat{\vbd}(\vbp)$};
    \draw[ray] (a1) -- (t1) (a2) -- (t2);
    \draw[line width=1pt] plot[smooth, tension=0.8] coordinates {($(o)+(42:2.25)$) ($(o)+(12:2.1)$) ($(o)+(-18:2.25)$) ($(o)+(-48:2.1)$) ($(o)+(-76:2.25)$) ($(o)+(-100:2.15)$)};
    \draw[line width=2.2pt, blue!70!black, line cap=round] (t1) -- (t0) -- (t2);
    \fill[red!75!black] (xf) circle (1.7pt);
    \node[anchor=west, inner sep=3pt] at (xf) {$\vbx_f(\vbp)$};
    \node[anchor=north] at (1.9,-2.45) {lifted fisheye};
    \node[anchor=south west, inner sep=1pt] at ($(o)+(42:2.25)$) {scene};
  \end{scope}
\end{tikzpicture}}%
        \caption{The radiometric camera lifts any central geometric model, here a perspective and a fisheye model, in a cross-section. The geometric model alone sets the direction $\hat{\vbd}(\vbp)$ in which the ray of the sensor point $\vbp$ leaves the center $\vbo$ of the aperture $\mathcal{A}$; the model does not trace the light inside the camera (dashed). The lift adds the aperture: the rays through all aperture points $\vba$ meet in the point of focus $\vbx_f(\vbp)$ on the focus surface $\mathcal{F}$, which is close to a plane for the perspective model and surrounds the camera for the fisheye. Where the scene is off $\mathcal{F}$, the cone of these rays meets it in a region (thick blue), from which $\vbp$ collects its light.}
        \label{fig:model}
\end{figure}
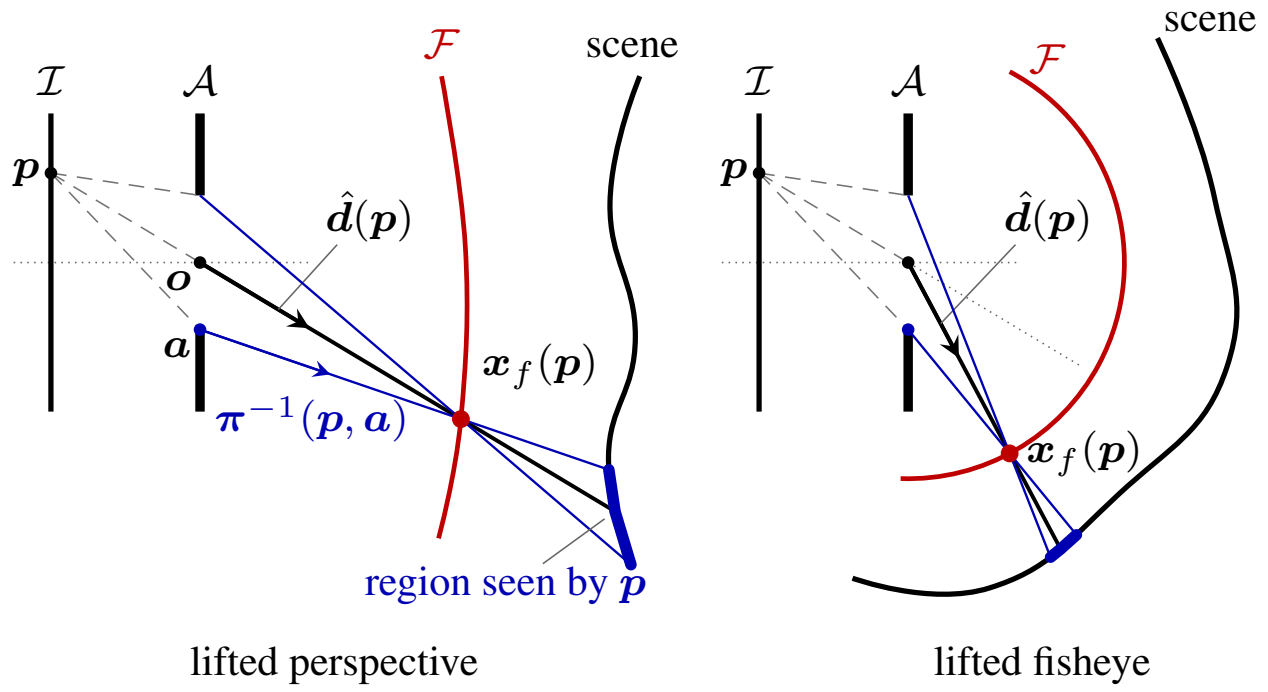
The thin lens~\cite{pharr2023physically} replaces the pinhole by an
aperture disc $\mathcal{A}$ of radius $r$, centered at $\vbo$ and
parallel to the sensor. All rays from a sensor point $\vbp$ through
$\mathcal{A}$ meet in one point $\vbx_f(\vbp)$ on the plane of focus.
Where the scene does not pass through $\vbx_f(\vbp)$, the cone of
these rays intersects it in a region and not in a point, and $\vbp$
collects the light of this whole region, which defocuses the image. For a real
lens, the points of focus form a surface $\mathcal{F}$
(\cref{fig:model}) that is not a plane in general \cite{Hecht+2023}: it is approximately a plane for a
thin-lens-like lens and a sphere for a fisheye
\cite{Kingslake2010,wide-field-of-view-2017}. We therefore learn the
focus surface. We represent it by the inverse distance $\tau(\vbp)$
from $\vbo$ to the point of focus $\vbx_f(\vbp)\in\mathcal{F}$ along
the ray of $\vbp$,
\begin{equation}
    \vbx_f(\vbp) \coloneqq \vbo + \frac{1}{\tau(\vbp)}\,\hat{\vbd}(\vbp),
\end{equation}
and extend the back-projection to the aperture
(\cref{fig:model}). The ray from $\vbp$
through an aperture point $\vba\in\mathcal{A}$ is directed from
$\vba$ towards $\vbx_f(\vbp)$:
\begin{equation}
\label{eq:lifted-back-projection}
    \vbpi^{-1}(\vbp,\vba)
    \coloneqq \frac{\vbx_f(\vbp) - \vba}{\lVert \vbx_f(\vbp) - \vba \rVert}
    = \frac{\tau(\vbp)(\vbo-\vba) + \hat{\vbd}(\vbp)}{\lVert\tau(\vbp)(\vbo-\vba) + \hat{\vbd}(\vbp)\rVert}.
\end{equation}
This is the lift of the geometric model. Setting the aperture $\mathcal{A}=\{\vbo\}$ reduces the lifted model back to the geometric one, since $\vbpi^{-1}(\vbp,\vbo)=\vbpi^{-1}(\vbp)$. Where the latter back-projects to a single ray,
$\vbpi^{-1}(\vbp,\mathcal{A})$ back-projects to a cone of rays
through $\vbx_f(\vbp)$. Similarly, with the perspective back-projection and a plane of focus, the model reduces to
the thin lens (\cref{sec:supp-thin-lens-equiv}). The inverse distance
avoids intersecting the rays with $\mathcal{F}$, and unlike the
distance itself, it keeps the gradient from vanishing for a far focus
\cite{unified-inverse-depth,direct-sparse-odometry}
(\cref{sec:supp-inverse-depth}). We model $\tau$ by a thin plate
spline \cite{thin-plate-spline} over a grid restricted to the illuminated region
of the sensor.

\subsection{Vignetting}
\label{sec:vignetting}
Written as an integral over the aperture, the flux of the thin lens
carries the factor $\cos^4\theta/s'^2$ that expresses its vignetting, with the image distance $s'$~\cite{Goossens_2019}
(\cref{sec:supp-thin-lens-vignetting}). Real lenses differ much
from it (\cref{fig:vignetting}). Hence, we replace this factor by a generic
vignetting function $V$. We also normalize by the aperture area $\pi r^2$,
so that the aperture radius changes the defocus but not the
brightness of the image, such that its scale is carried by the lighting (see \cref{sec:lighting}, \ref{sec:supp-gauge-freedoms}) alone. To spare us computing a boundary integral (see \cref{sec:boundary-terms}), we substitute the aperture point
$\vba=\vbo+r\vbxi$ with $\vbxi$ in the unit disc $\mathcal{D}$ of the
aperture plane. The area element $\odif{\vba}=r^2\odif{\vbxi}$
cancels the $r^2$ of the normalization, and the radiometric camera
measures
\begin{equation}
\label{eq:rad-flux-with-vignetting}
    \Phi(\vbu)
    = \frac{1}{\pi}\int_{\mathcal{I}_{\vbu}}V(\vbp)
    \int_{\mathcal{D}}L\bigl(\vbx(\vbp,\vba),-\vbpi^{-1}(\vbp,\vba)\bigr)\odif{\vbxi}\odif{\vbp},
\end{equation}
where $\vbx(\vbp,\vba)$ is the scene point that the ray from $\vba$ in
the direction $\vbpi^{-1}(\vbp,\vba)$ meets. The radius now
appears only in the position of the aperture point, and no longer in
the domain of the integral. We assume a vignetting
that is radially symmetric about the principal point
$\vbc\in\mathcal{I}$ and represent it by a cubic B-spline
\cite{piegl1997nurbs}
\begin{equation}
    V(\vbp) \coloneqq \sum_{i=0}^{n} N_{i,3}\Bigl(\frac{\lVert \vbp-\vbc \rVert}{r_{\text{max}} + \lVert \vbp-\vbc \rVert}\Bigr)P_{i},
\end{equation}
with the basis functions $N_{i,3}$, the control points
$\vbTheta_v = \{P_{i}\}_{i\in\{0,\dots,n\}}$, $n=15$, and the largest
distance $r_{\text{max}}$ from the principal point to an illuminated
pixel. Its knots are uniform in the image
(\cref{sec:supp-vignetting-uniform-knots}).

\section{Calibration by Inverse Rendering}
\label{sec:calibration-pipeline}
We calibrate the radiometric camera by inverse rendering: we render
the images of a calibration target with
\cref{eq:rad-flux-with-vignetting} and fit them to the captured RAW
images. The rendered flux depends on the scene as well as on the
camera, so we estimate jointly the poses $\vbTheta_e$, the
projection $\vbTheta_d$, the focus surface $\vbTheta_f$, the aperture
radius $r$ and the vignetting $\vbTheta_v$ of the camera, the
lighting $\vbTheta_l$, the material $\vbTheta_m$ and the shape
$\vbTheta_b$ of the target. We denote all of them by
$\bar{\vbTheta}$. Let $\phi_j(\vbu)$ be the value of the RAW
photosite $\vbu$ in view $j$ and $\sigma_j(\vbu)$ its standard
deviation under the Poisson--Gaussian noise model of the sensor. We
minimize
\begin{equation}
\label{eq:objective}
    \sum_{j}\sum_{\vbu\in\mathcal{S}_j}\!
    \Bigl(\frac{\phi_j(\vbu)-\Phi_j(\vbu;\bar{\vbTheta})}{\sigma_j(\vbu)}\Bigr)^{2}\!
    + \lambda\bigl\lVert \mathrm{D}^3(\hat{\vbd}-\hat{\vbd}^{0})\bigr\rVert^{2}
\end{equation}
over $\bar{\vbTheta}$, where $\mathcal{S}_j$ are photosites drawn
uniformly from those of view $j$ that see the target. The first term
is the negative log-likelihood of the photosites, so that for
$\lambda=0$ the minimizer is the maximum-likelihood estimate of the
camera, the poses and the scene. The second term regularizes the
central spline model only (\cref{sec:optimization}). The remainder of
this section gives what the objective requires: derivatives without
boundary terms, a lighting model, the
reflectance and the shape of the target, and a initialization value.

\subsection{Differentiation without Boundary Terms}
\label{sec:boundary-terms}
\begin{figure}[t]
        \centering
        \usetikzlibrary{arrows.meta}%
\resizebox{\linewidth}{!}{%
\begin{tikzpicture}[x=1cm, y=1cm, >=stealth, font=\footnotesize,
    smp/.style={fill=blue!70!black, draw=none},
    smpn/.style={fill=white, draw=blue!70!black, line width=0.4pt},
    bnd/.style={line width=0.7pt, black},
    moved/.style={line width=0.7pt, red!75!black, densely dashed},
    dim/.style={line width=0.3pt, black!60}]
  \def\pts{0/0, 0.36/40, 0.36/160, 0.36/280, 0.70/100, 0.70/220, 0.70/340, 0.58/5, 0.64/55, 0.60/185}
  \node[anchor=south] at (2.1,1.3) {\textbf{boundary moves}};
  \node[anchor=south] at (6.2,1.3) {\textbf{reformulated}};
  \node[rotate=90, anchor=south] at (0.55,0) {aperture radius $r$};
  \node[rotate=90, anchor=south] at (0.55,-3.0) {pattern edge};
  \draw[dim] (4.05,1.75) -- (4.05,-4.1);
  \fill[red!22, even odd rule] (2.1,0) circle (0.98) (2.1,0) circle (0.82);
  \draw[bnd] (2.1,0) circle (0.82);
  \draw[moved] (2.1,0) circle (0.98);
  \foreach \r/\a in \pts { \fill[smp] ($(2.1,0)+(\a:\r)$) circle (1.4pt); }
  \foreach \r/\a in {1.20/25, 1.18/75, 1.20/130, 1.18/185, 1.20/235, 1.18/285} { \fill[smp] ($(2.1,0)+(\a:\r)$) circle (1.4pt); }
  \node[anchor=west, inner sep=1pt] at (3.18,-0.92) {$\mathcal{A}(r)$};
  \draw[dim] (2.74,-0.52) -- (3.20,-0.88);
  \node[anchor=north] at (2.1,-1.36) {no sample changes};
  \draw[bnd] (4.85,0) circle (0.5);
  \foreach \r/\a in \pts { \fill[smp] ($(4.85,0)+(\a:0.6098*\r)$) circle (1.1pt); }
  \node[anchor=south] at (4.85,0.52) {$\mathcal{D}\ni\vbxi$};
  \draw[->, line width=0.5pt] (5.42,0) -- (5.82,0);
  \node[anchor=north, inner sep=1pt] at (5.4,-0.75) {$\vba=\vbo+r\,\vbxi$};
  \draw[bnd] (7.0,0) circle (0.82);
  \draw[moved] (7.0,0) circle (0.98);
  \foreach \r/\a in \pts {
    \draw[blue!70!black, line width=0.5pt] ($(7.0,0)+(\a:\r)$) -- ($(7.0,0)+(\a:1.1951*\r)$);
    \fill[smp] ($(7.0,0)+(\a:\r)$) circle (1.4pt);
    \draw[smpn] ($(7.0,0)+(\a:1.1951*\r)$) circle (1.4pt);
  }
  \node[anchor=north] at (7.0,-1.36) {every sample moves};
  \draw[dim, ->] (0.95,-3.75) -- (3.25,-3.75);
  \draw[dim, ->] (0.95,-3.75) -- (0.95,-2.15);
  \node[anchor=east, inner sep=2pt] at (0.95,-2.3) {$\rho$};
  \fill[red!22] (2.0,-3.6) rectangle (2.16,-2.5);
  \draw[bnd] (0.95,-3.6) -- (2.0,-3.6) -- (2.0,-2.5) -- (3.15,-2.5);
  \draw[moved] (2.16,-3.6) -- (2.16,-2.5);
  \foreach \x/\y in {1.2/-3.6, 1.6/-3.6, 2.55/-2.5, 2.95/-2.5} { \fill[smp] (\x,\y) circle (1.5pt); }
  \node[anchor=north] at (2.1,-3.78) {no sample changes};
  \draw[dim, ->] (4.55,-3.75) -- (7.95,-3.75);
  \draw[dim, ->] (4.55,-3.75) -- (4.55,-2.15);
  \node[anchor=east, inner sep=2pt] at (4.55,-2.3) {$\rho$};
  \draw[bnd, domain=4.55:7.85, samples=60, smooth] plot (\x, {-3.05 + 0.55*tanh((\x-6.05)/0.42)});
  \draw[moved, domain=4.55:7.85, samples=60, smooth] plot (\x, {-3.05 + 0.55*tanh((\x-6.22)/0.42)});
  \foreach \x in {5.45, 5.85, 6.25, 6.65, 7.05} {
    \draw[blue!70!black, line width=0.4pt] (\x, {-3.05 + 0.55*tanh((\x-6.05)/0.42)}) -- (\x, {-3.05 + 0.55*tanh((\x-6.22)/0.42)});
    \fill[smp] (\x, {-3.05 + 0.55*tanh((\x-6.05)/0.42)}) circle (1.5pt);
    \draw[smpn] (\x, {-3.05 + 0.55*tanh((\x-6.22)/0.42)}) circle (1.5pt);
  }
  \node[anchor=north] at (6.25,-3.78) {every sample changes};
\end{tikzpicture}}%
        \caption{Fixed samples (dots) miss a boundary that moves with a parameter (solid black to dashed red): none lies in the region that it sweeps, so the sampled gradient lacks its contribution (left). We remove both boundaries of the calibration (right). \emph{Top:} the samples of the aperture are placed from the fixed unit disc $\mathcal{D}$ and move with its radius $r$. \emph{Bottom:} the blur of the print makes the albedo $\rho$ smooth across an edge of the pattern. Hollow dots: after the change.}
        \label{fig:boundary}
\end{figure}
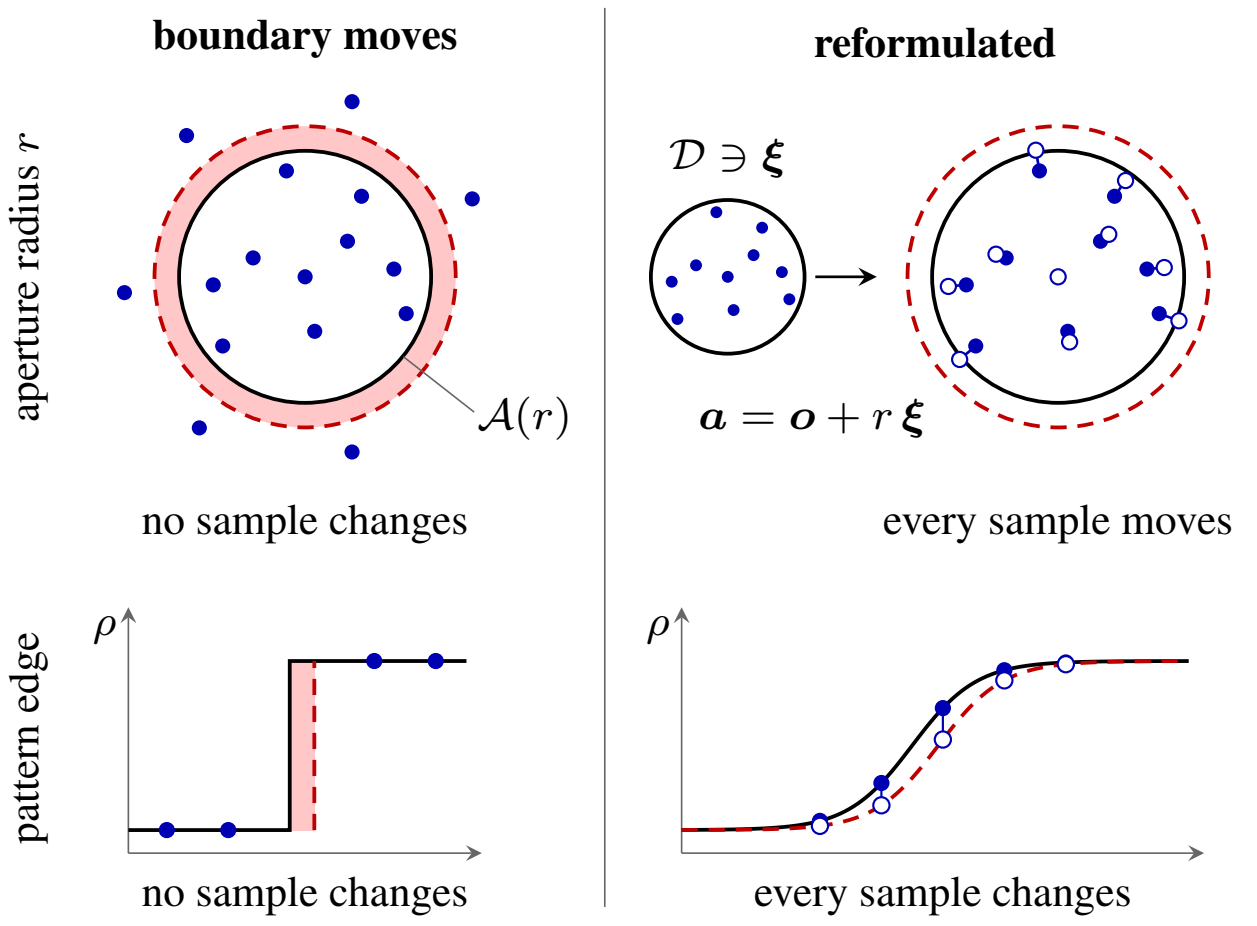
The derivative of an integral whose domain, or whose integrand's
discontinuities, depend on a parameter has a boundary term, which
automatic differentiation of a sampled estimate misses
(\cref{fig:boundary}, left). Differentiable renderers estimate such
terms by sampling the boundaries
\cite{Li:2018:DMC, Zhang:2020:PSDR, mitsuba-projective}. In the
calibration, three of them arise: The changing aperture radius, border of the calibration target, and sharp texture edges. We avoid these as follows.
The change of aperture radius is addressed by integrating over the fixed unit disc as described above in \cref{sec:vignetting}. 
Second, the hard edges of the calibration pattern are discontinuities of the albedo that
move over the aperture and over the pixel with the camera, the pose
and the target shape. We model the print as its design convolved
with a Gaussian, a $10$--$70\,\%$ edge width of $50\,\mathrm{\mu m}$
\cite{rosalen2019cardboards}, so that the albedo is smooth. The
boundary terms of a general scene, from silhouettes, shadows and
the visibility at every bounce of a ray
(\cref{sec:supp-boundaries}), do not arise in the calibration: we only 
render the masked target $\mathcal{M}\subseteq\mathcal{I}$, which is considered smoothly illuminated within a constant scene.

\subsection{Lighting}
\label{sec:lighting}
We represent the light of the environment by spherical harmonics of
degree two \cite{ramamoorthi-2001, peter-pikesloanand2002precomputed},
$L^{\text{SH}}(\vbomega;\vbTheta_l)$, with the coefficients
$\vbTheta_l$ per color channel (\cref{sec:supp-reflectance}). A
calibration target is planar, so every point of it sees the same
hemisphere, and a lighting at infinite distance gives the same
irradiance at every point of the target: any shading across the
target would have to be explained by the vignetting. We therefore
place the lighting on a sphere of finite radius $r_{\text{SH}}$
about the center $\vbb_0$ of the target, the larger of the diagonal
of the target and the largest distance of a camera from $\vbb_0$ at
the start. The radiance that arrives at $\vbx$ from the direction
$\vbomega$ is $L^{\text{SH}}$ in the direction
\begin{equation}
    \hat{\vbomega}(\vbx,\vbomega) \coloneqq \frac{\vbx + \gamma\vbomega - \vbb_0}{\lVert \vbx + \gamma\vbomega - \vbb_0 \rVert}
\end{equation}
from $\vbb_0$ to the point where the ray from $\vbx$ along $\vbomega$
meets the sphere, with $\gamma>0$ such that
$\lVert \vbx + \gamma \vbomega - \vbb_0\rVert = r_{\text{SH}}$. The
lighting then varies over the target and separates from the
vignetting.

\subsection{Reflectance}
\label{sec:reflectance}
Calibration targets usually consist of a pattern printed on background material. To model its appearance properly, we estimate the reflectance distribution of these materials. The radiance that leaves a scene point $\vbx$ in the direction $\vbomega_o$ is~\cite{pharr2023physically}
\begin{align}
    &L(\vbx,\vbomega_o)\label{eq:scene-rad}\\
    &\coloneqq \int_{\mathcal{H}^2(\vbx)}f(\vbx, \vbomega_o,\vbomega_i)L(\vbx'(\vbx,\vbomega_i),-\vbomega_i)\cos\theta_i\odif\vbomega_i\nonumber,
\end{align}
where $\mathcal{H}^2(\vbx)$ is the upper hemisphere over $\vbx$,
$\vbx'(\vbx,\vbomega_i)$ the scene or light point visible from $\vbx$
along $\vbomega_i$, $\theta_i$ the angle between $\vbomega_i$ and the
surface normal, and $f$ the bidirectional scattering distribution
function. On the target, the incident radiance is the lighting of
\cref{sec:lighting}. For $f$, we use the Cook--Torrance model
\cite{cookTorrance} with an isotropic GGX microfacet distribution of
roughness $\alpha$ \cite{Heitz2014Microfacet} and the Fresnel term
of a dielectric with the index of refraction $\eta$, and a weight
$\mu\in[0,1]$ between its diffuse part $\rho(\vbx)/\pi$, with the
albedo $\rho$ of the print, and its specular part. We evaluate the
integral by Monte Carlo sampling of the visible microfacet normals
\cite{sphericalCaps} (\cref{sec:supp-reflectance}). Ink fills the
microfacets of the print, so that inked areas are more specular than
blank ones. We therefore blend the roughness and the index of
refraction between a value for ink and one for the blank print by
the albedo, $\vbTheta_m = \{\alpha_{\neg\text{ink}},
\alpha_{\text{ink}},\eta_{\neg\text{ink}}, \eta_{\text{ink}}, \mu\}$.

\begin{figure*}[t]
        \centering
        \providecommand{\rcfigdir}{fig}%
\begingroup
\ifdefined\rcpanelwd\else\newlength\rcpanelwd\fi\ifdefined\rcpanelht\else\newlength\rcpanelht\fi
\setlength{\rcpanelwd}{\dimexpr(\linewidth-55pt)/6\relax}\setlength{\rcpanelht}{0.66917\rcpanelwd}%
\def\rcpanelwidth{\rcpanelwd}%
\ifdefined\rcflowkeybox\else\newsavebox\rcflowkeybox\fi
\sbox\rcflowkeybox{\scriptsize\input{\rcfigdir/detector-bias-mitsuba/flow-key}}%
\ifdefined\rcflowkeybeyondbox\else\newsavebox\rcflowkeybeyondbox\fi
\sbox\rcflowkeybeyondbox{\scriptsize\input{\rcfigdir/detector-bias-mitsuba/flow-key-beyond}}%
\footnotesize\setlength{\tabcolsep}{0pt}%
\begin{tabular}{@{}c@{\hspace{2pt}}c@{\hspace{10pt}}c@{\hspace{10pt}}c@{\hspace{1.5pt}}c@{\hspace{10pt}}c@{\hspace{1.5pt}}c@{}}
 & Render & \makebox[\rcpanelwd][c]{True Projected vs.\ Detected} & \multicolumn{2}{c@{\hspace{10pt}}}{Calibrated Camera \& Pose} & \multicolumn{2}{c@{}}{Calibrated Camera} \\
 & & & {\scriptsize Geometric} & {\scriptsize Radiometric} & {\scriptsize Geometric} & {\scriptsize Radiometric} \\\noalign{\vskip 2pt}
\raisebox{\dimexpr-0.5\height-0\rcpanelht\relax}[0pt][0pt]{\rotatebox{90}{\footnotesize \begin{tabular}{@{}c@{}}OpenCV +\\AprilGrid\end{tabular}}} & \providecommand{\rcfigdir}{fig}%
\providecommand{\rcpanelwidth}{\linewidth}%
\begin{tikzpicture}[baseline=(current axis.center)]
\begin{axis}[
  scale only axis, width=\rcpanelwidth, height=0.66917*\rcpanelwidth,
  grid=none, log ticks with fixed point, tick label style={font=\scriptsize}, label style={font=\small}, legend cell align=left, legend style={font=\scriptsize, draw=none, fill=none},
  xmin=0, xmax=4256, ymin=0, ymax=2848, enlargelimits=false,
  xtick=\empty, ytick=\empty, axis on top, axis line style={draw=none}]
\addplot graphics[xmin=0, xmax=4256, ymin=0, ymax=2848] {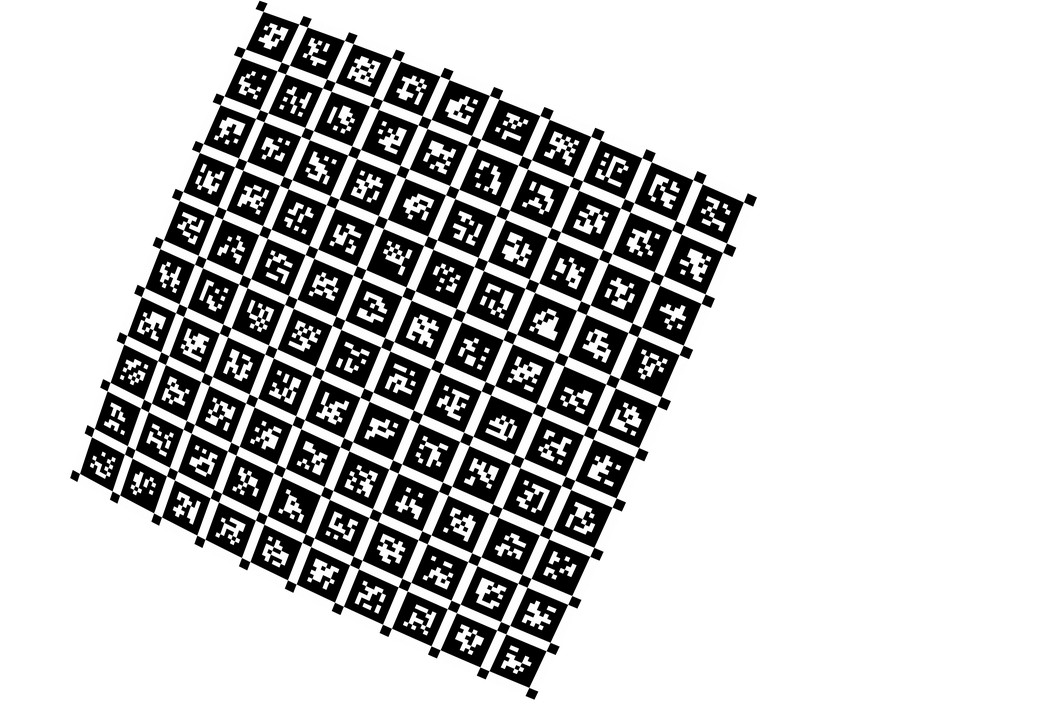};
\end{axis}
\end{tikzpicture}%
\makebox[0pt][r]{\makebox[\rcpanelwd][c]{\raisebox{\dimexpr0.5\rcpanelht-1.5pt\relax}[0pt][0pt]{\tikz[baseline=(k.north)]\node[fill=white,inner sep=1.2pt,font=\scriptsize](k){FOV};}}} & \providecommand{\rcfigdir}{fig}%
\providecommand{\rcpanelwidth}{\linewidth}%
\begin{tikzpicture}[baseline=(current axis.center)]
\begin{axis}[
  scale only axis, width=\rcpanelwidth, height=0.66917*\rcpanelwidth,
  grid=none, log ticks with fixed point, tick label style={font=\scriptsize}, label style={font=\small}, legend cell align=left, legend style={font=\scriptsize, draw=none, fill=none},
  xmin=0, xmax=4256, ymin=0, ymax=2848, enlargelimits=false,
  xtick=\empty, ytick=\empty, axis on top, axis line style={draw=none}]
\addplot graphics[xmin=0, xmax=4256, ymin=0, ymax=2848] {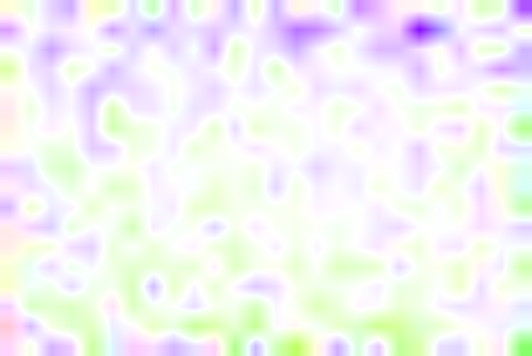};
\end{axis}
\end{tikzpicture}%
\makebox[0pt][r]{\raisebox{\dimexpr-0.5\rcpanelht+1.5pt\relax}[0pt][0pt]{\tikz[baseline=(k.south)]\node[fill=white,inner sep=1.2pt](k){\usebox\rcflowkeybox\,{\scriptsize $0.24$\,px}};}\kern2pt} & \providecommand{\rcfigdir}{fig}%
\providecommand{\rcpanelwidth}{\linewidth}%
\begin{tikzpicture}[baseline=(current axis.center)]
\begin{axis}[
  scale only axis, width=\rcpanelwidth, height=0.66917*\rcpanelwidth,
  grid=none, log ticks with fixed point, tick label style={font=\scriptsize}, label style={font=\small}, legend cell align=left, legend style={font=\scriptsize, draw=none, fill=none},
  xmin=0, xmax=4256, ymin=0, ymax=2848, enlargelimits=false,
  xtick=\empty, ytick=\empty, axis on top, axis line style={draw=none}]
\addplot graphics[xmin=0, xmax=4256, ymin=0, ymax=2848] {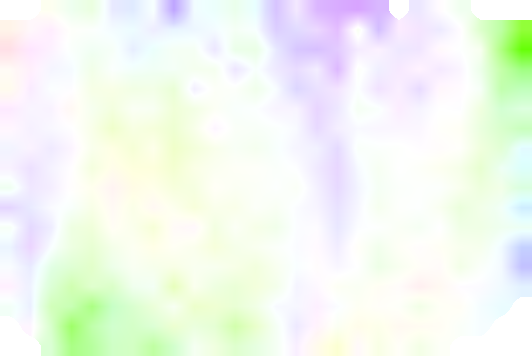};
\end{axis}
\end{tikzpicture}%
 & \providecommand{\rcfigdir}{fig}%
\providecommand{\rcpanelwidth}{\linewidth}%
\begin{tikzpicture}[baseline=(current axis.center)]
\begin{axis}[
  scale only axis, width=\rcpanelwidth, height=0.66917*\rcpanelwidth,
  grid=none, log ticks with fixed point, tick label style={font=\scriptsize}, label style={font=\small}, legend cell align=left, legend style={font=\scriptsize, draw=none, fill=none},
  xmin=0, xmax=4256, ymin=0, ymax=2848, enlargelimits=false,
  xtick=\empty, ytick=\empty, axis on top, axis line style={draw=none}]
\addplot graphics[xmin=0, xmax=4256, ymin=0, ymax=2848] {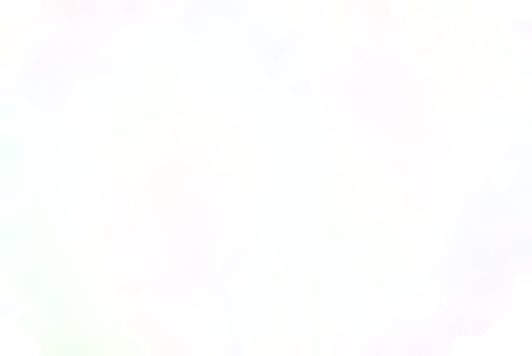};
\end{axis}
\end{tikzpicture}%
\makebox[0pt][r]{\makebox[\rcpanelwd][c]{\raisebox{-0.5ex}[0pt][0pt]{\scriptsize\textcolor{black!35}{error not visible}}}}\makebox[0pt][r]{\makebox[\dimexpr2\rcpanelwd+1.5pt\relax][c]{\raisebox{\dimexpr-0.5\rcpanelht+1.5pt\relax}[0pt][0pt]{\tikz[baseline=(k.south)]\node[fill=white,inner sep=1.2pt](k){\usebox\rcflowkeybox\,{\scriptsize $0.39$\,px}};}}}\makebox[0pt][r]{\makebox[\dimexpr2\rcpanelwd+1.5pt\relax][c]{\raisebox{\dimexpr0.5\rcpanelht-1.5pt\relax}[0pt][0pt]{\tikz[baseline=(k.north)]\node[fill=white,inner sep=1.2pt,font=\scriptsize](k){OpenCV-14};}}} & \providecommand{\rcfigdir}{fig}%
\providecommand{\rcpanelwidth}{\linewidth}%
\begin{tikzpicture}[baseline=(current axis.center)]
\begin{axis}[
  scale only axis, width=\rcpanelwidth, height=0.66917*\rcpanelwidth,
  grid=none, log ticks with fixed point, tick label style={font=\scriptsize}, label style={font=\small}, legend cell align=left, legend style={font=\scriptsize, draw=none, fill=none},
  xmin=0, xmax=4256, ymin=0, ymax=2848, enlargelimits=false,
  xtick=\empty, ytick=\empty, axis on top, axis line style={draw=none}]
\addplot graphics[xmin=0, xmax=4256, ymin=0, ymax=2848] {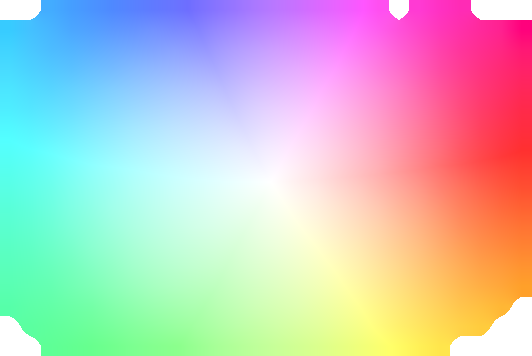};
\end{axis}
\end{tikzpicture}%
 & \providecommand{\rcfigdir}{fig}%
\providecommand{\rcpanelwidth}{\linewidth}%
\begin{tikzpicture}[baseline=(current axis.center)]
\begin{axis}[
  scale only axis, width=\rcpanelwidth, height=0.66917*\rcpanelwidth,
  grid=none, log ticks with fixed point, tick label style={font=\scriptsize}, label style={font=\small}, legend cell align=left, legend style={font=\scriptsize, draw=none, fill=none},
  xmin=0, xmax=4256, ymin=0, ymax=2848, enlargelimits=false,
  xtick=\empty, ytick=\empty, axis on top, axis line style={draw=none}]
\addplot graphics[xmin=0, xmax=4256, ymin=0, ymax=2848] {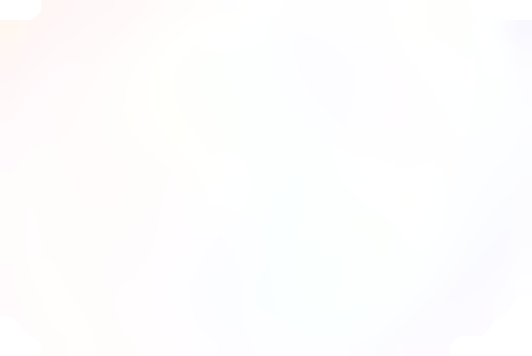};
\end{axis}
\end{tikzpicture}%
\makebox[0pt][r]{\makebox[\rcpanelwd][c]{\raisebox{-0.5ex}[0pt][0pt]{\scriptsize\textcolor{black!35}{error not visible}}}}\makebox[0pt][r]{\makebox[\dimexpr2\rcpanelwd+1.5pt\relax][c]{\raisebox{\dimexpr-0.5\rcpanelht+1.5pt\relax}[0pt][0pt]{\tikz[baseline=(k.south)]\node[fill=white,inner sep=1.2pt](k){\usebox\rcflowkeybox\,{\scriptsize $0.54$\,px}};}}}\makebox[0pt][r]{\makebox[\dimexpr2\rcpanelwd+1.5pt\relax][c]{\raisebox{\dimexpr0.5\rcpanelht-1.5pt\relax}[0pt][0pt]{\tikz[baseline=(k.north)]\node[fill=white,inner sep=1.2pt,font=\scriptsize](k){OpenCV-14, global rotation removed};}}} \\\noalign{\vskip 1.5pt}
\raisebox{\dimexpr-0.5\height-0\rcpanelht\relax}[0pt][0pt]{\rotatebox{90}{\footnotesize \begin{tabular}{@{}c@{}}Sch\"ops +\\Siemens stars\end{tabular}}} & \providecommand{\rcfigdir}{fig}%
\providecommand{\rcpanelwidth}{\linewidth}%
\begin{tikzpicture}[baseline=(current axis.center)]
\begin{axis}[
  scale only axis, width=\rcpanelwidth, height=0.66917*\rcpanelwidth,
  grid=none, log ticks with fixed point, tick label style={font=\scriptsize}, label style={font=\small}, legend cell align=left, legend style={font=\scriptsize, draw=none, fill=none},
  xmin=0, xmax=4256, ymin=0, ymax=2848, enlargelimits=false,
  xtick=\empty, ytick=\empty, axis on top, axis line style={draw=none}]
\addplot graphics[xmin=0, xmax=4256, ymin=0, ymax=2848] {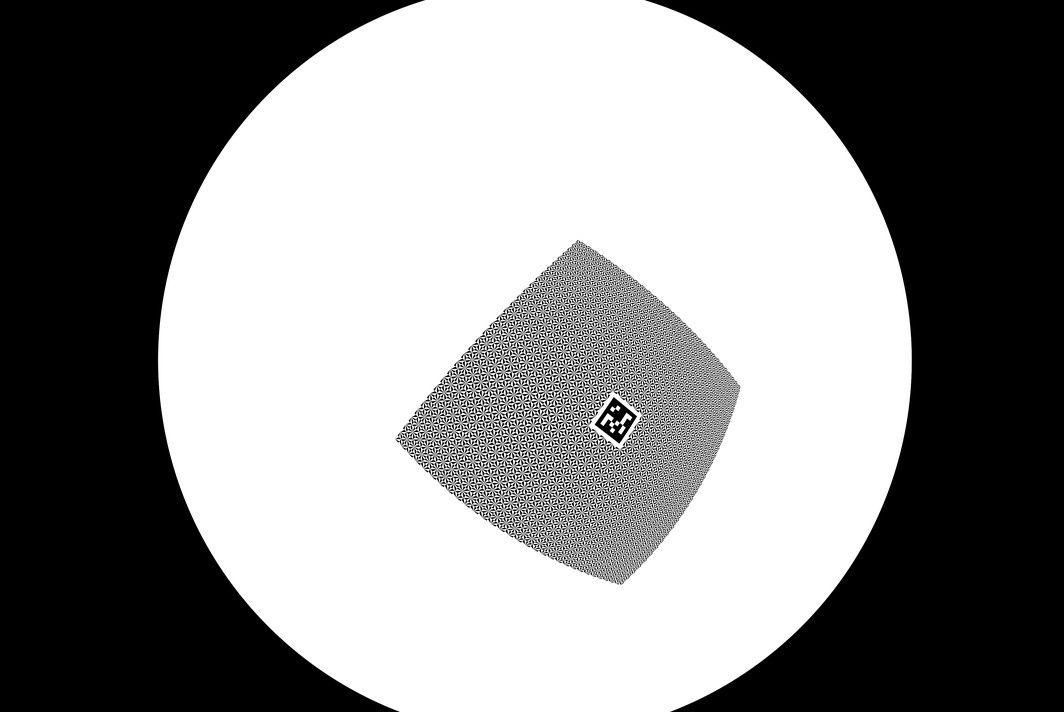};
\end{axis}
\end{tikzpicture}%
\makebox[0pt][r]{\makebox[\rcpanelwd][c]{\raisebox{\dimexpr0.5\rcpanelht-1.5pt\relax}[0pt][0pt]{\tikz[baseline=(k.north)]\node[fill=white,inner sep=1.2pt,font=\scriptsize](k){Double Sphere};}}} & \providecommand{\rcfigdir}{fig}%
\providecommand{\rcpanelwidth}{\linewidth}%
\begin{tikzpicture}[baseline=(current axis.center)]
\begin{axis}[
  scale only axis, width=\rcpanelwidth, height=0.66917*\rcpanelwidth,
  grid=none, log ticks with fixed point, tick label style={font=\scriptsize}, label style={font=\small}, legend cell align=left, legend style={font=\scriptsize, draw=none, fill=none},
  xmin=0, xmax=4256, ymin=0, ymax=2848, enlargelimits=false,
  xtick=\empty, ytick=\empty, axis on top, axis line style={draw=none}]
\addplot graphics[xmin=0, xmax=4256, ymin=0, ymax=2848] {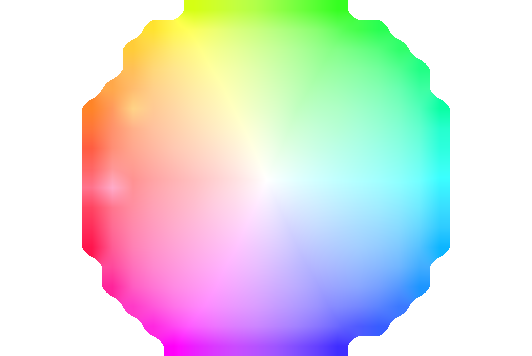};
\end{axis}
\end{tikzpicture}%
\makebox[0pt][r]{\raisebox{\dimexpr-0.5\rcpanelht+1.5pt\relax}[0pt][0pt]{\tikz[baseline=(k.south)]\node[fill=white,inner sep=1.2pt](k){\usebox\rcflowkeybox\,{\scriptsize $0.067$\,px}};}\kern2pt} & \providecommand{\rcfigdir}{fig}%
\providecommand{\rcpanelwidth}{\linewidth}%
\begin{tikzpicture}[baseline=(current axis.center)]
\begin{axis}[
  scale only axis, width=\rcpanelwidth, height=0.66917*\rcpanelwidth,
  grid=none, log ticks with fixed point, tick label style={font=\scriptsize}, label style={font=\small}, legend cell align=left, legend style={font=\scriptsize, draw=none, fill=none},
  xmin=0, xmax=4256, ymin=0, ymax=2848, enlargelimits=false,
  xtick=\empty, ytick=\empty, axis on top, axis line style={draw=none}]
\addplot graphics[xmin=0, xmax=4256, ymin=0, ymax=2848] {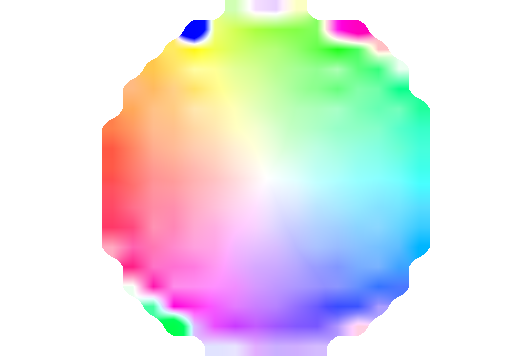};
\end{axis}
\end{tikzpicture}%
 & \providecommand{\rcfigdir}{fig}%
\providecommand{\rcpanelwidth}{\linewidth}%
\begin{tikzpicture}[baseline=(current axis.center)]
\begin{axis}[
  scale only axis, width=\rcpanelwidth, height=0.66917*\rcpanelwidth,
  grid=none, log ticks with fixed point, tick label style={font=\scriptsize}, label style={font=\small}, legend cell align=left, legend style={font=\scriptsize, draw=none, fill=none},
  xmin=0, xmax=4256, ymin=0, ymax=2848, enlargelimits=false,
  xtick=\empty, ytick=\empty, axis on top, axis line style={draw=none}]
\addplot graphics[xmin=0, xmax=4256, ymin=0, ymax=2848] {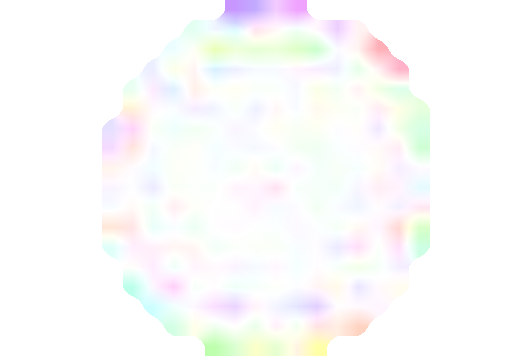};
\end{axis}
\end{tikzpicture}%
\makebox[0pt][r]{\makebox[\dimexpr2\rcpanelwd+1.5pt\relax][c]{\raisebox{\dimexpr-0.5\rcpanelht+1.5pt\relax}[0pt][0pt]{\tikz[baseline=(k.south)]\node[fill=white,inner sep=1.2pt](k){\usebox\rcflowkeybeyondbox\,{\scriptsize $\geq0.053$\,px}};}}}\makebox[0pt][r]{\makebox[\dimexpr2\rcpanelwd+1.5pt\relax][c]{\raisebox{\dimexpr0.5\rcpanelht-1.5pt\relax}[0pt][0pt]{\tikz[baseline=(k.north)]\node[fill=white,inner sep=1.2pt,font=\scriptsize](k){Spline};}}} & \providecommand{\rcfigdir}{fig}%
\providecommand{\rcpanelwidth}{\linewidth}%
\begin{tikzpicture}[baseline=(current axis.center)]
\begin{axis}[
  scale only axis, width=\rcpanelwidth, height=0.66917*\rcpanelwidth,
  grid=none, log ticks with fixed point, tick label style={font=\scriptsize}, label style={font=\small}, legend cell align=left, legend style={font=\scriptsize, draw=none, fill=none},
  xmin=0, xmax=4256, ymin=0, ymax=2848, enlargelimits=false,
  xtick=\empty, ytick=\empty, axis on top, axis line style={draw=none}]
\addplot graphics[xmin=0, xmax=4256, ymin=0, ymax=2848] {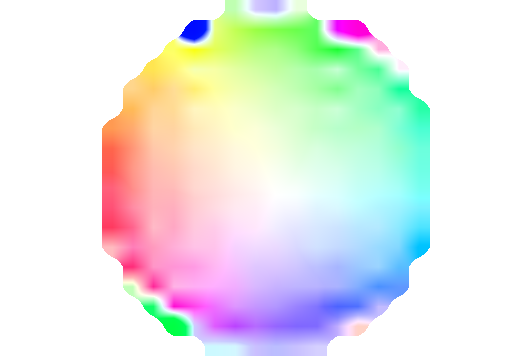};
\end{axis}
\end{tikzpicture}%
 & \providecommand{\rcfigdir}{fig}%
\providecommand{\rcpanelwidth}{\linewidth}%
\begin{tikzpicture}[baseline=(current axis.center)]
\begin{axis}[
  scale only axis, width=\rcpanelwidth, height=0.66917*\rcpanelwidth,
  grid=none, log ticks with fixed point, tick label style={font=\scriptsize}, label style={font=\small}, legend cell align=left, legend style={font=\scriptsize, draw=none, fill=none},
  xmin=0, xmax=4256, ymin=0, ymax=2848, enlargelimits=false,
  xtick=\empty, ytick=\empty, axis on top, axis line style={draw=none}]
\addplot graphics[xmin=0, xmax=4256, ymin=0, ymax=2848] {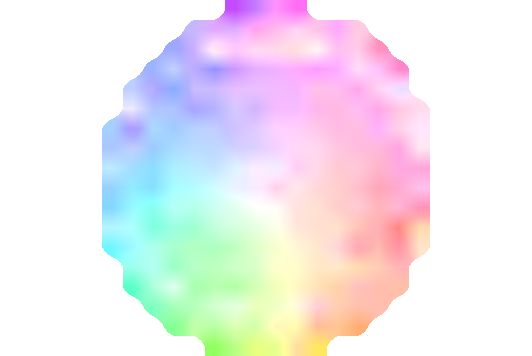};
\end{axis}
\end{tikzpicture}%
\makebox[0pt][r]{\makebox[\dimexpr2\rcpanelwd+1.5pt\relax][c]{\raisebox{\dimexpr-0.5\rcpanelht+1.5pt\relax}[0pt][0pt]{\tikz[baseline=(k.south)]\node[fill=white,inner sep=1.2pt](k){\usebox\rcflowkeybeyondbox\,{\scriptsize $\geq0.048$\,px}};}}}\makebox[0pt][r]{\makebox[\dimexpr2\rcpanelwd+1.5pt\relax][c]{\raisebox{\dimexpr0.5\rcpanelht-1.5pt\relax}[0pt][0pt]{\tikz[baseline=(k.north)]\node[fill=white,inner sep=1.2pt,font=\scriptsize](k){Spline, global rotation removed};}}} \\
\end{tabular}
\endgroup

        \caption{The bias of a detector (column 2) is absorbed by a geometrically calibrated camera (columns 3 and 4). With the calibrated poses, the bias is pushed into the camera model (column 3); with the true poses (column 4), it shows that the poses compensate for part of it. Our radiometric refinement does not rely on a detector and removes most of the bias in both evaluations.}
        \label{fig:detector-bias}
\end{figure*}
\subsection{Target Shape}
\label{sec:target-shape}
A printed target is not exactly planar, and its print is not exactly
its design. We chart the board by $\vbb\in\mathcal{B}$, in units of
its width, and model its shape by a height field
$h:\mathcal{B}\rightarrow\mathbb{R}$ along the board normal
$\vbn_{\mathcal{B}}$. In a view with the pose
$(\vbR,\vbt)\in\vbTheta_e$, the chart point $\vbb$ is the scene
point
\begin{equation}
\label{eq:board-surface}
    \vbs(\vbb) \coloneqq \vbR\bigl(\vbb_{\mathcal{B}} + h(\vbb)\,\vbn_{\mathcal{B}}\bigr) + \vbt,
\end{equation}
where $\vbb_{\mathcal{B}}$ is the point of $\vbb$ on the plane of the
board. The height field is a thin plate spline
\cite{thin-plate-spline} through the heights
$\vbh\in\mathbb{R}^{25}$ at a $5\times5$ grid of nodes. A thin plate
spline is the sum of an affine function and a bending, and affine
heights, an offset and two tilts, describe a plane, which is already
represented by the pose. We therefore restrict the heights to the
pure bendings, $\vbh=\vbB\vby$, where the columns of
$\vbB\in\mathbb{R}^{25\times22}$ span the nodal heights whose
spline has no affine part, and optimize $\vbTheta_b\coloneqq\vby$.
The shape then has no gauge freedom with the poses, without a
penalty or a fixed node (\cref{sec:supp-gauge-freedoms}), and only
22 parameters. For the spline camera, we also estimate the in-plane error of the print by
an aspect and a skew, at which the albedo is read.

The scene point of \cref{eq:rad-flux-with-vignetting} is the
intersection of the ray from $\vba$ in the direction
$\vbd\coloneqq\vbpi^{-1}(\vbp,\vba)$ with this surface. We solve for
its chart point: with an orthonormal pair $\vbe_1,\vbe_2$
perpendicular to $\vbd$, $\vbg(\vbb) \coloneqq
    \begin{pmatrix}\vbe_1 & \vbe_2\end{pmatrix}^{\!\top}
    \bigl(\vbs(\vbb)-\vba\bigr) = \mathbf{0}$,
a root of two equations in two unknowns that we find by
Levenberg--Marquardt iterations from the intersection with the plane
of the board. It gives $\vbx(\vbp,\vba)=\vbs(\vbb)$ together with the
chart point $\vbb$ at which the albedo is read. The root is
differentiated \wrt the camera, the pose and the heights by the
implicit function theorem,
$\partial\vbb/\partial\theta=-(\partial\vbg/\partial\vbb)^{-1}\,\partial\vbg/\partial\theta$,
and not through its iterations. The effect of the shape on the
reflectance is small, so the reflectance uses the normal of the
plane.

The reconstruction of the calibration target is consistent across three lenses and two boards: the learned board deformations agree to an RMS of less than $0.7\,\mathrm{mm}$ for every pair of lenses (\cref{sec:board-warping}).

\subsection{Initialization and Optimization}
\label{sec:optimization}
The projection and the poses start from a geometric calibration
(\cref{sec:supp-starts}), the target as a plane, the focus surface
as the plane parallel to the sensor at the mean distance of the
target, and the aperture radius as $f/(2N)$ with the focal length
$f$ and the f-number $N$ of the lens. The flux is linear in the
vignetting and in the lighting and affine in $\mu$
(\cref{sec:supp-linear}), so they start from one pass of linear
least-squares solves: the lighting with $V\equiv1$, then the
vignetting, then $\mu$. We minimize \cref{eq:objective} with the
Levenberg--Marquardt method \cite{marquardt1963algorithm}, with
$\lambda=0$ for the parametric camera models. The nodes of the central spline at
the border of the lens coverage are hardly constrained by the
photosites. For this model only, the second term regularizes the
change of the nodal directions $\hat{\vbd}\subseteq\bar{\vbTheta}$
from their start $\hat{\vbd}^{0}$ with a P-spline penalty
\cite{eilers1996flexible}, where $\mathrm{D}^3$ takes the third finite
differences, the discrete change of curvature, along both axes of the
grid and $\lambda=0.1$ relative to the data term. It
vanishes for a locally quadratic change, so the penalty leaves a
smooth change of the camera free and suppresses its oscillation, and
the border nodes follow the interior.
No other component is regularized. The parameters stay in their
physical ranges by their parameterization: we optimize $\log r$,
$\log\alpha$, $\log(\eta-1)$ and the logit of $\mu$, the poses on
$SE(3)$ \cite{solà2021microlietheorystate}, and the inverse focus
distances are kept non-negative. The gauge freedoms are removed by
the parameterization (\cref{sec:supp-gauge-freedoms}).

\section{Experiments}
\label{sec:experiments}
\label{sec:dataset}
We show that geometric calibration is biased by its feature detections
(\cref{sec:detector-bias}) and that the radiometric camera improves
the geometric accuracy (\cref{sec:geometric-accuracy}), compare the
learned vignetting with a measured flat field
(\cref{sec:experiment-vignetting}) and validate the focus model
(\cref{sec:experiment-defocus}). All captures are RAW images of a Sony
ILME-FX3 ($4256\times2848$ pixels) and are listed in
\cref{sec:supp-dataset} together with the runtime for each experiment; all errors are in pixels of this image. 

\begin{table*}[t]
        \centering
        \footnotesize
\setlength{\tabcolsep}{3pt}
\begin{tabular*}{\linewidth}{@{\extracolsep{\fill}}lcccccc@{}}
\toprule
 & \multicolumn{2}{c}{\textbf{Tamron 20\,mm}} & \multicolumn{2}{c}{\textbf{Sigma 14\,mm}} & \multicolumn{2}{c}{\textbf{Sigma 8\,mm fisheye}} \\
\cmidrule(lr){2-3}\cmidrule(lr){4-5}\cmidrule(lr){6-7}
\textbf{Model} & AprilGrid & Siemens stars & AprilGrid & Siemens stars & AprilGrid & Siemens stars \\
\midrule
\multicolumn{7}{@{}l}{\underline{\textit{Median reprojection error [px]}}} \\
Brown--Conrady & 0.119\,$\rightarrow$\,\textbf{0.067} & 0.288\,$\rightarrow$\,\textbf{0.135} & 0.499\,$\rightarrow$\,\textbf{0.377} & 0.687\,$\rightarrow$\,\textbf{0.359} & -- & -- \\
OpenCV-14 & 0.107\,$\rightarrow$\,\textbf{0.060} & 0.308\,$\rightarrow$\,\textbf{0.131} & 0.476\,$\rightarrow$\,\textbf{0.359} & 0.600\,$\rightarrow$\,\textbf{0.384} & 1.98\,$\rightarrow$\,\textbf{1.70} & 0.123\,$\rightarrow$\,\textbf{0.079} \\
Kannala--Brandt & 0.117\,$\rightarrow$\,\textbf{0.106} & 0.183\,$\rightarrow$\,0.176 & 0.456\,$\rightarrow$\,\textbf{0.351} & 0.567\,$\rightarrow$\,\textbf{0.329} & 2.00\,$\rightarrow$\,\textbf{1.52} & 0.113\,$\rightarrow$\,\textbf{0.063} \\
Double Sphere & 0.261\,$\rightarrow$\,\textit{0.286} & 0.219\,$\rightarrow$\,\textit{0.268} & 1.42\,$\rightarrow$\,1.34 & 0.734\,$\rightarrow$\,\textbf{0.461} & 2.09\,$\rightarrow$\,\textbf{1.49} & 0.109\,$\rightarrow$\,\textbf{0.067} \\
Central spline & 0.040\,$\rightarrow$\,\textit{0.048} & 0.161\,$\rightarrow$\,\textbf{0.136} & 0.479\,$\rightarrow$\,\textbf{0.331} & 0.504\,$\rightarrow$\,\textbf{0.366} & 1.99\,$\rightarrow$\,\textbf{1.45} & 0.121\,$\rightarrow$\,\textbf{0.100} \\
\midrule
\multicolumn{7}{@{}l}{\underline{\textit{95th percentile reprojection error [px]}}} \\
Brown--Conrady & 0.356\,$\rightarrow$\,\textbf{0.173} & 0.855\,$\rightarrow$\,\textbf{0.321} & 2.07\,$\rightarrow$\,\textbf{1.78} & 2.73\,$\rightarrow$\,\textbf{1.72} & -- & -- \\
OpenCV-14 & 0.373\,$\rightarrow$\,\textbf{0.155} & 0.969\,$\rightarrow$\,\textbf{0.337} & 2.16\,$\rightarrow$\,\textbf{1.74} & 2.87\,$\rightarrow$\,\textbf{1.93} & 9.15\,$\rightarrow$\,\textbf{5.68} & 0.818\,$\rightarrow$\,\textbf{0.440} \\
Kannala--Brandt & 0.446\,$\rightarrow$\,\textbf{0.285} & 0.730\,$\rightarrow$\,\textbf{0.515} & 1.99\,$\rightarrow$\,\textbf{1.79} & 2.37\,$\rightarrow$\,\textbf{1.75} & 6.80\,$\rightarrow$\,\textbf{5.43} & 0.590\,$\rightarrow$\,\textbf{0.254} \\
Double Sphere & 1.70\,$\rightarrow$\,\textbf{1.16} & 0.717\,$\rightarrow$\,\textbf{0.625} & 5.13\,$\rightarrow$\,\textbf{4.42} & 2.42\,$\rightarrow$\,\textbf{1.94} & 7.41\,$\rightarrow$\,\textbf{5.45} & 0.591\,$\rightarrow$\,\textbf{0.289} \\
Central spline & 0.131\,$\rightarrow$\,\textbf{0.119} & 0.436\,$\rightarrow$\,\textbf{0.331} & 2.45\,$\rightarrow$\,\textbf{1.76} & 2.35\,$\rightarrow$\,\textbf{1.97} & 7.58\,$\rightarrow$\,\textbf{5.51} & 0.996\,$\rightarrow$\,\textbf{0.538} \\
\bottomrule
\end{tabular*}

        \caption{Reprojection error on the held-out high-precision target, geometric start $\rightarrow$ radiometric refinement, for each camera model, lens and calibration target. The refined value is bold where it is significantly lower than the start and italic where it is significantly higher (95\,\% bootstrap interval of the difference, \cref{tab:supp-refinement-a}); the mean and the RMS are in \cref{tab:supp-refinement-b}. --: Brown--Conrady cannot represent the $180^\circ$ field of view of the fisheye.}
        \label{tab:refinement-chrome}
\end{table*}
\subsection{Detector Bias}
\label{sec:detector-bias}
Feature detectors introduce a measurement bias that degrades the
regressed camera model. We rendered 10,000 images each of the
AprilGrid \cite{furgale2013unified,olson2011apriltag} with an FOV
camera model \cite{devernay2001straight}, which has a mild
distortion, and of the Siemens-star target of Schöps \etal
\cite{10000params} with a Double Sphere model \cite{double-sphere},
which has a strong fisheye distortion, in evenly lit calibration
scenes. The second column of \cref{fig:detector-bias} shows
the displacement between the true projection of each feature and its
detection, for the AprilGrid detector of OpenCV \cite{opencv_library}
(top) and the Siemens-star detector of Schöps \etal (bottom). Both
detectors exhibit a bias that grows toward the image border. The third
column projects the features with the camera and the poses of a
calibration. In the geometric calibration, OpenCV-14 for the AprilGrid
and the central spline for the Siemens stars, the detection bias is
absorbed by both the camera and the poses, whereas the radiometric
refinement largely removes it. The fourth column substitutes the true
poses for the estimated ones, which isolates the camera model. This
column is the relevant one: the poses are discarded after the
calibration, and a downstream task depends on the camera alone.
The geometrically calibrated camera has absorbed a strong bias from
the feature detections, also for the highly flexible spline of Schöps
\etal, and our radiometric refinement removes most of it.

\subsection{Geometric Accuracy}
\label{sec:geometric-accuracy}
We lift four commonly used central parametric camera models,
Brown--Conrady \cite{Brown1966DecenteringDO}, the 14-parameter model
of OpenCV \cite{opencv_library}, Kannala--Brandt \cite{kannala-brandt}
and Double Sphere \cite{double-sphere}, and the central generic spline
model of Schöps \etal \cite{10000params} to the radiometric camera. We
calibrate them on three lenses chosen for their diverse fields of view
and distortions (\cref{tab:dataset-lenses}): a Tamron 20\,mm
wide-angle lens with barrel distortion ($94^\circ$ diagonal field of
view), a Sigma 14\,mm rectilinear wide-angle lens ($114^\circ$) and a
Sigma 8\,mm circular fisheye ($180^\circ$). We captured two
standard calibration targets, the Siemens-star target of Schöps \etal
\cite{10000params} and an AprilGrid \cite{furgale2013unified}, each
$1\,\mathrm{m}\times1\,\mathrm{m}$ and printed directly on Alu-Dibond,
at $f/22$ (the fisheye's Siemens
stars at $f/32$) and ISO\,100, to minimize defocus and thus give
optimal conditions to the geometric camera models, which do not model
defocus. Each model is started from the geometric calibration of its
native tool where the tool supports the model and the target, and
from our own estimator otherwise (\cref{sec:supp-starts}). The
radiometric refinement samples 8,192 photosites per image for the
parametric models and 16,384 for the spline.

The test set consists of backlit images of the Thorlabs R3L3S7N1
high-precision distortion target \cite{thorlabsR3L3S7N1}, captured
by rotating the camera about a pivot, which reduces the poses to one
rotation per view. Every dot is evaluated with a pose that it did
not help to fit (\cref{sec:supp-chrome-evaluation}). 

\begin{figure}[t]
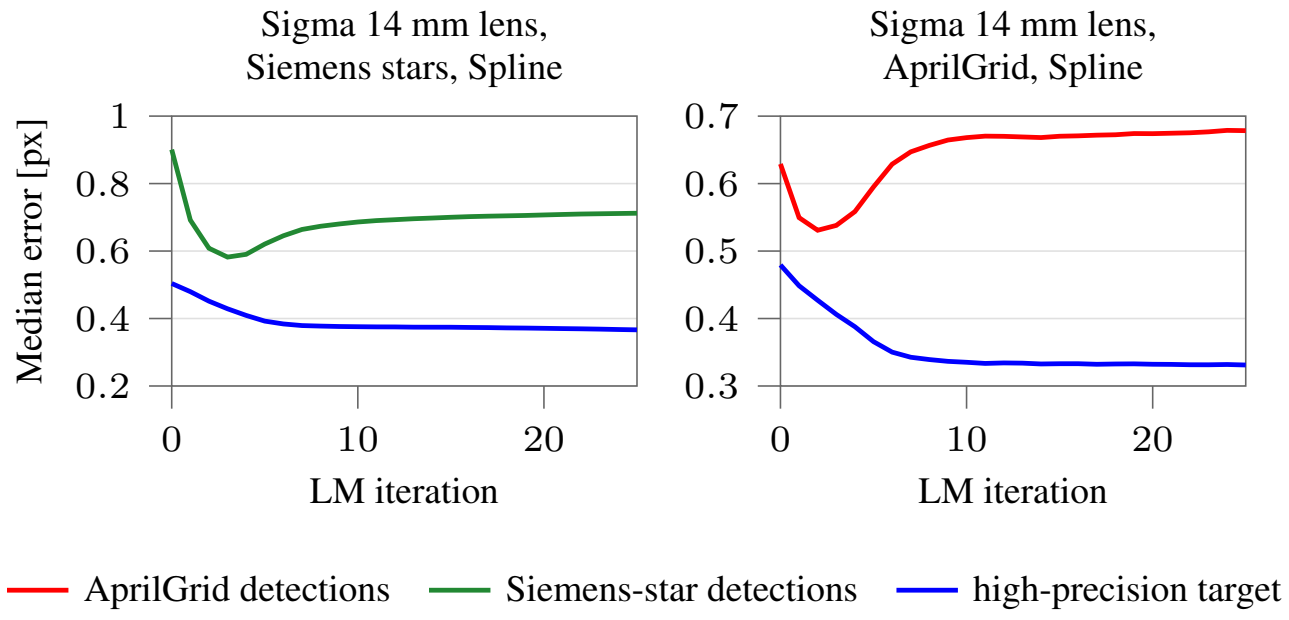

        \centering
        \providecommand{\rcfigdir}{fig}%
\begingroup
\ifdefined\rcpanelwd\else\newlength\rcpanelwd\fi\def\rcpanelwidth{\rcpanelwd}%
\setlength{\tabcolsep}{0pt}\definecolor{rcchr}{HTML}{0000FF}\definecolor{rcapr}{HTML}{FF0000}\definecolor{rcsie}{HTML}{228833}%
\def\rckey#1#2{\textcolor{#1}{\rule[0.5ex]{0.16in}{0.9pt}}\hspace{0.04in}#2}%
\ifdim\linewidth<4.5in\relax
  \setlength{\rcpanelwd}{\dimexpr(\linewidth-62pt)/2\relax}%
  \begin{tabular}{@{}l@{\hspace{4pt}}l@{}}
  {\def\rcpanelylabel{Median error [px]}\hspace*{31pt}\providecommand{\rcpanelwidth}{\linewidth}\providecommand{\rcpanelylabel}{Median error [px]}%
\definecolor{rcdet}{HTML}{228833}\definecolor{rcchr}{HTML}{0000FF}%
\begin{tikzpicture}[baseline=(current bounding box.south), trim axis left, trim axis right]
\begin{axis}[width=\rcpanelwidth, height=0.58*\rcpanelwidth, scale only axis, xmin=0, xmax=25, xtick distance=10, ymin=0.2, ymax=1, ytick distance=0.2, tick label style={font=\scriptsize}, label style={font=\scriptsize}, tick align=outside, tick style={black!60, line width=0.3pt}, axis line style={black!60, line width=0.3pt}, scaled y ticks=false, title style={font=\scriptsize, align=center, yshift=-4pt}, title={Sigma 14 mm lens,\\Siemens stars, Spline}, xlabel={LM iteration}, xtick pos=left, ytick pos=left, ymajorgrids, grid style={black!12, line width=0.3pt}, clip=false, yticklabel style={/pgf/number format/fixed, /pgf/number format/precision=2}, ylabel={\rcpanelylabel}, ylabel style={yshift=-2pt}, xlabel style={yshift=2pt}]
\addplot[rcchr, line width=0.9pt, no marks] coordinates {(0,0.503858) (1,0.479635) (2,0.451468) (3,0.428849) (4,0.409212) (5,0.392331) (6,0.383918) (7,0.379111) (8,0.377569) (9,0.376345) (10,0.375741) (11,0.375234) (12,0.375083) (13,0.374388) (14,0.374266) (15,0.37419) (16,0.373537) (17,0.373086) (18,0.372198) (19,0.371721) (20,0.370976) (21,0.370155) (22,0.36948) (23,0.368566) (24,0.3674) (25,0.366428)};
\addplot[rcdet, line width=0.9pt, no marks] coordinates {(0,0.900719) (1,0.691795) (2,0.608336) (3,0.582069) (4,0.590293) (5,0.620715) (6,0.645302) (7,0.663994) (8,0.673493) (9,0.680254) (10,0.686109) (11,0.690268) (12,0.692967) (13,0.695779) (14,0.697832) (15,0.700138) (16,0.702033) (17,0.703369) (18,0.704402) (19,0.705472) (20,0.70701) (21,0.708444) (22,0.709772) (23,0.710538) (24,0.711302) (25,0.712088)};
\end{axis}
\path ([yshift=-0.40in]current axis.south) -- ++(0pt,-1.9ex);
\end{tikzpicture}%
\hspace*{4pt}} &
  {\def\rcpanelylabel{}\hspace*{19pt}\providecommand{\rcpanelwidth}{\linewidth}\providecommand{\rcpanelylabel}{Median error [px]}%
\definecolor{rcdet}{HTML}{FF0000}\definecolor{rcchr}{HTML}{0000FF}%
\begin{tikzpicture}[baseline=(current bounding box.south), trim axis left, trim axis right]
\begin{axis}[width=\rcpanelwidth, height=0.58*\rcpanelwidth, scale only axis, xmin=0, xmax=25, xtick distance=10, ymin=0.3, ymax=0.7, ytick distance=0.1, tick label style={font=\scriptsize}, label style={font=\scriptsize}, tick align=outside, tick style={black!60, line width=0.3pt}, axis line style={black!60, line width=0.3pt}, scaled y ticks=false, title style={font=\scriptsize, align=center, yshift=-4pt}, title={Sigma 14 mm lens,\\AprilGrid, Spline}, xlabel={LM iteration}, xtick pos=left, ytick pos=left, ymajorgrids, grid style={black!12, line width=0.3pt}, clip=false, yticklabel style={/pgf/number format/fixed, /pgf/number format/precision=2}, ylabel={\rcpanelylabel}, ylabel style={yshift=-2pt}, xlabel style={yshift=2pt}]
\addplot[rcchr, line width=0.9pt, no marks] coordinates {(0,0.479401) (1,0.448594) (2,0.427077) (3,0.406175) (4,0.387851) (5,0.365746) (6,0.350214) (7,0.342611) (8,0.33918) (9,0.336529) (10,0.335144) (11,0.333448) (12,0.334201) (13,0.333871) (14,0.332649) (15,0.333017) (16,0.333033) (17,0.332171) (18,0.332636) (19,0.33282) (20,0.332211) (21,0.332024) (22,0.331501) (23,0.331445) (24,0.331851) (25,0.331107)};
\addplot[rcdet, line width=0.9pt, no marks] coordinates {(0,0.629097) (1,0.549551) (2,0.530836) (3,0.538027) (4,0.558502) (5,0.595073) (6,0.628786) (7,0.647052) (8,0.656638) (9,0.664582) (10,0.668141) (11,0.670354) (12,0.670069) (13,0.669146) (14,0.668246) (15,0.670206) (16,0.670782) (17,0.671808) (18,0.67239) (19,0.674187) (20,0.674022) (21,0.674677) (22,0.675312) (23,0.676721) (24,0.678901) (25,0.678465)};
\end{axis}
\path ([yshift=-0.40in]current axis.south) -- ++(0pt,-1.9ex);
\end{tikzpicture}%
\hspace*{4pt}} \\[-9pt]
  \multicolumn{2}{@{}c@{}}{\scriptsize \rckey{rcapr}{AprilGrid detections}\hspace{0.1in}\rckey{rcsie}{Siemens-star detections}\hspace{0.1in}\rckey{rcchr}{high-precision target}} \\
  \end{tabular}%
\else
  \setlength{\rcpanelwd}{\dimexpr(\linewidth-80pt)/2\relax}%
  \begin{tabular}{@{}l@{\hspace{10pt}}l@{}}
  {\def\rcpanelylabel{Median reprojection error [px]}\hspace*{36pt}\input{\rcfigdir/convergence-rms/si30_spline_warped}\hspace*{6pt}} &
  {\def\rcpanelylabel{}\hspace*{22pt}\input{\rcfigdir/convergence-rms/ag30_spline_warped}\hspace*{6pt}} \\[3pt]
  \multicolumn{2}{@{}c@{}}{\scriptsize \rckey{rcapr}{AprilGrid detections}\hspace{0.15in}\rckey{rcsie}{Siemens-star detections}\hspace{0.15in}\rckey{rcchr}{high-precision target}} \\
  \end{tabular}%
\fi
\endgroup

        \caption{Median reprojection error during the radiometric refinement of the central spline on the Sigma 14\,mm, with respect to the detections that the geometric calibration was fitted to and on the held-out high-precision target. After the first iterations the curves diverge in both panels: the error on the detections rises while the error on the held-out target falls, which shows that the geometric calibration had absorbed the bias of its detector.}
        \label{fig:convergence-rms}
\end{figure}

\Cref{tab:refinement-chrome} compares the reprojection error of each
model at its geometric start and after the radiometric refinement on
the Thorlabs test set (\cref{tab:supp-dataset}). We call a change significant where the 95\,\% bootstrap interval of
the paired difference excludes zero (\cref{sec:supp-chrome-evaluation}).
The radiometric refinement lowers the
median reprojection error significantly in 82\,\% of the 28\
combinations of model, lens and target (23, by 9 to 57\,\%), raises
it significantly in 11\,\% (three, all on the Tamron 20\,mm), and
lowers the $95$th percentile significantly in all of them, by 9 to
65\,\%. On the Sigma 14\,mm and on the fisheye, no model is
significantly worse in any statistic.
\Cref{fig:convergence-rms} follows the two refinements of the
central spline on the Sigma 14\,mm over their iterations. In both,
the error on the held-out target falls throughout, while the error
with respect to the detections that the geometric calibration was
fitted to rises after the first four iterations (from $0.59$ to
$0.71$\,px on the Siemens stars and from $0.56$ to $0.68$\,px on the
AprilGrid). The test error decreases while the training error
increases, which shows the bias: the refinement moves the camera
away from the biased detections that its start was fitted to. We attribute part of this bias to the freedom of the
geometric calibration: The feature wise parametrization of the target in \cite{10000params} allows to absorb the bias, while our lower parameter parametrization (\cref{sec:target-shape}) cannot. 

\subsection{Vignetting}
\label{sec:experiment-vignetting}
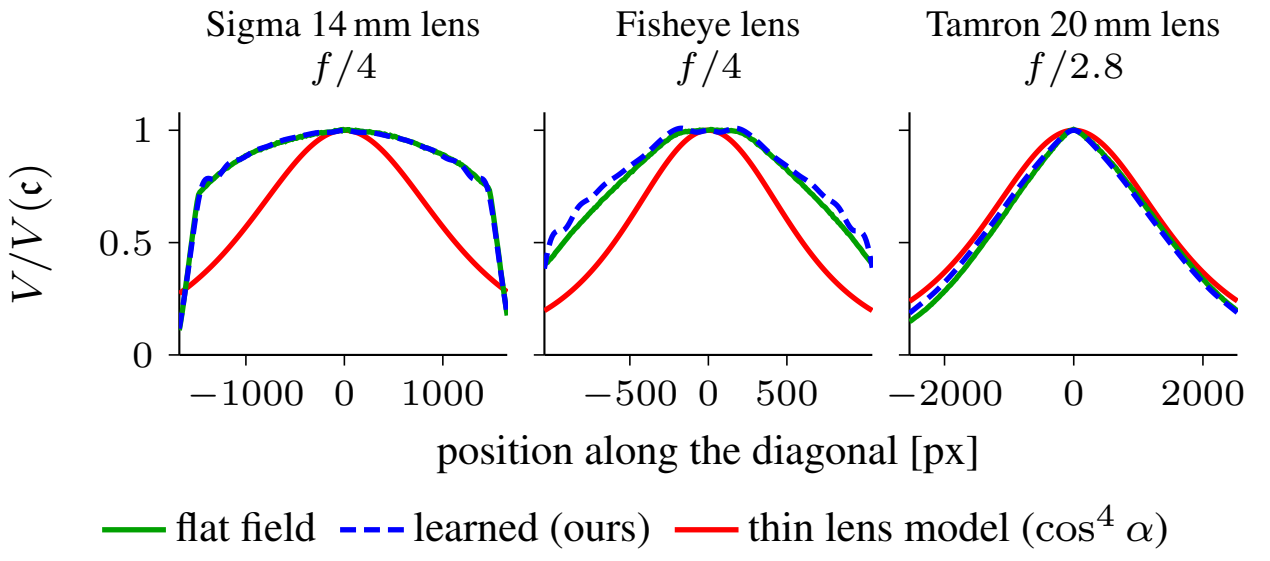
\begin{figure}[t]
        \centering
        \providecommand{\rcfigdir}{fig}%
\begingroup
\ifdefined\rcpanelwd\else\newlength\rcpanelwd\fi
\setlength{\rcpanelwd}{\dimexpr(\linewidth-52pt)/3\relax}\def\rcpanelwidth{\rcpanelwd}%
\definecolor{rcvlearned}{HTML}{0000FF}\definecolor{rcvflat}{HTML}{00A000}\definecolor{rcvthin}{HTML}{FF0000}%
\footnotesize\setlength{\tabcolsep}{0pt}%
\begin{tabular}{@{}c@{\hspace{24pt}}c@{\hspace{7pt}}c@{\hspace{7pt}}c@{\hspace{4pt}}}
  & \makebox[0pt]{\scriptsize\begin{tabular}[b]{@{}c@{}}Sigma 14\,mm lens\\$f/4$\end{tabular}} & \makebox[0pt]{\scriptsize\begin{tabular}[b]{@{}c@{}}Fisheye lens\\$f/4$\end{tabular}} & \makebox[0pt]{\scriptsize\begin{tabular}[b]{@{}c@{}}Tamron 20\,mm lens\\$f/2.8$\end{tabular}} \\[2pt]
  \smash{\raisebox{-0.5\height}{\rotatebox{90}{$V / V(\mathfrak{c})$}}} &
  \providecommand{\rcpanelwidth}{\linewidth}%
\begin{tikzpicture}[baseline=(current axis.center), trim axis left, trim axis right]
\definecolor{rcvlearned}{HTML}{0000FF}\definecolor{rcvflat}{HTML}{00A000}\definecolor{rcvthin}{HTML}{FF0000}
\begin{axis}[
  scale only axis, width=\rcpanelwidth, height=0.74*\rcpanelwidth,
  grid=none, log ticks with fixed point, tick label style={font=\scriptsize}, label style={font=\small}, legend cell align=left, legend style={font=\scriptsize, draw=none, fill=none},
  axis x line*=bottom, axis y line*=left, axis line style={line width=0.4pt}, tick align=outside, tick style={black, line width=0.3pt},
  major tick length=2pt, scaled ticks=false, /pgf/number format/1000 sep={}, enlargelimits=false,
  xmin=-1676, xmax=1654, xtick={-1000,0,1000}, ymin=0, ymax=1.08, ytick={0,0.5,1}]
\addplot[rcvthin, line width=1pt] coordinates {(-1676.5,0.2734) (-1662.5,0.2778) (-1648.5,0.2822) (-1634.5,0.2868) (-1620.5,0.2914) (-1606.5,0.2960) (-1592.5,0.3008) (-1578.5,0.3056) (-1564.5,0.3105) (-1550.5,0.3154) (-1536.5,0.3205) (-1522.5,0.3256) (-1508.5,0.3307) (-1494.5,0.3360) (-1480.5,0.3413) (-1466.5,0.3467) (-1452.5,0.3522) (-1438.5,0.3578) (-1424.5,0.3634) (-1410.5,0.3691) (-1396.5,0.3749) (-1382.5,0.3808) (-1368.5,0.3867) (-1354.5,0.3927) (-1340.5,0.3988) (-1326.5,0.4050) (-1312.5,0.4113) (-1298.5,0.4176) (-1284.5,0.4240) (-1270.5,0.4305) (-1256.5,0.4371) (-1242.5,0.4437) (-1228.5,0.4504) (-1214.5,0.4572) (-1200.5,0.4641) (-1186.5,0.4710) (-1172.5,0.4780) (-1158.5,0.4851) (-1144.5,0.4923) (-1130.5,0.4995) (-1116.5,0.5068) (-1102.5,0.5141) (-1088.5,0.5216) (-1074.5,0.5290) (-1060.5,0.5366) (-1046.5,0.5442) (-1032.5,0.5519) (-1018.5,0.5596) (-1004.5,0.5674) (-990.5,0.5752) (-976.5,0.5831) (-962.5,0.5910) (-948.5,0.5990) (-934.5,0.6070) (-920.5,0.6150) (-906.5,0.6231) (-892.5,0.6313) (-878.5,0.6394) (-864.5,0.6476) (-850.5,0.6558) (-836.5,0.6640) (-822.5,0.6723) (-808.5,0.6805) (-794.5,0.6888) (-780.5,0.6970) (-766.5,0.7053) (-752.5,0.7135) (-738.5,0.7218) (-724.5,0.7300) (-710.5,0.7382) (-696.5,0.7464) (-682.5,0.7545) (-668.5,0.7626) (-654.5,0.7707) (-640.5,0.7787) (-626.5,0.7867) (-612.5,0.7946) (-598.5,0.8025) (-584.5,0.8103) (-570.5,0.8180) (-556.5,0.8256) (-542.5,0.8332) (-528.5,0.8406) (-514.5,0.8480) (-500.5,0.8552) (-486.5,0.8624) (-472.5,0.8694) (-458.5,0.8763) (-444.5,0.8831) (-430.5,0.8897) (-416.5,0.8962) (-402.5,0.9026) (-388.5,0.9088) (-374.5,0.9148) (-360.5,0.9207) (-346.5,0.9264) (-332.5,0.9319) (-318.5,0.9372) (-304.5,0.9424) (-290.5,0.9474) (-276.5,0.9521) (-262.5,0.9567) (-248.5,0.9611) (-234.5,0.9652) (-220.5,0.9692) (-206.5,0.9729) (-192.5,0.9764) (-178.5,0.9796) (-164.5,0.9827) (-150.5,0.9855) (-136.5,0.9880) (-122.5,0.9903) (-108.5,0.9924) (-94.5,0.9942) (-80.5,0.9958) (-66.5,0.9971) (-52.5,0.9982) (-38.5,0.9990) (-24.5,0.9996) (-10.5,0.9999) (3.5,1.0000) (17.5,0.9998) (31.5,0.9994) (45.5,0.9987) (59.5,0.9977) (73.5,0.9965) (87.5,0.9950) (101.5,0.9933) (115.5,0.9914) (129.5,0.9892) (143.5,0.9868) (157.5,0.9841) (171.5,0.9812) (185.5,0.9780) (199.5,0.9746) (213.5,0.9710) (227.5,0.9672) (241.5,0.9632) (255.5,0.9589) (269.5,0.9545) (283.5,0.9498) (297.5,0.9449) (311.5,0.9398) (325.5,0.9346) (339.5,0.9292) (353.5,0.9235) (367.5,0.9177) (381.5,0.9118) (395.5,0.9057) (409.5,0.8994) (423.5,0.8930) (437.5,0.8864) (451.5,0.8797) (465.5,0.8729) (479.5,0.8659) (493.5,0.8588) (507.5,0.8516) (521.5,0.8443) (535.5,0.8369) (549.5,0.8294) (563.5,0.8218) (577.5,0.8141) (591.5,0.8064) (605.5,0.7986) (619.5,0.7907) (633.5,0.7827) (647.5,0.7747) (661.5,0.7667) (675.5,0.7586) (689.5,0.7505) (703.5,0.7423) (717.5,0.7341) (731.5,0.7259) (745.5,0.7176) (759.5,0.7094) (773.5,0.7011) (787.5,0.6929) (801.5,0.6846) (815.5,0.6764) (829.5,0.6681) (843.5,0.6599) (857.5,0.6517) (871.5,0.6435) (885.5,0.6353) (899.5,0.6272) (913.5,0.6191) (927.5,0.6110) (941.5,0.6030) (955.5,0.5950) (969.5,0.5870) (983.5,0.5791) (997.5,0.5713) (1011.5,0.5635) (1025.5,0.5557) (1039.5,0.5480) (1053.5,0.5404) (1067.5,0.5328) (1081.5,0.5253) (1095.5,0.5178) (1109.5,0.5104) (1123.5,0.5031) (1137.5,0.4959) (1151.5,0.4887) (1165.5,0.4816) (1179.5,0.4745) (1193.5,0.4675) (1207.5,0.4606) (1221.5,0.4538) (1235.5,0.4471) (1249.5,0.4404) (1263.5,0.4338) (1277.5,0.4273) (1291.5,0.4208) (1305.5,0.4144) (1319.5,0.4081) (1333.5,0.4019) (1347.5,0.3958) (1361.5,0.3897) (1375.5,0.3837) (1389.5,0.3778) (1403.5,0.3720) (1417.5,0.3663) (1431.5,0.3606) (1445.5,0.3550) (1459.5,0.3495) (1473.5,0.3440) (1487.5,0.3387) (1501.5,0.3334) (1515.5,0.3281) (1529.5,0.3230) (1543.5,0.3179) (1557.5,0.3129) (1571.5,0.3080) (1585.5,0.3032) (1599.5,0.2984) (1613.5,0.2937) (1627.5,0.2891) (1641.5,0.2845) (1653.5,0.2806)};
\addplot[rcvflat, line width=1pt] coordinates {(-1676.5,0.1099) (-1662.5,0.1452) (-1648.5,0.1847) (-1634.5,0.2278) (-1620.5,0.2733) (-1606.5,0.3211) (-1592.5,0.3641) (-1578.5,0.4107) (-1564.5,0.4593) (-1550.5,0.5046) (-1536.5,0.5521) (-1522.5,0.5982) (-1508.5,0.6423) (-1494.5,0.6829) (-1480.5,0.7133) (-1466.5,0.7250) (-1452.5,0.7310) (-1438.5,0.7371) (-1424.5,0.7445) (-1410.5,0.7491) (-1396.5,0.7555) (-1382.5,0.7630) (-1368.5,0.7692) (-1354.5,0.7738) (-1340.5,0.7796) (-1326.5,0.7840) (-1312.5,0.7905) (-1298.5,0.7960) (-1284.5,0.8023) (-1270.5,0.8078) (-1256.5,0.8104) (-1242.5,0.8158) (-1228.5,0.8223) (-1214.5,0.8273) (-1200.5,0.8302) (-1186.5,0.8322) (-1172.5,0.8378) (-1158.5,0.8432) (-1144.5,0.8469) (-1130.5,0.8516) (-1116.5,0.8560) (-1102.5,0.8587) (-1088.5,0.8633) (-1074.5,0.8683) (-1060.5,0.8690) (-1046.5,0.8741) (-1032.5,0.8786) (-1018.5,0.8806) (-1004.5,0.8864) (-990.5,0.8887) (-976.5,0.8897) (-962.5,0.8951) (-948.5,0.8972) (-934.5,0.9005) (-920.5,0.9022) (-906.5,0.9060) (-892.5,0.9092) (-878.5,0.9114) (-864.5,0.9138) (-850.5,0.9132) (-836.5,0.9175) (-822.5,0.9212) (-808.5,0.9233) (-794.5,0.9227) (-780.5,0.9263) (-766.5,0.9281) (-752.5,0.9299) (-738.5,0.9339) (-724.5,0.9355) (-710.5,0.9353) (-696.5,0.9398) (-682.5,0.9442) (-668.5,0.9432) (-654.5,0.9461) (-640.5,0.9462) (-626.5,0.9483) (-612.5,0.9534) (-598.5,0.9500) (-584.5,0.9543) (-570.5,0.9553) (-556.5,0.9574) (-542.5,0.9605) (-528.5,0.9594) (-514.5,0.9638) (-500.5,0.9644) (-486.5,0.9640) (-472.5,0.9671) (-458.5,0.9705) (-444.5,0.9722) (-430.5,0.9695) (-416.5,0.9746) (-402.5,0.9766) (-388.5,0.9745) (-374.5,0.9784) (-360.5,0.9795) (-346.5,0.9778) (-332.5,0.9801) (-318.5,0.9813) (-304.5,0.9838) (-290.5,0.9863) (-276.5,0.9843) (-262.5,0.9876) (-248.5,0.9887) (-234.5,0.9897) (-220.5,0.9879) (-206.5,0.9893) (-192.5,0.9905) (-178.5,0.9918) (-164.5,0.9918) (-150.5,0.9924) (-136.5,0.9912) (-122.5,0.9931) (-108.5,0.9943) (-94.5,0.9986) (-80.5,0.9980) (-66.5,0.9972) (-52.5,1.0002) (-38.5,1.0019) (-24.5,0.9980) (-10.5,1.0025) (3.5,0.9997) (17.5,1.0010) (31.5,1.0027) (45.5,1.0027) (59.5,0.9975) (73.5,0.9996) (87.5,0.9988) (101.5,0.9970) (115.5,1.0001) (129.5,0.9984) (143.5,0.9983) (157.5,0.9962) (171.5,0.9956) (185.5,0.9954) (199.5,0.9936) (213.5,0.9921) (227.5,0.9932) (241.5,0.9921) (255.5,0.9939) (269.5,0.9920) (283.5,0.9892) (297.5,0.9894) (311.5,0.9879) (325.5,0.9886) (339.5,0.9861) (353.5,0.9880) (367.5,0.9862) (381.5,0.9833) (395.5,0.9806) (409.5,0.9831) (423.5,0.9804) (437.5,0.9755) (451.5,0.9742) (465.5,0.9777) (479.5,0.9746) (493.5,0.9711) (507.5,0.9730) (521.5,0.9716) (535.5,0.9676) (549.5,0.9665) (563.5,0.9634) (577.5,0.9611) (591.5,0.9591) (605.5,0.9568) (619.5,0.9563) (633.5,0.9546) (647.5,0.9524) (661.5,0.9501) (675.5,0.9488) (689.5,0.9478) (703.5,0.9453) (717.5,0.9419) (731.5,0.9403) (745.5,0.9369) (759.5,0.9339) (773.5,0.9323) (787.5,0.9300) (801.5,0.9275) (815.5,0.9212) (829.5,0.9221) (843.5,0.9209) (857.5,0.9177) (871.5,0.9153) (885.5,0.9115) (899.5,0.9107) (913.5,0.9045) (927.5,0.9058) (941.5,0.9015) (955.5,0.8992) (969.5,0.8976) (983.5,0.8944) (997.5,0.8893) (1011.5,0.8897) (1025.5,0.8865) (1039.5,0.8810) (1053.5,0.8755) (1067.5,0.8751) (1081.5,0.8712) (1095.5,0.8701) (1109.5,0.8661) (1123.5,0.8613) (1137.5,0.8572) (1151.5,0.8533) (1165.5,0.8543) (1179.5,0.8466) (1193.5,0.8457) (1207.5,0.8404) (1221.5,0.8365) (1235.5,0.8323) (1249.5,0.8257) (1263.5,0.8215) (1277.5,0.8148) (1291.5,0.8100) (1305.5,0.8060) (1319.5,0.8013) (1333.5,0.7954) (1347.5,0.7884) (1361.5,0.7821) (1375.5,0.7779) (1389.5,0.7702) (1403.5,0.7670) (1417.5,0.7614) (1431.5,0.7583) (1445.5,0.7514) (1459.5,0.7453) (1473.5,0.7325) (1487.5,0.7009) (1501.5,0.6590) (1515.5,0.6155) (1529.5,0.5695) (1543.5,0.5219) (1557.5,0.4797) (1571.5,0.4338) (1585.5,0.3840) (1599.5,0.3406) (1613.5,0.2973) (1627.5,0.2559) (1641.5,0.2118) (1653.5,0.1754)};
\addplot[rcvlearned, line width=1pt, dash pattern=on 3pt off 1.5pt] coordinates {(-1676.5,0.1174) (-1662.5,0.1550) (-1648.5,0.1932) (-1634.5,0.2322) (-1620.5,0.2721) (-1606.5,0.3132) (-1592.5,0.3557) (-1578.5,0.3995) (-1564.5,0.4443) (-1550.5,0.4893) (-1536.5,0.5337) (-1522.5,0.5769) (-1508.5,0.6179) (-1494.5,0.6560) (-1480.5,0.6902) (-1466.5,0.7196) (-1452.5,0.7431) (-1438.5,0.7607) (-1424.5,0.7730) (-1410.5,0.7807) (-1396.5,0.7847) (-1382.5,0.7858) (-1368.5,0.7851) (-1354.5,0.7834) (-1340.5,0.7817) (-1326.5,0.7813) (-1312.5,0.7830) (-1298.5,0.7867) (-1284.5,0.7921) (-1270.5,0.7989) (-1256.5,0.8067) (-1242.5,0.8151) (-1228.5,0.8238) (-1214.5,0.8323) (-1200.5,0.8401) (-1186.5,0.8468) (-1172.5,0.8522) (-1158.5,0.8564) (-1144.5,0.8597) (-1130.5,0.8622) (-1116.5,0.8641) (-1102.5,0.8657) (-1088.5,0.8671) (-1074.5,0.8687) (-1060.5,0.8705) (-1046.5,0.8729) (-1032.5,0.8759) (-1018.5,0.8793) (-1004.5,0.8830) (-990.5,0.8869) (-976.5,0.8910) (-962.5,0.8951) (-948.5,0.8992) (-934.5,0.9030) (-920.5,0.9064) (-906.5,0.9095) (-892.5,0.9122) (-878.5,0.9146) (-864.5,0.9168) (-850.5,0.9188) (-836.5,0.9207) (-822.5,0.9226) (-808.5,0.9245) (-794.5,0.9265) (-780.5,0.9287) (-766.5,0.9311) (-752.5,0.9335) (-738.5,0.9361) (-724.5,0.9388) (-710.5,0.9414) (-696.5,0.9440) (-682.5,0.9466) (-668.5,0.9491) (-654.5,0.9513) (-640.5,0.9534) (-626.5,0.9554) (-612.5,0.9572) (-598.5,0.9589) (-584.5,0.9604) (-570.5,0.9619) (-556.5,0.9633) (-542.5,0.9646) (-528.5,0.9659) (-514.5,0.9673) (-500.5,0.9686) (-486.5,0.9699) (-472.5,0.9713) (-458.5,0.9726) (-444.5,0.9739) (-430.5,0.9753) (-416.5,0.9767) (-402.5,0.9781) (-388.5,0.9795) (-374.5,0.9809) (-360.5,0.9823) (-346.5,0.9836) (-332.5,0.9849) (-318.5,0.9861) (-304.5,0.9872) (-290.5,0.9882) (-276.5,0.9890) (-262.5,0.9897) (-248.5,0.9901) (-234.5,0.9903) (-220.5,0.9904) (-206.5,0.9904) (-192.5,0.9904) (-178.5,0.9903) (-164.5,0.9904) (-150.5,0.9905) (-136.5,0.9909) (-122.5,0.9915) (-108.5,0.9924) (-94.5,0.9935) (-80.5,0.9948) (-66.5,0.9960) (-52.5,0.9973) (-38.5,0.9984) (-24.5,0.9993) (-10.5,0.9999) (3.5,1.0000) (17.5,0.9996) (31.5,0.9989) (45.5,0.9978) (59.5,0.9967) (73.5,0.9954) (87.5,0.9941) (101.5,0.9930) (115.5,0.9920) (129.5,0.9912) (143.5,0.9907) (157.5,0.9904) (171.5,0.9903) (185.5,0.9903) (199.5,0.9904) (213.5,0.9904) (227.5,0.9904) (241.5,0.9902) (255.5,0.9899) (269.5,0.9894) (283.5,0.9887) (297.5,0.9878) (311.5,0.9867) (325.5,0.9855) (339.5,0.9843) (353.5,0.9829) (367.5,0.9816) (381.5,0.9802) (395.5,0.9788) (409.5,0.9774) (423.5,0.9760) (437.5,0.9746) (451.5,0.9733) (465.5,0.9719) (479.5,0.9706) (493.5,0.9693) (507.5,0.9679) (521.5,0.9666) (535.5,0.9653) (549.5,0.9640) (563.5,0.9626) (577.5,0.9612) (591.5,0.9597) (605.5,0.9580) (619.5,0.9563) (633.5,0.9544) (647.5,0.9524) (661.5,0.9502) (675.5,0.9478) (689.5,0.9453) (703.5,0.9427) (717.5,0.9401) (731.5,0.9374) (745.5,0.9348) (759.5,0.9323) (773.5,0.9299) (787.5,0.9276) (801.5,0.9255) (815.5,0.9235) (829.5,0.9217) (843.5,0.9198) (857.5,0.9178) (871.5,0.9157) (885.5,0.9134) (899.5,0.9109) (913.5,0.9080) (927.5,0.9047) (941.5,0.9011) (955.5,0.8972) (969.5,0.8931) (983.5,0.8890) (997.5,0.8849) (1011.5,0.8811) (1025.5,0.8775) (1039.5,0.8743) (1053.5,0.8717) (1067.5,0.8695) (1081.5,0.8679) (1095.5,0.8664) (1109.5,0.8650) (1123.5,0.8632) (1137.5,0.8610) (1151.5,0.8582) (1165.5,0.8544) (1179.5,0.8496) (1193.5,0.8436) (1207.5,0.8363) (1221.5,0.8281) (1235.5,0.8195) (1249.5,0.8109) (1263.5,0.8027) (1277.5,0.7953) (1291.5,0.7892) (1305.5,0.7846) (1319.5,0.7819) (1333.5,0.7813) (1347.5,0.7825) (1361.5,0.7843) (1375.5,0.7856) (1389.5,0.7855) (1403.5,0.7831) (1417.5,0.7773) (1431.5,0.7674) (1445.5,0.7526) (1459.5,0.7321) (1473.5,0.7055) (1487.5,0.6736) (1501.5,0.6374) (1515.5,0.5977) (1529.5,0.5555) (1543.5,0.5116) (1557.5,0.4668) (1571.5,0.4218) (1585.5,0.3774) (1599.5,0.3343) (1613.5,0.2925) (1627.5,0.2520) (1641.5,0.2126) (1653.5,0.1795)};
\end{axis}
\end{tikzpicture}%
 &
  \pgfplotsset{every axis/.append style={yticklabels=\empty}}\providecommand{\rcpanelwidth}{\linewidth}%
\begin{tikzpicture}[baseline=(current axis.center), trim axis left, trim axis right]
\definecolor{rcvlearned}{HTML}{0000FF}\definecolor{rcvflat}{HTML}{00A000}\definecolor{rcvthin}{HTML}{FF0000}
\begin{axis}[
  scale only axis, width=\rcpanelwidth, height=0.74*\rcpanelwidth,
  grid=none, log ticks with fixed point, tick label style={font=\scriptsize}, label style={font=\small}, legend cell align=left, legend style={font=\scriptsize, draw=none, fill=none},
  axis x line*=bottom, axis y line*=left, axis line style={line width=0.4pt}, tick align=outside, tick style={black, line width=0.3pt},
  major tick length=2pt, scaled ticks=false, /pgf/number format/1000 sep={}, enlargelimits=false,
  xmin=-1044, xmax=1044, xtick={-500,0,500}, ymin=0, ymax=1.08, ytick={0,0.5,1}]
\addplot[rcvthin, line width=1pt] coordinates {(-1044.5,0.1970) (-1034.5,0.2013) (-1024.5,0.2056) (-1014.5,0.2101) (-1004.5,0.2146) (-994.5,0.2192) (-984.5,0.2240) (-974.5,0.2288) (-964.5,0.2338) (-954.5,0.2388) (-944.5,0.2440) (-934.5,0.2493) (-924.5,0.2547) (-914.5,0.2602) (-904.5,0.2659) (-894.5,0.2716) (-884.5,0.2775) (-874.5,0.2835) (-864.5,0.2896) (-854.5,0.2959) (-844.5,0.3022) (-834.5,0.3088) (-824.5,0.3154) (-814.5,0.3222) (-804.5,0.3291) (-794.5,0.3361) (-784.5,0.3433) (-774.5,0.3506) (-764.5,0.3581) (-754.5,0.3657) (-744.5,0.3734) (-734.5,0.3813) (-724.5,0.3893) (-714.5,0.3974) (-704.5,0.4057) (-694.5,0.4142) (-684.5,0.4227) (-674.5,0.4315) (-664.5,0.4403) (-654.5,0.4493) (-644.5,0.4584) (-634.5,0.4677) (-624.5,0.4770) (-614.5,0.4866) (-604.5,0.4962) (-594.5,0.5060) (-584.5,0.5159) (-574.5,0.5259) (-564.5,0.5360) (-554.5,0.5462) (-544.5,0.5565) (-534.5,0.5670) (-524.5,0.5775) (-514.5,0.5881) (-504.5,0.5988) (-494.5,0.6096) (-484.5,0.6204) (-474.5,0.6313) (-464.5,0.6423) (-454.5,0.6533) (-444.5,0.6643) (-434.5,0.6754) (-424.5,0.6865) (-414.5,0.6976) (-404.5,0.7087) (-394.5,0.7198) (-384.5,0.7308) (-374.5,0.7418) (-364.5,0.7528) (-354.5,0.7637) (-344.5,0.7745) (-334.5,0.7853) (-324.5,0.7959) (-314.5,0.8064) (-304.5,0.8168) (-294.5,0.8271) (-284.5,0.8372) (-274.5,0.8471) (-264.5,0.8568) (-254.5,0.8663) (-244.5,0.8757) (-234.5,0.8847) (-224.5,0.8936) (-214.5,0.9022) (-204.5,0.9105) (-194.5,0.9185) (-184.5,0.9262) (-174.5,0.9336) (-164.5,0.9406) (-154.5,0.9474) (-144.5,0.9537) (-134.5,0.9597) (-124.5,0.9653) (-114.5,0.9706) (-104.5,0.9754) (-94.5,0.9798) (-84.5,0.9838) (-74.5,0.9874) (-64.5,0.9905) (-54.5,0.9932) (-44.5,0.9955) (-34.5,0.9973) (-24.5,0.9986) (-14.5,0.9995) (-4.5,1.0000) (5.5,0.9999) (15.5,0.9994) (25.5,0.9985) (35.5,0.9971) (45.5,0.9953) (55.5,0.9930) (65.5,0.9902) (75.5,0.9870) (85.5,0.9834) (95.5,0.9794) (105.5,0.9749) (115.5,0.9701) (125.5,0.9648) (135.5,0.9591) (145.5,0.9531) (155.5,0.9467) (165.5,0.9399) (175.5,0.9328) (185.5,0.9254) (195.5,0.9177) (205.5,0.9096) (215.5,0.9013) (225.5,0.8927) (235.5,0.8838) (245.5,0.8747) (255.5,0.8654) (265.5,0.8558) (275.5,0.8461) (285.5,0.8362) (295.5,0.8260) (305.5,0.8158) (315.5,0.8054) (325.5,0.7948) (335.5,0.7842) (345.5,0.7734) (355.5,0.7626) (365.5,0.7517) (375.5,0.7407) (385.5,0.7297) (395.5,0.7187) (405.5,0.7076) (415.5,0.6965) (425.5,0.6854) (435.5,0.6743) (445.5,0.6632) (455.5,0.6522) (465.5,0.6412) (475.5,0.6303) (485.5,0.6194) (495.5,0.6085) (505.5,0.5978) (515.5,0.5871) (525.5,0.5765) (535.5,0.5659) (545.5,0.5555) (555.5,0.5452) (565.5,0.5350) (575.5,0.5249) (585.5,0.5149) (595.5,0.5050) (605.5,0.4952) (615.5,0.4856) (625.5,0.4761) (635.5,0.4667) (645.5,0.4575) (655.5,0.4484) (665.5,0.4394) (675.5,0.4306) (685.5,0.4219) (695.5,0.4133) (705.5,0.4049) (715.5,0.3966) (725.5,0.3885) (735.5,0.3805) (745.5,0.3726) (755.5,0.3649) (765.5,0.3573) (775.5,0.3499) (785.5,0.3426) (795.5,0.3354) (805.5,0.3284) (815.5,0.3215) (825.5,0.3147) (835.5,0.3081) (845.5,0.3016) (855.5,0.2952) (865.5,0.2890) (875.5,0.2829) (885.5,0.2769) (895.5,0.2710) (905.5,0.2653) (915.5,0.2597) (925.5,0.2542) (935.5,0.2488) (945.5,0.2435) (955.5,0.2383) (965.5,0.2333) (975.5,0.2283) (985.5,0.2235) (995.5,0.2188) (1005.5,0.2141) (1015.5,0.2096) (1025.5,0.2052) (1035.5,0.2009) (1043.5,0.1975)};
\addplot[rcvflat, line width=1pt] coordinates {(-1044.5,0.4078) (-1034.5,0.4079) (-1024.5,0.4120) (-1014.5,0.4223) (-1004.5,0.4275) (-994.5,0.4360) (-984.5,0.4435) (-974.5,0.4548) (-964.5,0.4631) (-954.5,0.4713) (-944.5,0.4798) (-934.5,0.4893) (-924.5,0.4976) (-914.5,0.5067) (-904.5,0.5150) (-894.5,0.5235) (-884.5,0.5324) (-874.5,0.5418) (-864.5,0.5478) (-854.5,0.5571) (-844.5,0.5644) (-834.5,0.5701) (-824.5,0.5788) (-814.5,0.5875) (-804.5,0.5936) (-794.5,0.6011) (-784.5,0.6096) (-774.5,0.6155) (-764.5,0.6246) (-754.5,0.6316) (-744.5,0.6398) (-734.5,0.6455) (-724.5,0.6548) (-714.5,0.6581) (-704.5,0.6677) (-694.5,0.6741) (-684.5,0.6844) (-674.5,0.6913) (-664.5,0.6969) (-654.5,0.7021) (-644.5,0.7100) (-634.5,0.7182) (-624.5,0.7248) (-614.5,0.7297) (-604.5,0.7376) (-594.5,0.7458) (-584.5,0.7491) (-574.5,0.7583) (-564.5,0.7660) (-554.5,0.7731) (-544.5,0.7783) (-534.5,0.7817) (-524.5,0.7915) (-514.5,0.7982) (-504.5,0.8073) (-494.5,0.8135) (-484.5,0.8157) (-474.5,0.8247) (-464.5,0.8320) (-454.5,0.8390) (-444.5,0.8452) (-434.5,0.8516) (-424.5,0.8568) (-414.5,0.8628) (-404.5,0.8692) (-394.5,0.8777) (-384.5,0.8818) (-374.5,0.8901) (-364.5,0.8940) (-354.5,0.9017) (-344.5,0.9066) (-334.5,0.9153) (-324.5,0.9211) (-314.5,0.9259) (-304.5,0.9310) (-294.5,0.9396) (-284.5,0.9447) (-274.5,0.9510) (-264.5,0.9532) (-254.5,0.9591) (-244.5,0.9669) (-234.5,0.9699) (-224.5,0.9752) (-214.5,0.9773) (-204.5,0.9803) (-194.5,0.9872) (-184.5,0.9871) (-174.5,0.9884) (-164.5,0.9917) (-154.5,0.9927) (-144.5,0.9948) (-134.5,0.9935) (-124.5,0.9958) (-114.5,0.9946) (-104.5,0.9955) (-94.5,0.9972) (-84.5,0.9966) (-74.5,0.9979) (-64.5,0.9975) (-54.5,1.0008) (-44.5,0.9992) (-34.5,1.0006) (-24.5,0.9995) (-14.5,0.9992) (-4.5,1.0001) (5.5,1.0011) (15.5,1.0034) (25.5,1.0029) (35.5,0.9996) (45.5,0.9999) (55.5,1.0010) (65.5,0.9997) (75.5,0.9990) (85.5,1.0000) (95.5,0.9969) (105.5,0.9990) (115.5,0.9997) (125.5,0.9969) (135.5,0.9980) (145.5,0.9989) (155.5,0.9975) (165.5,0.9964) (175.5,0.9951) (185.5,0.9954) (195.5,0.9918) (205.5,0.9888) (215.5,0.9889) (225.5,0.9839) (235.5,0.9788) (245.5,0.9761) (255.5,0.9725) (265.5,0.9645) (275.5,0.9597) (285.5,0.9571) (295.5,0.9522) (305.5,0.9429) (315.5,0.9358) (325.5,0.9334) (335.5,0.9274) (345.5,0.9224) (355.5,0.9157) (365.5,0.9070) (375.5,0.8999) (385.5,0.8966) (395.5,0.8903) (405.5,0.8833) (415.5,0.8771) (425.5,0.8733) (435.5,0.8646) (445.5,0.8576) (455.5,0.8516) (465.5,0.8470) (475.5,0.8378) (485.5,0.8343) (495.5,0.8295) (505.5,0.8200) (515.5,0.8119) (525.5,0.8063) (535.5,0.8016) (545.5,0.7931) (555.5,0.7845) (565.5,0.7788) (575.5,0.7727) (585.5,0.7682) (595.5,0.7612) (605.5,0.7523) (615.5,0.7467) (625.5,0.7385) (635.5,0.7325) (645.5,0.7242) (655.5,0.7181) (665.5,0.7126) (675.5,0.7037) (685.5,0.6971) (695.5,0.6901) (705.5,0.6823) (715.5,0.6746) (725.5,0.6663) (735.5,0.6617) (745.5,0.6558) (755.5,0.6469) (765.5,0.6388) (775.5,0.6299) (785.5,0.6259) (795.5,0.6130) (805.5,0.6070) (815.5,0.6021) (825.5,0.5910) (835.5,0.5852) (845.5,0.5761) (855.5,0.5692) (865.5,0.5602) (875.5,0.5527) (885.5,0.5422) (895.5,0.5377) (905.5,0.5295) (915.5,0.5203) (925.5,0.5104) (935.5,0.5022) (945.5,0.4923) (955.5,0.4835) (965.5,0.4758) (975.5,0.4664) (985.5,0.4550) (995.5,0.4463) (1005.5,0.4377) (1015.5,0.4291) (1025.5,0.4200) (1035.5,0.4118) (1043.5,0.4099)};
\addplot[rcvlearned, line width=1pt, dash pattern=on 3pt off 1.5pt] coordinates {(-1044.5,0.3841) (-1034.5,0.4225) (-1024.5,0.4569) (-1014.5,0.4864) (-1004.5,0.5102) (-994.5,0.5276) (-984.5,0.5395) (-974.5,0.5467) (-964.5,0.5505) (-954.5,0.5518) (-944.5,0.5520) (-934.5,0.5523) (-924.5,0.5540) (-914.5,0.5585) (-904.5,0.5660) (-894.5,0.5760) (-884.5,0.5877) (-874.5,0.6008) (-864.5,0.6144) (-854.5,0.6280) (-844.5,0.6408) (-834.5,0.6520) (-824.5,0.6612) (-814.5,0.6688) (-804.5,0.6749) (-794.5,0.6798) (-784.5,0.6841) (-774.5,0.6880) (-764.5,0.6919) (-754.5,0.6962) (-744.5,0.7014) (-734.5,0.7073) (-724.5,0.7139) (-714.5,0.7209) (-704.5,0.7282) (-694.5,0.7357) (-684.5,0.7430) (-674.5,0.7500) (-664.5,0.7565) (-654.5,0.7623) (-644.5,0.7677) (-634.5,0.7727) (-624.5,0.7773) (-614.5,0.7817) (-604.5,0.7861) (-594.5,0.7905) (-584.5,0.7951) (-574.5,0.8000) (-564.5,0.8051) (-554.5,0.8104) (-544.5,0.8159) (-534.5,0.8214) (-524.5,0.8270) (-514.5,0.8325) (-504.5,0.8380) (-494.5,0.8432) (-484.5,0.8483) (-474.5,0.8533) (-464.5,0.8582) (-454.5,0.8631) (-444.5,0.8681) (-434.5,0.8732) (-424.5,0.8784) (-414.5,0.8840) (-404.5,0.8898) (-394.5,0.8959) (-384.5,0.9022) (-374.5,0.9087) (-364.5,0.9154) (-354.5,0.9221) (-344.5,0.9289) (-334.5,0.9357) (-324.5,0.9424) (-314.5,0.9490) (-304.5,0.9554) (-294.5,0.9617) (-284.5,0.9678) (-274.5,0.9737) (-264.5,0.9793) (-254.5,0.9846) (-244.5,0.9895) (-234.5,0.9940) (-224.5,0.9980) (-214.5,1.0015) (-204.5,1.0044) (-194.5,1.0067) (-184.5,1.0082) (-174.5,1.0090) (-164.5,1.0088) (-154.5,1.0078) (-144.5,1.0061) (-134.5,1.0039) (-124.5,1.0012) (-114.5,0.9985) (-104.5,0.9959) (-94.5,0.9936) (-84.5,0.9921) (-74.5,0.9915) (-64.5,0.9917) (-54.5,0.9927) (-44.5,0.9941) (-34.5,0.9959) (-24.5,0.9976) (-14.5,0.9990) (-4.5,0.9999) (5.5,0.9998) (15.5,0.9989) (25.5,0.9974) (35.5,0.9957) (45.5,0.9940) (55.5,0.9926) (65.5,0.9917) (75.5,0.9915) (85.5,0.9922) (95.5,0.9938) (105.5,0.9961) (115.5,0.9988) (125.5,1.0015) (135.5,1.0041) (145.5,1.0063) (155.5,1.0080) (165.5,1.0089) (175.5,1.0089) (185.5,1.0081) (195.5,1.0065) (205.5,1.0042) (215.5,1.0012) (225.5,0.9976) (235.5,0.9936) (245.5,0.9890) (255.5,0.9841) (265.5,0.9788) (275.5,0.9731) (285.5,0.9672) (295.5,0.9611) (305.5,0.9548) (315.5,0.9483) (325.5,0.9417) (335.5,0.9350) (345.5,0.9282) (355.5,0.9214) (365.5,0.9147) (375.5,0.9081) (385.5,0.9016) (395.5,0.8953) (405.5,0.8892) (415.5,0.8834) (425.5,0.8779) (435.5,0.8726) (445.5,0.8676) (455.5,0.8626) (465.5,0.8577) (475.5,0.8528) (485.5,0.8478) (495.5,0.8427) (505.5,0.8374) (515.5,0.8320) (525.5,0.8264) (535.5,0.8209) (545.5,0.8153) (555.5,0.8099) (565.5,0.8046) (575.5,0.7995) (585.5,0.7946) (595.5,0.7901) (605.5,0.7857) (615.5,0.7813) (625.5,0.7769) (635.5,0.7722) (645.5,0.7672) (655.5,0.7618) (665.5,0.7558) (675.5,0.7493) (685.5,0.7423) (695.5,0.7349) (705.5,0.7275) (715.5,0.7202) (725.5,0.7132) (735.5,0.7067) (745.5,0.7008) (755.5,0.6957) (765.5,0.6915) (775.5,0.6876) (785.5,0.6837) (795.5,0.6794) (805.5,0.6743) (815.5,0.6681) (825.5,0.6604) (835.5,0.6510) (845.5,0.6396) (855.5,0.6267) (865.5,0.6131) (875.5,0.5994) (885.5,0.5865) (895.5,0.5749) (905.5,0.5652) (915.5,0.5580) (925.5,0.5537) (935.5,0.5522) (945.5,0.5520) (955.5,0.5517) (965.5,0.5502) (975.5,0.5462) (985.5,0.5385) (995.5,0.5261) (1005.5,0.5081) (1015.5,0.4837) (1025.5,0.4536) (1035.5,0.4188) (1043.5,0.3881)};
\end{axis}
\end{tikzpicture}%
 &
  \pgfplotsset{every axis/.append style={yticklabels=\empty}}\providecommand{\rcpanelwidth}{\linewidth}%
\begin{tikzpicture}[baseline=(current axis.center), trim axis left, trim axis right]
\definecolor{rcvlearned}{HTML}{0000FF}\definecolor{rcvflat}{HTML}{00A000}\definecolor{rcvthin}{HTML}{FF0000}
\begin{axis}[
  scale only axis, width=\rcpanelwidth, height=0.74*\rcpanelwidth,
  grid=none, log ticks with fixed point, tick label style={font=\scriptsize}, label style={font=\small}, legend cell align=left, legend style={font=\scriptsize, draw=none, fill=none},
  axis x line*=bottom, axis y line*=left, axis line style={line width=0.4pt}, tick align=outside, tick style={black, line width=0.3pt},
  major tick length=2pt, scaled ticks=false, /pgf/number format/1000 sep={}, enlargelimits=false,
  xmin=-2542, xmax=2534, xtick={-2000,0,2000}, ymin=0, ymax=1.08, ytick={0,0.5,1}]
\addplot[rcvthin, line width=1pt] coordinates {(-2542.5,0.2385) (-2522.5,0.2424) (-2502.5,0.2463) (-2482.5,0.2503) (-2462.5,0.2544) (-2442.5,0.2585) (-2422.5,0.2627) (-2402.5,0.2670) (-2382.5,0.2713) (-2362.5,0.2757) (-2342.5,0.2802) (-2322.5,0.2847) (-2302.5,0.2893) (-2282.5,0.2940) (-2262.5,0.2988) (-2242.5,0.3036) (-2222.5,0.3085) (-2202.5,0.3135) (-2182.5,0.3185) (-2162.5,0.3236) (-2142.5,0.3288) (-2122.5,0.3341) (-2102.5,0.3395) (-2082.5,0.3449) (-2062.5,0.3504) (-2042.5,0.3560) (-2022.5,0.3616) (-2002.5,0.3674) (-1982.5,0.3732) (-1962.5,0.3791) (-1942.5,0.3851) (-1922.5,0.3911) (-1902.5,0.3973) (-1882.5,0.4035) (-1862.5,0.4098) (-1842.5,0.4162) (-1822.5,0.4226) (-1802.5,0.4291) (-1782.5,0.4357) (-1762.5,0.4424) (-1742.5,0.4492) (-1722.5,0.4560) (-1702.5,0.4629) (-1682.5,0.4699) (-1662.5,0.4770) (-1642.5,0.4841) (-1622.5,0.4913) (-1602.5,0.4986) (-1582.5,0.5060) (-1562.5,0.5134) (-1542.5,0.5209) (-1522.5,0.5284) (-1502.5,0.5360) (-1482.5,0.5437) (-1462.5,0.5515) (-1442.5,0.5592) (-1422.5,0.5671) (-1402.5,0.5750) (-1382.5,0.5830) (-1362.5,0.5910) (-1342.5,0.5990) (-1322.5,0.6071) (-1302.5,0.6152) (-1282.5,0.6234) (-1262.5,0.6316) (-1242.5,0.6398) (-1222.5,0.6481) (-1202.5,0.6564) (-1182.5,0.6646) (-1162.5,0.6730) (-1142.5,0.6813) (-1122.5,0.6896) (-1102.5,0.6980) (-1082.5,0.7063) (-1062.5,0.7146) (-1042.5,0.7229) (-1022.5,0.7312) (-1002.5,0.7395) (-982.5,0.7478) (-962.5,0.7560) (-942.5,0.7642) (-922.5,0.7723) (-902.5,0.7804) (-882.5,0.7884) (-862.5,0.7964) (-842.5,0.8043) (-822.5,0.8122) (-802.5,0.8199) (-782.5,0.8276) (-762.5,0.8352) (-742.5,0.8427) (-722.5,0.8501) (-702.5,0.8574) (-682.5,0.8646) (-662.5,0.8716) (-642.5,0.8785) (-622.5,0.8853) (-602.5,0.8920) (-582.5,0.8985) (-562.5,0.9048) (-542.5,0.9110) (-522.5,0.9171) (-502.5,0.9229) (-482.5,0.9286) (-462.5,0.9341) (-442.5,0.9394) (-422.5,0.9446) (-402.5,0.9495) (-382.5,0.9542) (-362.5,0.9587) (-342.5,0.9630) (-322.5,0.9671) (-302.5,0.9710) (-282.5,0.9746) (-262.5,0.9780) (-242.5,0.9812) (-222.5,0.9841) (-202.5,0.9868) (-182.5,0.9893) (-162.5,0.9915) (-142.5,0.9935) (-122.5,0.9952) (-102.5,0.9966) (-82.5,0.9978) (-62.5,0.9987) (-42.5,0.9994) (-22.5,0.9998) (-2.5,1.0000) (17.5,0.9999) (37.5,0.9995) (57.5,0.9989) (77.5,0.9981) (97.5,0.9969) (117.5,0.9955) (137.5,0.9939) (157.5,0.9920) (177.5,0.9899) (197.5,0.9875) (217.5,0.9848) (237.5,0.9820) (257.5,0.9789) (277.5,0.9755) (297.5,0.9719) (317.5,0.9681) (337.5,0.9641) (357.5,0.9598) (377.5,0.9554) (397.5,0.9507) (417.5,0.9458) (437.5,0.9407) (457.5,0.9355) (477.5,0.9300) (497.5,0.9244) (517.5,0.9185) (537.5,0.9126) (557.5,0.9064) (577.5,0.9001) (597.5,0.8936) (617.5,0.8870) (637.5,0.8802) (657.5,0.8733) (677.5,0.8663) (697.5,0.8592) (717.5,0.8519) (737.5,0.8446) (757.5,0.8371) (777.5,0.8295) (797.5,0.8219) (817.5,0.8141) (837.5,0.8063) (857.5,0.7984) (877.5,0.7904) (897.5,0.7824) (917.5,0.7743) (937.5,0.7662) (957.5,0.7580) (977.5,0.7498) (997.5,0.7416) (1017.5,0.7333) (1037.5,0.7250) (1057.5,0.7167) (1077.5,0.7084) (1097.5,0.7000) (1117.5,0.6917) (1137.5,0.6834) (1157.5,0.6750) (1177.5,0.6667) (1197.5,0.6584) (1217.5,0.6501) (1237.5,0.6419) (1257.5,0.6336) (1277.5,0.6254) (1297.5,0.6173) (1317.5,0.6091) (1337.5,0.6010) (1357.5,0.5930) (1377.5,0.5849) (1397.5,0.5770) (1417.5,0.5691) (1437.5,0.5612) (1457.5,0.5534) (1477.5,0.5456) (1497.5,0.5380) (1517.5,0.5303) (1537.5,0.5228) (1557.5,0.5153) (1577.5,0.5078) (1597.5,0.5005) (1617.5,0.4932) (1637.5,0.4859) (1657.5,0.4788) (1677.5,0.4717) (1697.5,0.4647) (1717.5,0.4578) (1737.5,0.4509) (1757.5,0.4441) (1777.5,0.4374) (1797.5,0.4308) (1817.5,0.4242) (1837.5,0.4178) (1857.5,0.4114) (1877.5,0.4051) (1897.5,0.3988) (1917.5,0.3927) (1937.5,0.3866) (1957.5,0.3806) (1977.5,0.3747) (1997.5,0.3688) (2017.5,0.3631) (2037.5,0.3574) (2057.5,0.3518) (2077.5,0.3463) (2097.5,0.3408) (2117.5,0.3354) (2137.5,0.3301) (2157.5,0.3249) (2177.5,0.3198) (2197.5,0.3147) (2217.5,0.3097) (2237.5,0.3048) (2257.5,0.3000) (2277.5,0.2952) (2297.5,0.2905) (2317.5,0.2859) (2337.5,0.2813) (2357.5,0.2768) (2377.5,0.2724) (2397.5,0.2680) (2417.5,0.2638) (2437.5,0.2596) (2457.5,0.2554) (2477.5,0.2513) (2497.5,0.2473) (2517.5,0.2434) (2533.5,0.2403)};
\addplot[rcvflat, line width=1pt] coordinates {(-2542.5,0.1488) (-2522.5,0.1515) (-2502.5,0.1556) (-2482.5,0.1603) (-2462.5,0.1647) (-2442.5,0.1692) (-2422.5,0.1727) (-2402.5,0.1771) (-2382.5,0.1814) (-2362.5,0.1861) (-2342.5,0.1908) (-2322.5,0.1957) (-2302.5,0.1993) (-2282.5,0.2040) (-2262.5,0.2098) (-2242.5,0.2149) (-2222.5,0.2207) (-2202.5,0.2245) (-2182.5,0.2312) (-2162.5,0.2360) (-2142.5,0.2426) (-2122.5,0.2477) (-2102.5,0.2533) (-2082.5,0.2586) (-2062.5,0.2656) (-2042.5,0.2710) (-2022.5,0.2762) (-2002.5,0.2831) (-1982.5,0.2879) (-1962.5,0.2947) (-1942.5,0.3020) (-1922.5,0.3071) (-1902.5,0.3126) (-1882.5,0.3198) (-1862.5,0.3258) (-1842.5,0.3329) (-1822.5,0.3379) (-1802.5,0.3467) (-1782.5,0.3531) (-1762.5,0.3608) (-1742.5,0.3673) (-1722.5,0.3735) (-1702.5,0.3802) (-1682.5,0.3881) (-1662.5,0.3946) (-1642.5,0.4024) (-1622.5,0.4103) (-1602.5,0.4181) (-1582.5,0.4251) (-1562.5,0.4325) (-1542.5,0.4391) (-1522.5,0.4483) (-1502.5,0.4563) (-1482.5,0.4635) (-1462.5,0.4712) (-1442.5,0.4786) (-1422.5,0.4861) (-1402.5,0.4951) (-1382.5,0.5036) (-1362.5,0.5104) (-1342.5,0.5190) (-1322.5,0.5257) (-1302.5,0.5333) (-1282.5,0.5402) (-1262.5,0.5495) (-1242.5,0.5564) (-1222.5,0.5659) (-1202.5,0.5727) (-1182.5,0.5814) (-1162.5,0.5889) (-1142.5,0.5972) (-1122.5,0.6076) (-1102.5,0.6132) (-1082.5,0.6229) (-1062.5,0.6292) (-1042.5,0.6373) (-1022.5,0.6484) (-1002.5,0.6553) (-982.5,0.6652) (-962.5,0.6743) (-942.5,0.6824) (-922.5,0.6878) (-902.5,0.6970) (-882.5,0.7056) (-862.5,0.7153) (-842.5,0.7217) (-822.5,0.7306) (-802.5,0.7389) (-782.5,0.7471) (-762.5,0.7535) (-742.5,0.7633) (-722.5,0.7714) (-702.5,0.7770) (-682.5,0.7852) (-662.5,0.7916) (-642.5,0.8013) (-622.5,0.8085) (-602.5,0.8179) (-582.5,0.8252) (-562.5,0.8306) (-542.5,0.8392) (-522.5,0.8469) (-502.5,0.8530) (-482.5,0.8623) (-462.5,0.8701) (-442.5,0.8773) (-422.5,0.8851) (-402.5,0.8904) (-382.5,0.8984) (-362.5,0.9046) (-342.5,0.9133) (-322.5,0.9177) (-302.5,0.9282) (-282.5,0.9332) (-262.5,0.9412) (-242.5,0.9474) (-222.5,0.9557) (-202.5,0.9590) (-182.5,0.9642) (-162.5,0.9718) (-142.5,0.9754) (-122.5,0.9843) (-102.5,0.9892) (-82.5,0.9941) (-62.5,0.9985) (-42.5,0.9993) (-22.5,1.0011) (-2.5,1.0039) (17.5,1.0011) (37.5,0.9992) (57.5,0.9951) (77.5,0.9926) (97.5,0.9875) (117.5,0.9842) (137.5,0.9802) (157.5,0.9745) (177.5,0.9708) (197.5,0.9630) (217.5,0.9617) (237.5,0.9555) (257.5,0.9505) (277.5,0.9467) (297.5,0.9421) (317.5,0.9344) (337.5,0.9299) (357.5,0.9273) (377.5,0.9184) (397.5,0.9109) (417.5,0.9065) (437.5,0.9008) (457.5,0.8927) (477.5,0.8884) (497.5,0.8804) (517.5,0.8753) (537.5,0.8704) (557.5,0.8621) (577.5,0.8563) (597.5,0.8512) (617.5,0.8433) (637.5,0.8375) (657.5,0.8309) (677.5,0.8212) (697.5,0.8162) (717.5,0.8078) (737.5,0.8013) (757.5,0.7942) (777.5,0.7863) (797.5,0.7795) (817.5,0.7709) (837.5,0.7630) (857.5,0.7580) (877.5,0.7497) (897.5,0.7415) (917.5,0.7360) (937.5,0.7274) (957.5,0.7212) (977.5,0.7133) (997.5,0.7061) (1017.5,0.6979) (1037.5,0.6895) (1057.5,0.6820) (1077.5,0.6741) (1097.5,0.6684) (1117.5,0.6582) (1137.5,0.6509) (1157.5,0.6443) (1177.5,0.6354) (1197.5,0.6300) (1217.5,0.6224) (1237.5,0.6142) (1257.5,0.6038) (1277.5,0.5977) (1297.5,0.5877) (1317.5,0.5814) (1337.5,0.5733) (1357.5,0.5666) (1377.5,0.5575) (1397.5,0.5514) (1417.5,0.5444) (1437.5,0.5356) (1457.5,0.5279) (1477.5,0.5203) (1497.5,0.5136) (1517.5,0.5047) (1537.5,0.4987) (1557.5,0.4904) (1577.5,0.4838) (1597.5,0.4759) (1617.5,0.4685) (1637.5,0.4612) (1657.5,0.4547) (1677.5,0.4462) (1697.5,0.4396) (1717.5,0.4324) (1737.5,0.4238) (1757.5,0.4190) (1777.5,0.4114) (1797.5,0.4047) (1817.5,0.3971) (1837.5,0.3907) (1857.5,0.3825) (1877.5,0.3769) (1897.5,0.3704) (1917.5,0.3642) (1937.5,0.3591) (1957.5,0.3502) (1977.5,0.3446) (1997.5,0.3384) (2017.5,0.3323) (2037.5,0.3264) (2057.5,0.3192) (2077.5,0.3130) (2097.5,0.3077) (2117.5,0.3019) (2137.5,0.2960) (2157.5,0.2911) (2177.5,0.2847) (2197.5,0.2801) (2217.5,0.2743) (2237.5,0.2683) (2257.5,0.2623) (2277.5,0.2575) (2297.5,0.2525) (2317.5,0.2461) (2337.5,0.2419) (2357.5,0.2366) (2377.5,0.2317) (2397.5,0.2268) (2417.5,0.2227) (2437.5,0.2180) (2457.5,0.2133) (2477.5,0.2093) (2497.5,0.2039) (2517.5,0.1998) (2533.5,0.1968)};
\addplot[rcvlearned, line width=1pt, dash pattern=on 3pt off 1.5pt] coordinates {(-2542.5,0.1860) (-2522.5,0.1906) (-2502.5,0.1952) (-2482.5,0.1999) (-2462.5,0.2045) (-2442.5,0.2091) (-2422.5,0.2138) (-2402.5,0.2184) (-2382.5,0.2230) (-2362.5,0.2276) (-2342.5,0.2323) (-2322.5,0.2369) (-2302.5,0.2416) (-2282.5,0.2464) (-2262.5,0.2511) (-2242.5,0.2560) (-2222.5,0.2609) (-2202.5,0.2660) (-2182.5,0.2711) (-2162.5,0.2764) (-2142.5,0.2818) (-2122.5,0.2872) (-2102.5,0.2928) (-2082.5,0.2985) (-2062.5,0.3044) (-2042.5,0.3103) (-2022.5,0.3162) (-2002.5,0.3223) (-1982.5,0.3285) (-1962.5,0.3347) (-1942.5,0.3410) (-1922.5,0.3473) (-1902.5,0.3538) (-1882.5,0.3602) (-1862.5,0.3668) (-1842.5,0.3734) (-1822.5,0.3801) (-1802.5,0.3869) (-1782.5,0.3937) (-1762.5,0.4005) (-1742.5,0.4075) (-1722.5,0.4144) (-1702.5,0.4215) (-1682.5,0.4285) (-1662.5,0.4357) (-1642.5,0.4429) (-1622.5,0.4501) (-1602.5,0.4573) (-1582.5,0.4646) (-1562.5,0.4720) (-1542.5,0.4794) (-1522.5,0.4868) (-1502.5,0.4942) (-1482.5,0.5017) (-1462.5,0.5091) (-1442.5,0.5166) (-1422.5,0.5241) (-1402.5,0.5316) (-1382.5,0.5391) (-1362.5,0.5466) (-1342.5,0.5541) (-1322.5,0.5616) (-1302.5,0.5691) (-1282.5,0.5766) (-1262.5,0.5842) (-1242.5,0.5918) (-1222.5,0.5994) (-1202.5,0.6071) (-1182.5,0.6148) (-1162.5,0.6227) (-1142.5,0.6305) (-1122.5,0.6384) (-1102.5,0.6464) (-1082.5,0.6544) (-1062.5,0.6623) (-1042.5,0.6704) (-1022.5,0.6784) (-1002.5,0.6863) (-982.5,0.6943) (-962.5,0.7022) (-942.5,0.7101) (-922.5,0.7179) (-902.5,0.7258) (-882.5,0.7335) (-862.5,0.7412) (-842.5,0.7489) (-822.5,0.7565) (-802.5,0.7640) (-782.5,0.7715) (-762.5,0.7789) (-742.5,0.7862) (-722.5,0.7935) (-702.5,0.8007) (-682.5,0.8079) (-662.5,0.8149) (-642.5,0.8220) (-622.5,0.8289) (-602.5,0.8359) (-582.5,0.8427) (-562.5,0.8495) (-542.5,0.8563) (-522.5,0.8630) (-502.5,0.8697) (-482.5,0.8763) (-462.5,0.8829) (-442.5,0.8894) (-422.5,0.8959) (-402.5,0.9024) (-382.5,0.9088) (-362.5,0.9152) (-342.5,0.9215) (-322.5,0.9278) (-302.5,0.9340) (-282.5,0.9401) (-262.5,0.9461) (-242.5,0.9521) (-222.5,0.9579) (-202.5,0.9636) (-182.5,0.9691) (-162.5,0.9744) (-142.5,0.9795) (-122.5,0.9842) (-102.5,0.9884) (-82.5,0.9922) (-62.5,0.9953) (-42.5,0.9977) (-22.5,0.9993) (-2.5,1.0000) (17.5,0.9996) (37.5,0.9982) (57.5,0.9960) (77.5,0.9930) (97.5,0.9894) (117.5,0.9853) (137.5,0.9807) (157.5,0.9757) (177.5,0.9704) (197.5,0.9649) (217.5,0.9593) (237.5,0.9535) (257.5,0.9476) (277.5,0.9416) (297.5,0.9355) (317.5,0.9293) (337.5,0.9231) (357.5,0.9167) (377.5,0.9104) (397.5,0.9040) (417.5,0.8975) (437.5,0.8910) (457.5,0.8845) (477.5,0.8780) (497.5,0.8713) (517.5,0.8647) (537.5,0.8580) (557.5,0.8512) (577.5,0.8444) (597.5,0.8376) (617.5,0.8307) (637.5,0.8237) (657.5,0.8167) (677.5,0.8096) (697.5,0.8025) (717.5,0.7953) (737.5,0.7880) (757.5,0.7807) (777.5,0.7733) (797.5,0.7659) (817.5,0.7583) (837.5,0.7508) (857.5,0.7431) (877.5,0.7354) (897.5,0.7277) (917.5,0.7199) (937.5,0.7121) (957.5,0.7042) (977.5,0.6963) (997.5,0.6883) (1017.5,0.6803) (1037.5,0.6724) (1057.5,0.6643) (1077.5,0.6563) (1097.5,0.6484) (1117.5,0.6404) (1137.5,0.6325) (1157.5,0.6246) (1177.5,0.6168) (1197.5,0.6090) (1217.5,0.6013) (1237.5,0.5937) (1257.5,0.5861) (1277.5,0.5785) (1297.5,0.5710) (1317.5,0.5635) (1337.5,0.5560) (1357.5,0.5485) (1377.5,0.5410) (1397.5,0.5335) (1417.5,0.5260) (1437.5,0.5185) (1457.5,0.5110) (1477.5,0.5035) (1497.5,0.4961) (1517.5,0.4886) (1537.5,0.4812) (1557.5,0.4738) (1577.5,0.4665) (1597.5,0.4592) (1617.5,0.4519) (1637.5,0.4447) (1657.5,0.4375) (1677.5,0.4303) (1697.5,0.4232) (1717.5,0.4162) (1737.5,0.4092) (1757.5,0.4023) (1777.5,0.3954) (1797.5,0.3886) (1817.5,0.3818) (1837.5,0.3751) (1857.5,0.3685) (1877.5,0.3619) (1897.5,0.3554) (1917.5,0.3489) (1937.5,0.3426) (1957.5,0.3362) (1977.5,0.3300) (1997.5,0.3238) (2017.5,0.3177) (2037.5,0.3117) (2057.5,0.3058) (2077.5,0.3000) (2097.5,0.2943) (2117.5,0.2886) (2137.5,0.2831) (2157.5,0.2777) (2177.5,0.2724) (2197.5,0.2672) (2217.5,0.2622) (2237.5,0.2572) (2257.5,0.2524) (2277.5,0.2475) (2297.5,0.2428) (2317.5,0.2381) (2337.5,0.2334) (2357.5,0.2288) (2377.5,0.2242) (2397.5,0.2195) (2417.5,0.2149) (2437.5,0.2103) (2457.5,0.2057) (2477.5,0.2010) (2497.5,0.1964) (2517.5,0.1918) (2533.5,0.1881)};
\end{axis}
\end{tikzpicture}%
 \\\noalign{\vskip 1pt}
  & \multicolumn{3}{@{}c@{\hspace{4pt}}}{position along the diagonal [px]} \\\noalign{\vskip 5pt}
  \multicolumn{4}{@{}c@{}}{\resizebox{\ifdim\width>\linewidth\linewidth\else\width\fi}{!}{\mbox{\tikz[baseline=-0.6ex]{\draw[rcvflat, line width=1pt] (0,0) -- (12pt,0);}\,flat field}\enspace \mbox{\tikz[baseline=-0.6ex]{\draw[rcvlearned, line width=1pt, dash pattern=on 3pt off 1.5pt] (0,0) -- (12pt,0);}\,learned (ours)}\enspace \mbox{\tikz[baseline=-0.6ex]{\draw[rcvthin, line width=1pt] (0,0) -- (12pt,0);}\,thin lens model ($\cos^4\alpha$)}}} \\
\end{tabular}%
\endgroup

        \caption{Normalized vignetting along the direction of the sensor diagonal through the principal point: the $\cos^4$ falloff of the thin lens and our learned vignetting against the measured flat field.}
        \label{fig:vignetting}
\end{figure}
\begin{figure*}[t]
        \centering
        \begin{minipage}[c]{0.43\textwidth}
                \centering
                \providecommand{\rcfigdir}{fig}%
\begingroup
\footnotesize\setlength{\tabcolsep}{0pt}%
\begin{tabular}{@{\hspace{22pt}}c@{\hspace{34pt}}c@{\hspace{20pt}}}
  \makebox[0pt]{Fisheye lens, $f/4$} & \makebox[0pt]{Tamron 20\,mm lens, $f/8$} \\\noalign{\vskip 3pt}
  \def\rcpanelwidth{\dimexpr(\linewidth-76pt)*563/1000\relax}\input{\rcfigdir/focus-surfaces/fisheye} &
  \def\rcpanelwidth{\dimexpr(\linewidth-76pt)*437/1000\relax}\input{\rcfigdir/focus-surfaces/tamron} \\\noalign{\vskip 3pt}
  \makebox[0pt]{{\scriptsize 0.52\,m\,\tikz[baseline=0.5pt]{\fill[fill={rgb,255:red,230;green,239;blue,249}] (0pt,0) rectangle (2.05pt,4pt);\fill[fill={rgb,255:red,222;green,234;blue,246}] (2pt,0) rectangle (4.05pt,4pt);\fill[fill={rgb,255:red,214;green,228;blue,243}] (4pt,0) rectangle (6.05pt,4pt);\fill[fill={rgb,255:red,206;green,222;blue,240}] (6pt,0) rectangle (8.05pt,4pt);\fill[fill={rgb,255:red,198;green,217;blue,237}] (8pt,0) rectangle (10.05pt,4pt);\fill[fill={rgb,255:red,190;green,211;blue,234}] (10pt,0) rectangle (12.05pt,4pt);\fill[fill={rgb,255:red,182;green,206;blue,231}] (12pt,0) rectangle (14.05pt,4pt);\fill[fill={rgb,255:red,174;green,200;blue,228}] (14pt,0) rectangle (16.05pt,4pt);\fill[fill={rgb,255:red,166;green,195;blue,224}] (16pt,0) rectangle (18.05pt,4pt);\fill[fill={rgb,255:red,158;green,189;blue,221}] (18pt,0) rectangle (20.05pt,4pt);\fill[fill={rgb,255:red,150;green,183;blue,218}] (20pt,0) rectangle (22.05pt,4pt);\fill[fill={rgb,255:red,142;green,178;blue,215}] (22pt,0) rectangle (24.05pt,4pt);\fill[fill={rgb,255:red,135;green,172;blue,212}] (24pt,0) rectangle (26.05pt,4pt);\fill[fill={rgb,255:red,127;green,167;blue,209}] (26pt,0) rectangle (28.05pt,4pt);\fill[fill={rgb,255:red,119;green,161;blue,206}] (28pt,0) rectangle (30.05pt,4pt);\fill[fill={rgb,255:red,111;green,155;blue,203}] (30pt,0) rectangle (32.05pt,4pt);\fill[fill={rgb,255:red,103;green,150;blue,199}] (32pt,0) rectangle (34.05pt,4pt);\fill[fill={rgb,255:red,95;green,144;blue,196}] (34pt,0) rectangle (36.05pt,4pt);\fill[fill={rgb,255:red,87;green,139;blue,193}] (36pt,0) rectangle (38.05pt,4pt);\fill[fill={rgb,255:red,79;green,133;blue,190}] (38pt,0) rectangle (40.05pt,4pt);\fill[fill={rgb,255:red,71;green,128;blue,187}] (40pt,0) rectangle (42.05pt,4pt);\fill[fill={rgb,255:red,63;green,122;blue,184}] (42pt,0) rectangle (44.05pt,4pt);\fill[fill={rgb,255:red,55;green,116;blue,181}] (44pt,0) rectangle (46.05pt,4pt);\fill[fill={rgb,255:red,47;green,111;blue,178}] (46pt,0) rectangle (48.05pt,4pt);\draw[line width=0.2pt] (0,0) rectangle (48pt,4pt);}\,0.76\,m}} & \makebox[0pt]{{\scriptsize 0.22\,m\,\tikz[baseline=0.5pt]{\fill[fill={rgb,255:red,230;green,239;blue,249}] (0pt,0) rectangle (2.05pt,4pt);\fill[fill={rgb,255:red,222;green,234;blue,246}] (2pt,0) rectangle (4.05pt,4pt);\fill[fill={rgb,255:red,214;green,228;blue,243}] (4pt,0) rectangle (6.05pt,4pt);\fill[fill={rgb,255:red,206;green,222;blue,240}] (6pt,0) rectangle (8.05pt,4pt);\fill[fill={rgb,255:red,198;green,217;blue,237}] (8pt,0) rectangle (10.05pt,4pt);\fill[fill={rgb,255:red,190;green,211;blue,234}] (10pt,0) rectangle (12.05pt,4pt);\fill[fill={rgb,255:red,182;green,206;blue,231}] (12pt,0) rectangle (14.05pt,4pt);\fill[fill={rgb,255:red,174;green,200;blue,228}] (14pt,0) rectangle (16.05pt,4pt);\fill[fill={rgb,255:red,166;green,195;blue,224}] (16pt,0) rectangle (18.05pt,4pt);\fill[fill={rgb,255:red,158;green,189;blue,221}] (18pt,0) rectangle (20.05pt,4pt);\fill[fill={rgb,255:red,150;green,183;blue,218}] (20pt,0) rectangle (22.05pt,4pt);\fill[fill={rgb,255:red,142;green,178;blue,215}] (22pt,0) rectangle (24.05pt,4pt);\fill[fill={rgb,255:red,135;green,172;blue,212}] (24pt,0) rectangle (26.05pt,4pt);\fill[fill={rgb,255:red,127;green,167;blue,209}] (26pt,0) rectangle (28.05pt,4pt);\fill[fill={rgb,255:red,119;green,161;blue,206}] (28pt,0) rectangle (30.05pt,4pt);\fill[fill={rgb,255:red,111;green,155;blue,203}] (30pt,0) rectangle (32.05pt,4pt);\fill[fill={rgb,255:red,103;green,150;blue,199}] (32pt,0) rectangle (34.05pt,4pt);\fill[fill={rgb,255:red,95;green,144;blue,196}] (34pt,0) rectangle (36.05pt,4pt);\fill[fill={rgb,255:red,87;green,139;blue,193}] (36pt,0) rectangle (38.05pt,4pt);\fill[fill={rgb,255:red,79;green,133;blue,190}] (38pt,0) rectangle (40.05pt,4pt);\fill[fill={rgb,255:red,71;green,128;blue,187}] (40pt,0) rectangle (42.05pt,4pt);\fill[fill={rgb,255:red,63;green,122;blue,184}] (42pt,0) rectangle (44.05pt,4pt);\fill[fill={rgb,255:red,55;green,116;blue,181}] (44pt,0) rectangle (46.05pt,4pt);\fill[fill={rgb,255:red,47;green,111;blue,178}] (46pt,0) rectangle (48.05pt,4pt);\draw[line width=0.2pt] (0,0) rectangle (48pt,4pt);}\,0.31\,m}} \\\noalign{\vskip 1pt}
  \multicolumn{2}{@{\hspace{22pt}}c@{\hspace{20pt}}}{\scriptsize focus distance from the camera centre} \\
\end{tabular}%
\endgroup

        \end{minipage}\hfill
        \begin{minipage}[c]{0.54\textwidth}
                \centering
                \providecommand{\rcfigdir}{fig}%
\begingroup
\ifdefined\rcpanelwd\else\newlength\rcpanelwd\fi
\setlength{\rcpanelwd}{\dimexpr\linewidth-48pt\relax}\def\rcpanelwidth{\rcpanelwd}%
\footnotesize\setlength{\tabcolsep}{0pt}%
\begin{tabular}{@{}c@{\hspace{4pt}}c@{}}
  \tikz[baseline=(n.center)]{\node[inner sep=0pt, outer sep=0pt, minimum width=44pt, minimum height=0.13078\rcpanelwd, align=center, font=\fontsize{6.5}{7.5}\selectfont] (n) {ground\\truth};\draw[overlay, black!60, line width=0.4pt] (n.south west) -- (n.south east);} &
  \providecommand{\rcfigdir}{fig}%
\providecommand{\rcpanelwidth}{\linewidth}%
\begin{tikzpicture}[baseline=(current axis.center)]
\begin{axis}[
  scale only axis, width=\rcpanelwidth, height=0.13078*\rcpanelwidth,
  grid=none, log ticks with fixed point, tick label style={font=\scriptsize}, label style={font=\small}, legend cell align=left, legend style={font=\scriptsize, draw=none, fill=none},
  xmin=0, xmax=627, ymin=0, ymax=82, enlargelimits=false,
  xtick=\empty, ytick=\empty, axis on top, axis line style={draw=none}]
\addplot graphics[xmin=0, xmax=627, ymin=0, ymax=82] {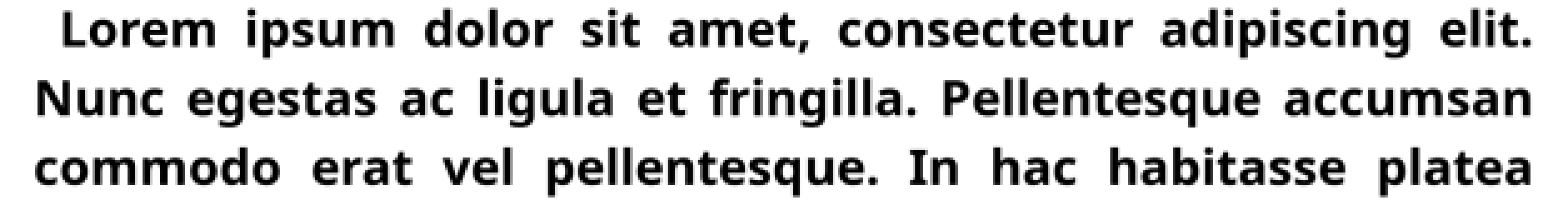};
\end{axis}
\end{tikzpicture}%
 \\
  \tikz[baseline=(n.center)]{\node[inner sep=0pt, outer sep=0pt, minimum width=44pt, minimum height=0.13264\rcpanelwd, align=center, font=\fontsize{6.5}{7.5}\selectfont] (n) {training\\image\\39 of 63};\draw[overlay, black!60, line width=0.4pt] (n.south west) -- (n.south east);} &
  \providecommand{\rcfigdir}{fig}%
\providecommand{\rcpanelwidth}{\linewidth}%
\begin{tikzpicture}[baseline=(current axis.center)]
\begin{axis}[
  scale only axis, width=\rcpanelwidth, height=0.13264*\rcpanelwidth,
  grid=none, log ticks with fixed point, tick label style={font=\scriptsize}, label style={font=\small}, legend cell align=left, legend style={font=\scriptsize, draw=none, fill=none},
  xmin=0, xmax=573, ymin=0, ymax=76, enlargelimits=false,
  xtick=\empty, ytick=\empty, axis on top, axis line style={draw=none}]
\addplot graphics[xmin=0, xmax=573, ymin=0, ymax=76] {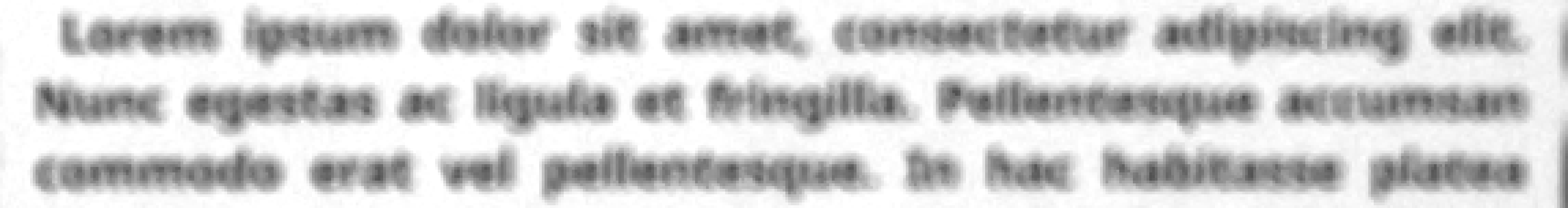};
\end{axis}
\end{tikzpicture}%
 \\
  \tikz[baseline=(n.center)]{\node[inner sep=0pt, outer sep=0pt, minimum width=44pt, minimum height=0.13078\rcpanelwd, align=center, font=\fontsize{6.5}{7.5}\selectfont] (n) {pinhole};\draw[overlay, black!60, line width=0.4pt] (n.south west) -- (n.south east);} &
  \providecommand{\rcfigdir}{fig}%
\providecommand{\rcpanelwidth}{\linewidth}%
\begin{tikzpicture}[baseline=(current axis.center)]
\begin{axis}[
  scale only axis, width=\rcpanelwidth, height=0.13078*\rcpanelwidth,
  grid=none, log ticks with fixed point, tick label style={font=\scriptsize}, label style={font=\small}, legend cell align=left, legend style={font=\scriptsize, draw=none, fill=none},
  xmin=0, xmax=627, ymin=0, ymax=82, enlargelimits=false,
  xtick=\empty, ytick=\empty, axis on top, axis line style={draw=none}]
\addplot graphics[xmin=0, xmax=627, ymin=0, ymax=82] {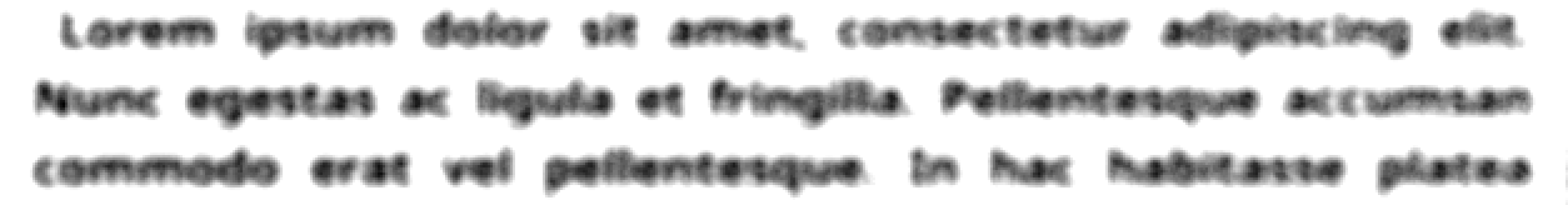};
\end{axis}
\end{tikzpicture}%
 \\
  \tikz[baseline=(n.center)]{\node[inner sep=0pt, outer sep=0pt, minimum width=44pt, minimum height=0.13078\rcpanelwd, align=center, font=\fontsize{6.5}{7.5}\selectfont] (n) {ours};} &
  \providecommand{\rcfigdir}{fig}%
\providecommand{\rcpanelwidth}{\linewidth}%
\begin{tikzpicture}[baseline=(current axis.center)]
\begin{axis}[
  scale only axis, width=\rcpanelwidth, height=0.13078*\rcpanelwidth,
  grid=none, log ticks with fixed point, tick label style={font=\scriptsize}, label style={font=\small}, legend cell align=left, legend style={font=\scriptsize, draw=none, fill=none},
  xmin=0, xmax=627, ymin=0, ymax=82, enlargelimits=false,
  xtick=\empty, ytick=\empty, axis on top, axis line style={draw=none}]
\addplot graphics[xmin=0, xmax=627, ymin=0, ymax=82] {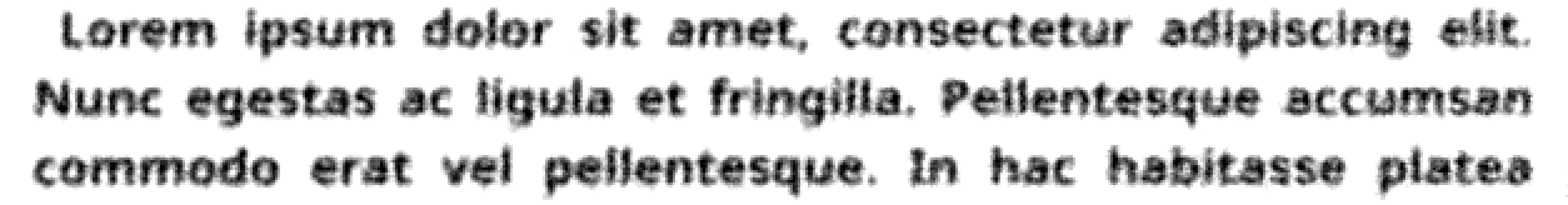};
\end{axis}
\end{tikzpicture}%
 \\
\end{tabular}%
\endgroup

        \end{minipage}
        \caption{The focus model. \emph{Left:} the learned focus surfaces, a hemisphere for the fisheye at $f/4$ and a plane for the Tamron 20\,mm at $f/8$. \emph{Right:} a texture estimated from defocused images. The radiometric camera recovers sharp text, whereas the pinhole camera has to put the blur of the images into the texture.}
        \label{fig:focus-surfaces}
        \label{fig:texture-learned}
\end{figure*}
We use the three lenses of \cref{sec:geometric-accuracy} and the
spline model with Siemens-star captures at a large aperture ($f/4$; $f/2.8$ for the
Tamron) for stronger vignetting, and one flat-field image per lens
(\cref{tab:supp-dataset}): the lens, with its hood on, covered by a
white acrylic plate and lit by diffuse daylight through a light
diffuser. \Cref{fig:vignetting} compares, along the
direction of the sensor diagonal through the principal point, the
learned vignetting, a B-spline (\cref{sec:vignetting}),
with the flat field, read off the green photosites, and with the
$\cos^4$ falloff of the thin lens. All profiles are normalized at the
principal point. The learned vignetting follows the flat field for all
three lenses: its RMS deviation along this line, in units of the
value at the principal point, is $0.009$ for the 14\,mm lens, $0.042$ for the
fisheye and $0.025$ for the Tamron, against $0.244$, $0.199$ and
$0.059$ for the $\cos^4$ falloff.

\subsection{Defocus}
\label{sec:experiment-defocus}
We validate the focus model in three ways. First,
\cref{fig:focus-surfaces} (left) shows the focus surfaces learned with the
spline model for the Tamron 20\,mm at $f/8$ and for the fisheye at
$f/4$: a plane for the wide-angle lens and a hemisphere for the
fisheye, as expected for these lens types
\cite{Kingslake2010,wide-field-of-view-2017}.

Second, we solve a first multi-view radiometric problem with the
calibrated camera: we estimate the unknown reflectance of a surface
from many defocused views of it by inverse rendering. With a pinhole
aperture, the estimated texture has to explain the defocus of the
images and is blurred. We calibrate the Sigma 14\,mm at $f/2.8$ on
Siemens stars, once with the aperture and the focus surface and once
as a pinhole, and estimate the texture of a board with printed text
from 63 defocused images with each camera. \Cref{fig:texture-learned} (right)
compares both textures with the printed design: the radiometric
camera recovers sharp text where the pinhole camera returns the blur
of the images. Since the calibration works with relative irradiance,
a texture is recovered only up to a scale and an offset, which we fit
once per texture before comparing it with the design. With the design
ranging from 0 to 1, the RMS difference to it is $0.177$ for the
radiometric camera and $0.235$ for the pinhole, and it is lower for
the radiometric camera on 99\,\% of the page.

Third, we ablate the focus model on two lenses with visible defocus,
the Tamron 20\,mm at $f/8$ with the spline model and a 24--70\,mm zoom
lens at 48\,mm and $f/10$ with Brown--Conrady, by refining the same
geometric start once with the aperture and the focus
surface and once as a pinhole. \Cref{tab:defocus} shows that modeling
the aperture and the focus surface lowers the median reprojection
error on the Thorlabs target by 25\,\% and 16\,\% compared with the
pinhole refinement, both significantly; on the Tamron, the pinhole
refinement is even worse than the geometric start.
\begin{table}[t]
        \centering
        \footnotesize
\setlength{\tabcolsep}{1.5pt}
\begin{tabular*}{\linewidth}{@{\extracolsep{\fill}}lll@{}}
\toprule
\textbf{Refinement} & \textbf{Median [px]} & \textbf{95th perc. [px]} \\
\midrule
\multicolumn{3}{@{}l}{\underline{\textit{Tamron 20\,mm, $f/8$, central spline}}} \\
Geometric start & 0.160 & 0.499 \\
Pinhole & \textit{0.199} & \textit{0.534} \\
Aperture and focus & \textbf{0.149} & \textbf{0.410} \\
\quad $-$ pinhole & $-0.051$ {\scriptsize[$-0.054$,\,$-0.047$]} & $-0.124$ {\scriptsize[$-0.139$,\,$-0.103$]} \\
\midrule
\multicolumn{3}{@{}l}{\underline{\textit{Zoom at 48\,mm, $f/10$, Brown--Conrady}}} \\
Geometric start & 0.068 & 0.194 \\
Pinhole & \textbf{0.050} & \textbf{0.150} \\
Aperture and focus & \textbf{0.042} & \textbf{0.117} \\
\quad $-$ pinhole & $-0.008$ {\scriptsize[$-0.013$,\,$-0.003$]} & $-0.032$ {\scriptsize[$-0.047$,\,$-0.017$]} \\
\bottomrule
\end{tabular*}

        \caption{Defocus ablation on the held-out target: the refinement as a pinhole and with the aperture and the focus surface, and their difference with its 95\,\% bootstrap interval. Bold and italic as in \cref{tab:refinement-chrome}.}
        \label{tab:defocus}
\end{table}

\section{Conclusion}
We presented the radiometric camera, which lifts any central geometric
camera model to a radiometric one with a finite aperture, a free-form
focus surface and a generic vignetting function. We also presented its
calibration from the RAW photosites of ordinary calibration images,
formulated without parameter-dependent discontinuities or integration
domains and with every gauge freedom fixed exactly. To our knowledge,
it is the first model to unite aperture, focus and vignetting with
any central projection and to calibrate all of them jointly from the
images of a geometric calibration (\cref{tab:related}). The calibrated
cameras have a lower reprojection error on a held-out high-precision
target than their geometric starts for most models and lenses, their
vignetting reproduces a measured flat field, and their focus surface
is learned as a plane for a wide-angle lens and as a hemisphere for a
fisheye. In a first multi-view radiometric reconstruction, they
recover the in-focus reflectance of a texture from defocused images.
The estimation of light sources and of the attenuation of water
is future work. We will
release the code, the data and the configurations of every experiment.

{
    \small
    \bibliographystyle{ieeenat_fullname}
    \bibliography{main}
}

\clearpage
\setcounter{page}{1}
\maketitlesupplementary
\setcounter{section}{0}\renewcommand{\thesection}{A.\arabic{section}}
\setcounter{table}{0}\renewcommand{\thetable}{A\arabic{table}}
\setcounter{figure}{0}\renewcommand{\thefigure}{A\arabic{figure}}
\setcounter{equation}{0}\renewcommand{\theequation}{A\arabic{equation}}
\renewcommand{\theHsection}{A.\arabic{section}}\renewcommand{\theHtable}{A\arabic{table}}%
\renewcommand{\theHfigure}{A\arabic{figure}}\renewcommand{\theHequation}{A\arabic{equation}}%

\section{Board Warping}
\label{sec:board-warping}
\begin{figure*}[t]
        \centering
        \providecommand{\rcfigdir}{fig}%
\begingroup
\ifdefined\rcpanelwd\else\newlength\rcpanelwd\fi
\ifdefined\rcdiffwd\else\newlength\rcdiffwd\fi
\setlength{\rcpanelwd}{\dimexpr(\linewidth-49pt)*100/420\relax}\setlength{\rcdiffwd}{0.4\rcpanelwd}\def\rcpanelwidth{\rcpanelwd}%
\footnotesize\setlength{\tabcolsep}{0pt}\renewcommand{\arraystretch}{1}%
\newcommand{\rcdiffpanel}[1]{{\def\rcpanelwidth{\rcdiffwd}\input{\rcfigdir/board-warping-lenses-offsets-difference-b6/#1}}}%
\newcommand{\rcmaphead}[1]{\llap{\smash{\raisebox{\dimexpr0.5\rcdiffwd+3pt\relax}{\makebox[\rcdiffwd]{\tiny\begin{tabular}[b]{@{}c@{}}#1\end{tabular}}}}}}%
\newcommand{\rcmapstat}[1]{\llap{\raisebox{\dimexpr-0.5\rcdiffwd-1pt-\height\relax}{\makebox[\rcdiffwd]{\tiny\begin{tabular}[t]{@{}c@{}}#1\end{tabular}}}}}%
\begin{tabular}{@{}p{9pt}@{\hspace{3pt}}c@{\hspace{4pt}}c@{\hspace{4pt}}c@{\hspace{10pt}}c@{\hspace{8pt}}c@{\hspace{8pt}}c@{\hspace{3pt}}}
  & \makebox[0pt]{\scriptsize Sigma 14\,mm lens} & \makebox[0pt]{\scriptsize Tamron 20\,mm lens} & \makebox[0pt]{\scriptsize Fisheye lens} &  &  &  \\\noalign{\vskip 2pt}
  \makebox[9pt][c]{\rotatebox[origin=c]{90}{\scriptsize AprilGrid}} & \input{\rcfigdir/board-warping-lenses-offsets/bw_ag_14mm} & \input{\rcfigdir/board-warping-lenses-offsets/bw_ag_tamron} & \input{\rcfigdir/board-warping-lenses-offsets/bw_ag_fisheye} & {\def\rcpanelwidth{\rcdiffwd}\providecommand{\rcfigdir}{fig}%
\providecommand{\rcpanelwidth}{\linewidth}%
\begin{tikzpicture}[baseline=(current axis.center)]
\begin{axis}[
  scale only axis, width=\rcpanelwidth, height=1.00000*\rcpanelwidth,
  grid=none, log ticks with fixed point, tick label style={font=\scriptsize}, label style={font=\small}, legend cell align=left, legend style={font=\scriptsize, draw=none, fill=none},
  xmin=0, xmax=979.000, ymin=0, ymax=979.000, enlargelimits=false, clip=false,
  xtick=\empty, ytick=\empty, axis on top, axis line style={line width=0.2pt}]
\addplot graphics[xmin=0, xmax=979.000, ymin=0, ymax=979.000] {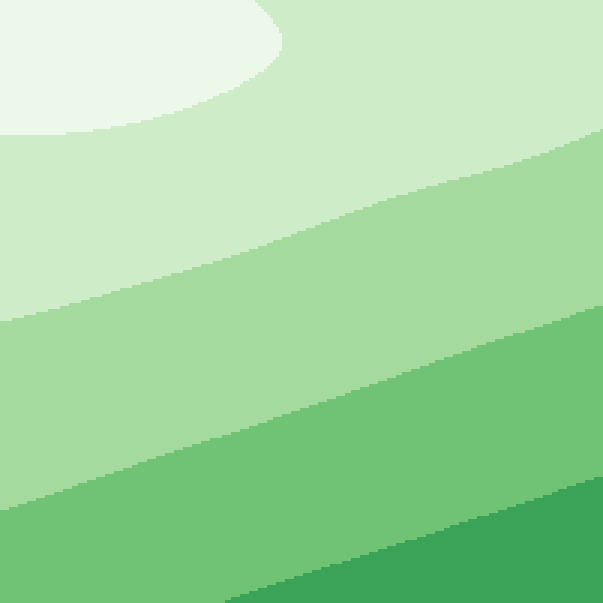};
\addplot[only marks, mark=*, mark size=0.45pt, black] coordinates {(0.00,979.00) (244.75,979.00) (489.50,979.00) (734.25,979.00) (979.00,979.00) (0.00,734.25) (244.75,734.25) (489.50,734.25) (734.25,734.25) (979.00,734.25) (0.00,489.50) (244.75,489.50) (489.50,489.50) (734.25,489.50) (979.00,489.50) (0.00,244.75) (244.75,244.75) (489.50,244.75) (734.25,244.75) (979.00,244.75) (0.00,0.00) (244.75,0.00) (489.50,0.00) (734.25,0.00) (979.00,0.00)};
\end{axis}
\pgfresetboundingbox\path (current axis.south west) rectangle (current axis.north east);   % the box is the page
\end{tikzpicture}%
}\rcmaphead{Sigma 14\,mm $-$\\Tamron 20\,mm}\rcmapstat{RMS 0.53\,mm\\max 0.95\,mm} & {\def\rcpanelwidth{\rcdiffwd}\providecommand{\rcfigdir}{fig}%
\providecommand{\rcpanelwidth}{\linewidth}%
\begin{tikzpicture}[baseline=(current axis.center)]
\begin{axis}[
  scale only axis, width=\rcpanelwidth, height=1.00000*\rcpanelwidth,
  grid=none, log ticks with fixed point, tick label style={font=\scriptsize}, label style={font=\small}, legend cell align=left, legend style={font=\scriptsize, draw=none, fill=none},
  xmin=0, xmax=979.000, ymin=0, ymax=979.000, enlargelimits=false, clip=false,
  xtick=\empty, ytick=\empty, axis on top, axis line style={line width=0.2pt}]
\addplot graphics[xmin=0, xmax=979.000, ymin=0, ymax=979.000] {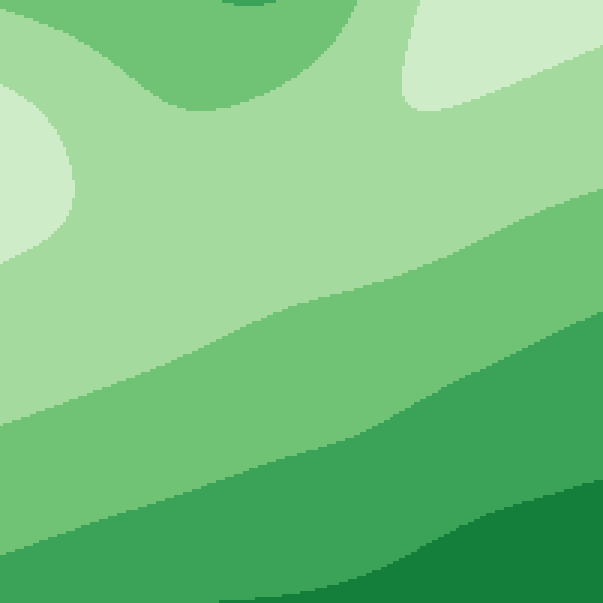};
\addplot[only marks, mark=*, mark size=0.45pt, black] coordinates {(0.00,979.00) (244.75,979.00) (489.50,979.00) (734.25,979.00) (979.00,979.00) (0.00,734.25) (244.75,734.25) (489.50,734.25) (734.25,734.25) (979.00,734.25) (0.00,489.50) (244.75,489.50) (489.50,489.50) (734.25,489.50) (979.00,489.50) (0.00,244.75) (244.75,244.75) (489.50,244.75) (734.25,244.75) (979.00,244.75) (0.00,0.00) (244.75,0.00) (489.50,0.00) (734.25,0.00) (979.00,0.00)};
\end{axis}
\pgfresetboundingbox\path (current axis.south west) rectangle (current axis.north east);   % the box is the page
\end{tikzpicture}%
}\rcmaphead{Fisheye $-$\\Tamron 20\,mm}\rcmapstat{RMS 0.69\,mm\\max 1.18\,mm} & {\def\rcpanelwidth{\rcdiffwd}\providecommand{\rcfigdir}{fig}%
\providecommand{\rcpanelwidth}{\linewidth}%
\begin{tikzpicture}[baseline=(current axis.center)]
\begin{axis}[
  scale only axis, width=\rcpanelwidth, height=1.00000*\rcpanelwidth,
  grid=none, log ticks with fixed point, tick label style={font=\scriptsize}, label style={font=\small}, legend cell align=left, legend style={font=\scriptsize, draw=none, fill=none},
  xmin=0, xmax=979.000, ymin=0, ymax=979.000, enlargelimits=false, clip=false,
  xtick=\empty, ytick=\empty, axis on top, axis line style={line width=0.2pt}]
\addplot graphics[xmin=0, xmax=979.000, ymin=0, ymax=979.000] {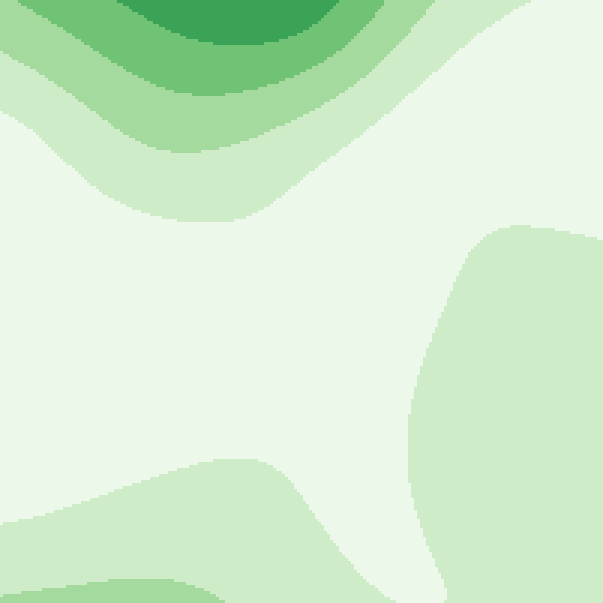};
\addplot[only marks, mark=*, mark size=0.45pt, black] coordinates {(0.00,979.00) (244.75,979.00) (489.50,979.00) (734.25,979.00) (979.00,979.00) (0.00,734.25) (244.75,734.25) (489.50,734.25) (734.25,734.25) (979.00,734.25) (0.00,489.50) (244.75,489.50) (489.50,489.50) (734.25,489.50) (979.00,489.50) (0.00,244.75) (244.75,244.75) (489.50,244.75) (734.25,244.75) (979.00,244.75) (0.00,0.00) (244.75,0.00) (489.50,0.00) (734.25,0.00) (979.00,0.00)};
\end{axis}
\pgfresetboundingbox\path (current axis.south west) rectangle (current axis.north east);   % the box is the page
\end{tikzpicture}%
}\rcmaphead{Sigma 14\,mm $-$\\Fisheye}\rcmapstat{RMS 0.30\,mm\\max 0.97\,mm} \\\noalign{\vskip 5pt}
  \makebox[9pt][c]{\rotatebox[origin=c]{90}{\scriptsize Siemens stars}} & \input{\rcfigdir/board-warping-lenses-offsets/bw_si_14mm} & \input{\rcfigdir/board-warping-lenses-offsets/bw_si_tamron} & \input{\rcfigdir/board-warping-lenses-offsets/bw_si_fisheye} & {\def\rcpanelwidth{\rcdiffwd}\providecommand{\rcfigdir}{fig}%
\providecommand{\rcpanelwidth}{\linewidth}%
\begin{tikzpicture}[baseline=(current axis.center)]
\begin{axis}[
  scale only axis, width=\rcpanelwidth, height=1.00000*\rcpanelwidth,
  grid=none, log ticks with fixed point, tick label style={font=\scriptsize}, label style={font=\small}, legend cell align=left, legend style={font=\scriptsize, draw=none, fill=none},
  xmin=0, xmax=1000.000, ymin=0, ymax=1000.000, enlargelimits=false, clip=false,
  xtick=\empty, ytick=\empty, axis on top, axis line style={line width=0.2pt}]
\addplot graphics[xmin=0, xmax=1000.000, ymin=0, ymax=1000.000] {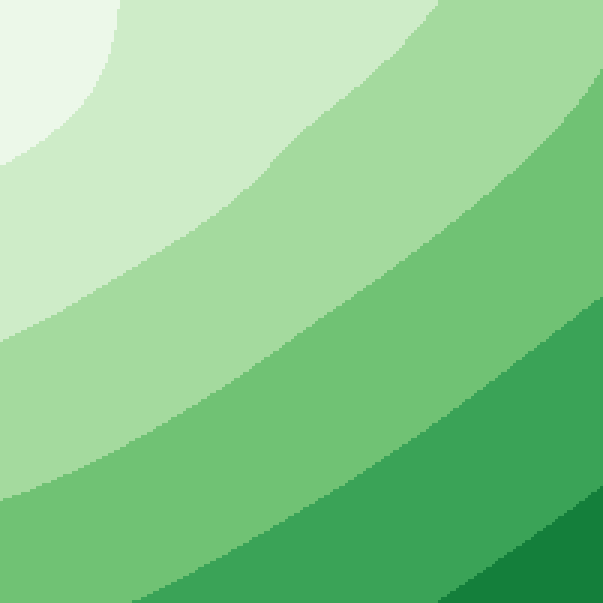};
\addplot[only marks, mark=*, mark size=0.45pt, black] coordinates {(0.00,1000.00) (250.00,1000.00) (500.00,1000.00) (750.00,1000.00) (1000.00,1000.00) (0.00,750.00) (250.00,750.00) (500.00,750.00) (750.00,750.00) (1000.00,750.00) (0.00,500.00) (250.00,500.00) (500.00,500.00) (750.00,500.00) (1000.00,500.00) (0.00,250.00) (250.00,250.00) (500.00,250.00) (750.00,250.00) (1000.00,250.00) (0.00,0.00) (250.00,0.00) (500.00,0.00) (750.00,0.00) (1000.00,0.00)};
\end{axis}
\pgfresetboundingbox\path (current axis.south west) rectangle (current axis.north east);   % the box is the page
\end{tikzpicture}%
}\rcmapstat{RMS 0.62\,mm\\max 1.13\,mm} & {\def\rcpanelwidth{\rcdiffwd}\providecommand{\rcfigdir}{fig}%
\providecommand{\rcpanelwidth}{\linewidth}%
\begin{tikzpicture}[baseline=(current axis.center)]
\begin{axis}[
  scale only axis, width=\rcpanelwidth, height=1.00000*\rcpanelwidth,
  grid=none, log ticks with fixed point, tick label style={font=\scriptsize}, label style={font=\small}, legend cell align=left, legend style={font=\scriptsize, draw=none, fill=none},
  xmin=0, xmax=1000.000, ymin=0, ymax=1000.000, enlargelimits=false, clip=false,
  xtick=\empty, ytick=\empty, axis on top, axis line style={line width=0.2pt}]
\addplot graphics[xmin=0, xmax=1000.000, ymin=0, ymax=1000.000] {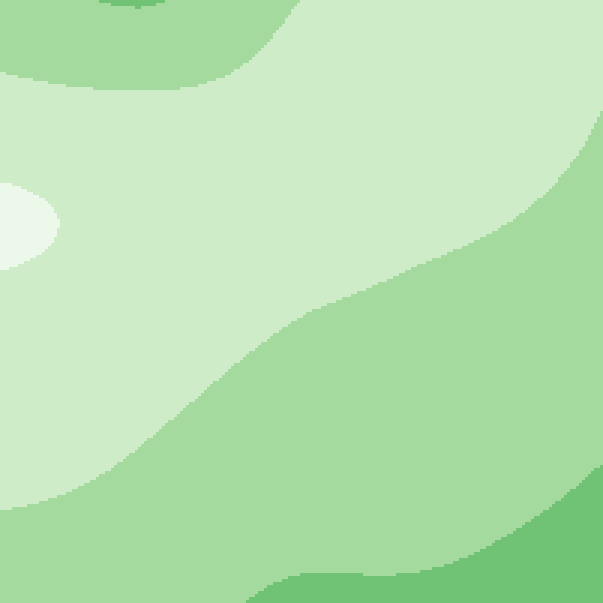};
\addplot[only marks, mark=*, mark size=0.45pt, black] coordinates {(0.00,1000.00) (250.00,1000.00) (500.00,1000.00) (750.00,1000.00) (1000.00,1000.00) (0.00,750.00) (250.00,750.00) (500.00,750.00) (750.00,750.00) (1000.00,750.00) (0.00,500.00) (250.00,500.00) (500.00,500.00) (750.00,500.00) (1000.00,500.00) (0.00,250.00) (250.00,250.00) (500.00,250.00) (750.00,250.00) (1000.00,250.00) (0.00,0.00) (250.00,0.00) (500.00,0.00) (750.00,0.00) (1000.00,0.00)};
\end{axis}
\pgfresetboundingbox\path (current axis.south west) rectangle (current axis.north east);   % the box is the page
\end{tikzpicture}%
}\rcmapstat{RMS 0.43\,mm\\max 0.73\,mm} & {\def\rcpanelwidth{\rcdiffwd}\providecommand{\rcfigdir}{fig}%
\providecommand{\rcpanelwidth}{\linewidth}%
\begin{tikzpicture}[baseline=(current axis.center)]
\begin{axis}[
  scale only axis, width=\rcpanelwidth, height=1.00000*\rcpanelwidth,
  grid=none, log ticks with fixed point, tick label style={font=\scriptsize}, label style={font=\small}, legend cell align=left, legend style={font=\scriptsize, draw=none, fill=none},
  xmin=0, xmax=1000.000, ymin=0, ymax=1000.000, enlargelimits=false, clip=false,
  xtick=\empty, ytick=\empty, axis on top, axis line style={line width=0.2pt}]
\addplot graphics[xmin=0, xmax=1000.000, ymin=0, ymax=1000.000] {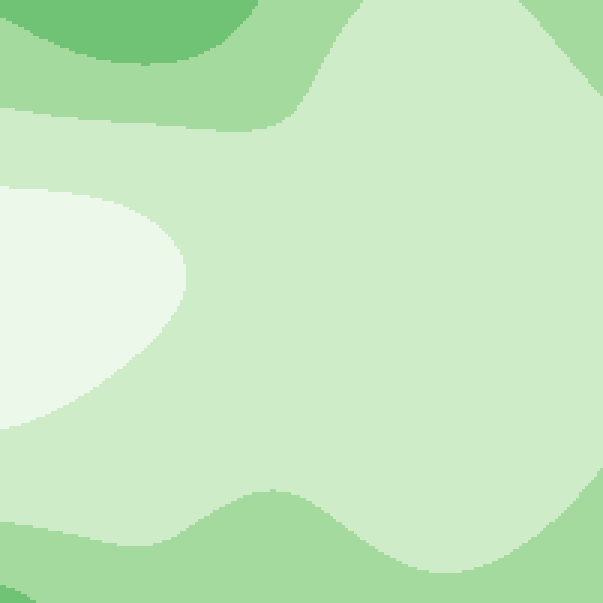};
\addplot[only marks, mark=*, mark size=0.45pt, black] coordinates {(0.00,1000.00) (250.00,1000.00) (500.00,1000.00) (750.00,1000.00) (1000.00,1000.00) (0.00,750.00) (250.00,750.00) (500.00,750.00) (750.00,750.00) (1000.00,750.00) (0.00,500.00) (250.00,500.00) (500.00,500.00) (750.00,500.00) (1000.00,500.00) (0.00,250.00) (250.00,250.00) (500.00,250.00) (750.00,250.00) (1000.00,250.00) (0.00,0.00) (250.00,0.00) (500.00,0.00) (750.00,0.00) (1000.00,0.00)};
\end{axis}
\pgfresetboundingbox\path (current axis.south west) rectangle (current axis.north east);   % the box is the page
\end{tikzpicture}%
}\rcmapstat{RMS 0.36\,mm\\max 0.79\,mm} \\\noalign{\vskip 6pt}
  \multicolumn{7}{@{}c@{}}{%
    \begin{minipage}[t]{0.49\linewidth}\centering\begin{tikzpicture}[baseline=(current bounding box.north)]
\fill[fill={rgb,255:red,17;green,70;blue,128}] (0*\dimexpr0.62\linewidth/12\relax,0) rectangle (1*\dimexpr0.62\linewidth/12\relax+0.05pt,5pt);
\fill[fill={rgb,255:red,42;green,113;blue,178}] (1*\dimexpr0.62\linewidth/12\relax,0) rectangle (2*\dimexpr0.62\linewidth/12\relax+0.05pt,5pt);
\fill[fill={rgb,255:red,74;green,151;blue,197}] (2*\dimexpr0.62\linewidth/12\relax,0) rectangle (3*\dimexpr0.62\linewidth/12\relax+0.05pt,5pt);
\fill[fill={rgb,255:red,139;green,193;blue,220}] (3*\dimexpr0.62\linewidth/12\relax,0) rectangle (4*\dimexpr0.62\linewidth/12\relax+0.05pt,5pt);
\fill[fill={rgb,255:red,193;green,221;blue,236}] (4*\dimexpr0.62\linewidth/12\relax,0) rectangle (5*\dimexpr0.62\linewidth/12\relax+0.05pt,5pt);
\fill[fill={rgb,255:red,231;green,240;blue,244}] (5*\dimexpr0.62\linewidth/12\relax,0) rectangle (6*\dimexpr0.62\linewidth/12\relax+0.05pt,5pt);
\fill[fill={rgb,255:red,250;green,235;blue,227}] (6*\dimexpr0.62\linewidth/12\relax,0) rectangle (7*\dimexpr0.62\linewidth/12\relax+0.05pt,5pt);
\fill[fill={rgb,255:red,251;green,206;blue,182}] (7*\dimexpr0.62\linewidth/12\relax,0) rectangle (8*\dimexpr0.62\linewidth/12\relax+0.05pt,5pt);
\fill[fill={rgb,255:red,241;green,159;blue,126}] (8*\dimexpr0.62\linewidth/12\relax,0) rectangle (9*\dimexpr0.62\linewidth/12\relax+0.05pt,5pt);
\fill[fill={rgb,255:red,217;green,102;blue,81}] (9*\dimexpr0.62\linewidth/12\relax,0) rectangle (10*\dimexpr0.62\linewidth/12\relax+0.05pt,5pt);
\fill[fill={rgb,255:red,187;green,42;blue,52}] (10*\dimexpr0.62\linewidth/12\relax,0) rectangle (11*\dimexpr0.62\linewidth/12\relax+0.05pt,5pt);
\fill[fill={rgb,255:red,134;green,10;blue,36}] (11*\dimexpr0.62\linewidth/12\relax,0) rectangle (12*\dimexpr0.62\linewidth/12\relax+0.05pt,5pt);
\draw[line width=0.2pt] (0,0) rectangle (12*\dimexpr0.62\linewidth/12\relax,5pt);
\draw[line width=0.2pt] (0*\dimexpr0.62\linewidth/12\relax,0) -- ++(0,-1.5pt) node[below, inner sep=1pt, font=\scriptsize] {${-}3$};
\draw[line width=0.2pt] (2*\dimexpr0.62\linewidth/12\relax,0) -- ++(0,-1.5pt) node[below, inner sep=1pt, font=\scriptsize] {${-}2$};
\draw[line width=0.2pt] (4*\dimexpr0.62\linewidth/12\relax,0) -- ++(0,-1.5pt) node[below, inner sep=1pt, font=\scriptsize] {${-}1$};
\draw[line width=0.2pt] (6*\dimexpr0.62\linewidth/12\relax,0) -- ++(0,-1.5pt) node[below, inner sep=1pt, font=\scriptsize] {$0$};
\draw[line width=0.2pt] (8*\dimexpr0.62\linewidth/12\relax,0) -- ++(0,-1.5pt) node[below, inner sep=1pt, font=\scriptsize] {${+}1$};
\draw[line width=0.2pt] (10*\dimexpr0.62\linewidth/12\relax,0) -- ++(0,-1.5pt) node[below, inner sep=1pt, font=\scriptsize] {${+}2$};
\draw[line width=0.2pt] (12*\dimexpr0.62\linewidth/12\relax,0) -- ++(0,-1.5pt) node[below, inner sep=1pt, font=\scriptsize] {${+}3$};
\node[right, inner sep=2pt, font=\scriptsize] at (12*\dimexpr0.62\linewidth/12\relax,2.5pt) {mm};
\end{tikzpicture}%
\\[1pt]{\scriptsize surfaces: the planar board plus the board offsets learned with this lens (in the plane and along the normal), drawn 50 times enlarged; colour: height above the board's plane (positive: towards the camera)}\end{minipage}\hfill
    \begin{minipage}[t]{0.49\linewidth}\centering\begin{tikzpicture}[baseline=(current bounding box.north)]
\fill[fill={rgb,255:red,236;green,248;blue,233}] (0*\dimexpr0.62\linewidth/6\relax,0) rectangle (1*\dimexpr0.62\linewidth/6\relax+0.05pt,5pt);
\fill[fill={rgb,255:red,206;green,236;blue,200}] (1*\dimexpr0.62\linewidth/6\relax,0) rectangle (2*\dimexpr0.62\linewidth/6\relax+0.05pt,5pt);
\fill[fill={rgb,255:red,164;green,218;blue,158}] (2*\dimexpr0.62\linewidth/6\relax,0) rectangle (3*\dimexpr0.62\linewidth/6\relax+0.05pt,5pt);
\fill[fill={rgb,255:red,112;green,194;blue,116}] (3*\dimexpr0.62\linewidth/6\relax,0) rectangle (4*\dimexpr0.62\linewidth/6\relax+0.05pt,5pt);
\fill[fill={rgb,255:red,58;green,163;blue,87}] (4*\dimexpr0.62\linewidth/6\relax,0) rectangle (5*\dimexpr0.62\linewidth/6\relax+0.05pt,5pt);
\fill[fill={rgb,255:red,20;green,127;blue,59}] (5*\dimexpr0.62\linewidth/6\relax,0) rectangle (6*\dimexpr0.62\linewidth/6\relax+0.05pt,5pt);
\draw[line width=0.2pt] (0,0) rectangle (6*\dimexpr0.62\linewidth/6\relax,5pt);
\draw[line width=0.2pt] (0*\dimexpr0.62\linewidth/6\relax,0) -- ++(0,-1.5pt) node[below, inner sep=1pt, font=\scriptsize] {$0$};
\draw[line width=0.2pt] (1*\dimexpr0.62\linewidth/6\relax,0) -- ++(0,-1.5pt) node[below, inner sep=1pt, font=\scriptsize] {$0.2$};
\draw[line width=0.2pt] (2*\dimexpr0.62\linewidth/6\relax,0) -- ++(0,-1.5pt) node[below, inner sep=1pt, font=\scriptsize] {$0.4$};
\draw[line width=0.2pt] (3*\dimexpr0.62\linewidth/6\relax,0) -- ++(0,-1.5pt) node[below, inner sep=1pt, font=\scriptsize] {$0.6$};
\draw[line width=0.2pt] (4*\dimexpr0.62\linewidth/6\relax,0) -- ++(0,-1.5pt) node[below, inner sep=1pt, font=\scriptsize] {$0.8$};
\draw[line width=0.2pt] (5*\dimexpr0.62\linewidth/6\relax,0) -- ++(0,-1.5pt) node[below, inner sep=1pt, font=\scriptsize] {$1$};
\draw[line width=0.2pt] (6*\dimexpr0.62\linewidth/6\relax,0) -- ++(0,-1.5pt) node[below, inner sep=1pt, font=\scriptsize] {$1.2$};
\node[right, inner sep=2pt, font=\scriptsize] at (6*\dimexpr0.62\linewidth/6\relax,2.5pt) {mm};
\end{tikzpicture}%
\\[1pt]{\scriptsize maps: length of the difference between the board offsets learned with two lenses (in the plane and along the normal); below: RMS and largest value}\end{minipage}} \\
\end{tabular}%
\endgroup

        \caption{\protect\input{fig/board-warping-caption}}
        \label{fig:board-warping}
\end{figure*}

In \cref{fig:board-warping}, we compare the learned warping of the $1\,\mathrm{m}\,\times\,1\,\mathrm{m}$ Siemens stars and AprilGrid board using the Tamron $20\,\mathrm{mm}$, Sigma $14\,\mathrm{mm}$ and fisheye lens.
The boards were not moved between the calibration of the three different lenses.
For each pair of lenses, we compare the two reconstructions point by point: every point of the board is displaced by the offset each calibration learned for it, in the plane (print error) and along the normal (height), and we take the difference of the two offsets.
\Cref{fig:board-warping} shows that the learned surfaces are similar; the RMS of this difference is below $0.7\,\mathrm{mm}$ for every pair of lenses on both boards.

\section{Spectral Formulation of the Image Formation}
\label{sec:supp-image-formation}
The camera sensor has a certain spectral sensitivity $S(\lambda)$~\cite{akkaynak2017space,
akkaynak2018revised, Darrodi:15, pharr2023physically,
toivonen2020practical} and measures the spectral radiant flux as~\cite{space-of-spectral-sensitivity-2013}
\begin{align*}
    S(\lambda)\Phi(\vbu,\lambda) = \int_{\mathcal{I}_{\vbu}}S(\lambda)E(\vbp,\lambda)\odif{\vbp},
\end{align*}
where
\begin{align*}
    E(\vbp,\lambda)
    =\int_{\calOmega(\vbp)} L(\vbx(\vbp,\vbomega), \vbomega_o(\vbp,\vbomega),\lambda) \cos\theta(\vbomega)\odif{\vbomega}.
\end{align*}
This spectral formulation requires knowledge of the camera's spectral sensitivity
$S(\lambda)$ which is difficult to recover in practice because it
requires specialized equipment~\cite{akkaynak2017space, Darrodi:15, toivonen2020practical}. Tri-color (RGB) renderers avoid this by working directly with per-channel quantities~\cite{Finlayson-PAMI-2017, rgb}.
Let $\Lambda_c$ be the set of wavelengths associated with the channel $c\in\{r,g,b\}$.
The per-channel radiant flux measured by the sensor in pixel $\vbu$ becomes
\begin{align*}
    &\Phi_c(\vbu) \coloneqq \int_{\mathcal{I}_{\vbu}}\underbrace{\int_{\Lambda_c}S_c(\lambda)E(\vbp,\lambda)\odif{\lambda}}_{\eqqcolon E_c(\vbp)}\odif{\vbp}\nonumber\\
    &=\int_{\mathcal{I}_{\vbu}}\int_{\calOmega(\vbp)}\underbrace{\int_{\Lambda_c}S_c(\lambda)L(\vbx(\vbp,\vbomega),\vbomega_o(\vbp,\vbomega),\lambda)\odif{\lambda}}_{\eqqcolon L_c(\vbx(\vbp,\vbomega),\vbomega_o(\vbp,\vbomega))}\nonumber\\
    &\qquad\qquad\qquad\qquad\qquad\qquad\qquad\cdot\cos\theta(\vbomega)\odif{\vbomega}\odif{\vbp}\nonumber,
\end{align*}
where Fubini's theorem~\cite{heuser2004} is used to swap the order of the integrals.
In the main text and in the following, all quantities are per color channel and the channel index $c$ is omitted.

\section{Geometric Camera Models and the Pinhole Aperture}
\label{sec:supp-pinhole-radiometry}
\label{sec:supp-camera-models}
The pinhole model can be constructed from the thin lens model by making the aperture radius infinitesimal so that for each sensor position only a single ray passes through the pinhole $\vbo\in\mathcal{A}$~\cite{pharr2023physically}.
We model the pinhole with the Dirac delta distribution $\delta$, which
satisfies the sifting property \cite{Bracewell1966TheFT}
\begin{align}
\label{eq:sifting}
    \int_{\calOmega(\vbp)} g(\vbomega)\,\delta(\vbomega-\vbomega_0)\odif{\vbomega} = g(\vbomega_0)
\end{align}
for any function $g$ that is continuous at $\vbomega_0$. The delta
selects the radiance along the single ray direction $\vbomega_0 \coloneqq \vbo - \vbp$ through the pinhole \cite{pinholedelta}, the radiant flux \cref{eq:per-channel-rad-flux} simplifies to
\begin{align}
    &\int_{\mathcal{I}_{\vbu}} \int_{\calOmega(\vbp)} \hspace{-0.3cm}L(\vbx(\vbp,\vbomega),\vbomega_o(\vbp,\vbomega))\cos\theta(\vbomega)\delta(\vbomega-\vbomega_0) \odif{\vbomega}\odif{\vbp}\nonumber\\
    \overset{(\ref{eq:sifting})}&{=} \int_{\mathcal{I}_{\vbu}}L(\vbx(\vbp,\vbomega_0),\vbomega_o(\vbp,\vbomega_0))\cos\theta(\vbomega_0)\odif{\vbp}\label{eq:pinhole-rad-flux} = \Phi(\vbu).
\end{align}
\cref{eq:pinhole-rad-flux} shows that the pinhole model is a pure geometric model and physically incorrect:
a vignetting falloff is missing and every scene point is in perfect focus.

The geometric camera models that the radiometric camera lifts are all
models of a pinhole aperture. The radiometric camera uses them through
their back-projection $\vbpi^{-1}$ alone
(\cref{eq:lifted-back-projection}), so it places no condition on how a
model is parameterized. The remainder of this section defines the five
models of \cref{sec:geometric-accuracy}, derives their
back-projections, gives the geometric meaning of their parameters
$\vbTheta_d$, and separates them into \emph{global} and \emph{local}
models. The distinction decides how a model is regularized and how the
vignetting, which is radially symmetric about the principal point
(\cref{sec:vignetting}), is attached to it.
\begin{center}
  \footnotesize\setlength{\tabcolsep}{3pt}
  \begin{tabular*}{\linewidth}{@{\extracolsep{\fill}}llll@{}}
    \toprule
    \textbf{Model} & \textbf{Parameters} $\vbTheta_d$ & \textbf{Scope} & \textbf{Principal point} \\
    \midrule
    Brown--Conrady & $f_x,f_y,\vbc$, 5 coeff. & global & parameter $\vbc$ \\
    OpenCV-14 & $f_x,f_y,\vbc$, 14 coeff. & global & parameter $\vbc$ \\
    Kannala--Brandt & $f_x,f_y,\vbc$, 4 coeff. & global & parameter $\vbc$ \\
    Double Sphere & $f_x,f_y,\vbc$, $\xi,\alpha$ & global & parameter $\vbc$ \\
    Central spline & $\vbd_{ij}$ per node & local & estimated, \cref{eq:supp-axis-fit} \\
    \bottomrule
  \end{tabular*}
\end{center}

\subsection{Global Models}
\label{sec:supp-global-models}
Let the $z$-axis of the camera frame be the optical axis. A
parametric model maps a direction $\vbd=(X,Y,Z)^\top$ of this frame
to a point $(x_d,y_d)$ of the normalized image plane and from there
to the sensor point
\begin{equation}
\label{eq:supp-intrinsics}
    \vbp = \bigl(f_x x_d + c_x,\; f_y y_d + c_y\bigr)^\top,
\end{equation}
with the focal lengths $f_x,f_y$ in pixels and the principal point
$\vbc=(c_x,c_y)^\top$. The models differ in the map
$\vbd\mapsto(x_d,y_d)$, which we write with the angle $\theta$
between $\vbd$ and the optical axis and the azimuth $\varphi$ about
it,
\begin{equation}
\label{eq:supp-angles}
    \theta=\operatorname{atan2}\bigl(\sqrt{X^2+Y^2},Z\bigr),
    \qquad \varphi=\operatorname{atan2}(Y,X).
\end{equation}
All four models are defined by this forward projection. The
back-projection $\vbpi^{-1}$ inverts it: \cref{eq:supp-intrinsics}
gives $(x_d,y_d)$ from $\vbp$, and the inversion of the map
$\vbd\mapsto(x_d,y_d)$, derived below for each model, gives the
direction. We call these models \emph{global}: each of their few
parameters changes the back-projection of every sensor point.

\paragraph{Brown--Conrady.} The model starts from the perspective
projection $(x,y)=(X/Z,\,Y/Z)$, whose radius is
$\rho=\sqrt{x^2+y^2}=\tan\theta$. A lens that is rotationally
symmetric about its optical axis displaces an image point along its
radius, and by symmetry the displaced radius is an odd function of
$\rho$, so that $\rho_d=\rho\,R(\rho)$ with a factor $R$ that is even
in $\rho$. Brown \cite{Brown1966DecenteringDO} truncates its series
after three terms and adds the displacement that Conrady
\cite{Conrady1919DecentredL} derived for lens elements whose centers
are not on a common axis, which is not radial:
\begin{align}
\label{eq:supp-brown-conrady}
    x_d &= x\,R(\rho) + 2p_1xy + p_2(\rho^2+2x^2),\nonumber\\
    y_d &= y\,R(\rho) + p_1(\rho^2+2y^2) + 2p_2xy,\\
    R(\rho) &= 1+k_1\rho^2+k_2\rho^4+k_3\rho^6.\nonumber
\end{align}
The radial coefficients $k_1,k_2,k_3$ describe barrel distortion
where $R<1$ and pincushion distortion where $R>1$. The tangential
coefficients $p_1,p_2$ are the components of the decentering along
the two image axes. \Cref{eq:supp-brown-conrady} has no closed-form
inverse. We solve it for $(x,y)$ iteratively from
$(x,y)=(x_d,y_d)$, which has a unique solution where the distortion is
monotonic, \ie where $\rho\,R(\rho)$ increases with $\rho$, and the
back-projection is
$\vbpi^{-1}(\vbp)=(x,y,1)^\top/\lVert(x,y,1)\rVert$. Since
$\rho=\tan\theta$ diverges for $\theta\to90^\circ$, the model cannot
represent a field of view of $180^\circ$.

\paragraph{OpenCV-14.} The 14-parameter model of OpenCV
\cite{opencv_library} extends \cref{eq:supp-brown-conrady} by three
effects. First, the radial factor becomes a rational function,
\begin{equation}
\label{eq:supp-rational}
    R(\rho) = \frac{1+k_1\rho^2+k_2\rho^4+k_3\rho^6}{1+k_4\rho^2+k_5\rho^4+k_6\rho^6},
\end{equation}
which tends to a constant for large $\rho$, where a polynomial
diverges, and therefore follows a wide-angle lens further toward the
image border. Second, thin-prism terms describe lens elements that
are tilted against the optical axis, which displace the image by even
powers of the radius along a fixed direction:
\begin{align}
\label{eq:supp-thin-prism}
    x' &= x\,R(\rho) + 2p_1xy + p_2(\rho^2+2x^2) + s_1\rho^2+s_2\rho^4,\nonumber\\
    y' &= y\,R(\rho) + p_1(\rho^2+2y^2) + 2p_2xy + s_3\rho^2+s_4\rho^4.
\end{align}
Third, the sensor may be tilted against the plane perpendicular to
the optical axis. With the rotation
$\vbT=\vbR_y(\tau_y)\,\vbR_x(\tau_x)$ of the sensor
by the angles $\tau_x,\tau_y$ about the image axes, with the entries
$T_{ij}$, the point is mapped by the homography
\begin{equation}
\label{eq:supp-tilt}
    \lambda\begin{pmatrix}x_d\\y_d\\1\end{pmatrix}
    = \begin{pmatrix} T_{33} & 0 & -T_{13}\\ 0 & T_{33} & -T_{23}\\ 0 & 0 & 1\end{pmatrix}
    \vbT\begin{pmatrix}x'\\y'\\1\end{pmatrix},
\end{equation}
which rotates the image plane into the plane of the tilted sensor
and projects it back along the optical axis, so that the principal
point stays where the optical axis meets the sensor. The 14
parameters are $k_1,\dots,k_6$, $p_1,p_2$, $s_1,\dots,s_4$ and
$\tau_x,\tau_y$, and $k_4=\dots=\tau_y=0$ is the Brown--Conrady
model. For the back-projection, \cref{eq:supp-tilt} is inverted in
closed form, as a homography, and \cref{eq:supp-thin-prism} is then
solved for $(x,y)$ iteratively as above. The model distorts
the perspective projection as well and shares its limit of
$180^\circ$.

\paragraph{Kannala--Brandt.} A fisheye lens is not designed to image
a direction at the radius $\tan\theta$ of the perspective projection,
but at a radius that stays finite at $\theta=90^\circ$, \eg $\theta$
(equidistant), $2\tan\frac{\theta}{2}$ (stereographic) or
$2\sin\frac{\theta}{2}$ (equisolid angle). All of these, and the
perspective projection itself, are odd functions of $\theta$.
Kannala and Brandt \cite{kannala-brandt} therefore model a
rotationally symmetric lens by an odd polynomial in $\theta$, whose
linear coefficient is absorbed by the focal length:
\begin{align}
\label{eq:supp-kannala-brandt}
    (x_d,y_d) &= r(\theta)\,(\cos\varphi,\sin\varphi),\nonumber\\
    r(\theta) &= \theta+k_1\theta^3+k_2\theta^5+k_3\theta^7+k_4\theta^9,
\end{align}
with $\theta$ and $\varphi$ of \cref{eq:supp-angles}. The first term
is the equidistant projection, and $k_1,\dots,k_4$ describe the
deviation of the lens from it.
We use the radially symmetric part only, as in OpenCV's fisheye model; the asymmetric radial and tangential terms of the full model \cite{kannala-brandt} are omitted.
For the back-projection, the azimuth is that of the image point,
$\varphi=\operatorname{atan2}(y_d,x_d)$, and the angle is the root of
$r(\theta)=\sqrt{x_d^2+y_d^2}$, which we find iteratively
from $\theta=\sqrt{x_d^2+y_d^2}$ and which is unique where $r$ is
monotonic. Then
\begin{equation}
    \vbpi^{-1}(\vbp)=(\sin\theta\cos\varphi,\;\sin\theta\sin\varphi,\;\cos\theta)^\top .
\end{equation}
Since $\theta$ is not restricted to less than $90^\circ$, the model
represents fields of view of $180^\circ$ and more.

\paragraph{Double Sphere.} The model of Usenko \etal
\cite{double-sphere} is constructed geometrically. The direction is
projected onto a unit sphere about the center of projection, which
gives $\vbd/d_1$ with $d_1=\lVert\vbd\rVert$. This point is expressed
in the frame of a second unit sphere whose center is displaced by
$\xi$ along the optical axis, where it is
$(X,Y,Z+\xi d_1)^\top/d_1$, and is projected onto the second sphere
by dividing by its norm $d_2/d_1$,
\begin{equation}
    d_2=\sqrt{X^2+Y^2+(\xi d_1+Z)^2}.
\end{equation}
Finally, the point $(X,Y,Z+\xi d_1)^\top/d_2$ on the second sphere is
projected perspectively from a center that is displaced by
$\frac{\alpha}{1-\alpha}$ behind the center of this sphere, which
gives
\begin{equation}
\label{eq:supp-double-sphere}
    (x_d,y_d) = \frac{(X,Y)}{\alpha d_2+(1-\alpha)(\xi d_1+Z)}.
\end{equation}
The parameter $\xi$ is thus the distance between the two spheres, and
$\alpha\in[0,1]$ the position of the final center of projection.
For $\xi=0$ and $\alpha=0$, the model reduces to the perspective projection. Both describe a
rotationally symmetric lens, and the model has no decentering terms.
Unlike the three models above, it has a closed-form inverse. With
$\rho^2=x_d^2+y_d^2$,
\begin{align}
\label{eq:supp-double-sphere-inverse}
    m_z &= \frac{1-\alpha^2\rho^2}{\alpha\sqrt{1-(2\alpha-1)\rho^2}+1-\alpha},\nonumber\\
    \vbpi^{-1}(\vbp) &= \frac{m_z\xi+\sqrt{m_z^2+(1-\xi^2)\rho^2}}{m_z^2+\rho^2}
    \begin{pmatrix}x_d\\y_d\\m_z\end{pmatrix}
    -\begin{pmatrix}0\\0\\\xi\end{pmatrix},
\end{align}
which follows from inverting the three steps in reverse order: the
first line intersects the ray of the shifted perspective projection
with the second sphere, and the second line intersects the ray from
the center of the second sphere through this point with the first
sphere. The inverse is a unit vector, and it is defined where
$(2\alpha-1)\rho^2\le1$, which bounds the image radius for
$\alpha>\frac{1}{2}$. The model represents fields of view beyond
$180^\circ$.

\subsection{The Local Model}
\label{sec:supp-local-model}
The central generic model in \cite{spline-camera,10000params} has no
projection function with physical parameters. It stores the
back-projection itself: directions at the nodes of a regular lattice
over the sensor, interpolated by a bicubic B-spline
\cite{piegl1997nurbs} and normalized. With the control directions
$\vbd_{ij}\in\mathbb{R}^3$ at the nodes $(i,j)$ and the cubic
B-spline basis functions $N_{i,3}$, $N_{j,3}$ over the lattice,
\begin{equation}
\label{eq:supp-spline-lens}
    \hat{\vbd}(\vbp) = \frac{\tilde{\vbd}(\vbp)}{\lVert\tilde{\vbd}(\vbp)\rVert},
    \quad
    \tilde{\vbd}(\vbp) = \sum_{i,j}N_{i,3}(p_x)\,N_{j,3}(p_y)\,\vbd_{ij},
\end{equation}
and $\vbTheta_d=\{\vbd_{ij}\}$. On the Sigma 14\,mm, for example, the
lattice has $35\times25$ nodes at a spacing of about 160 pixels. We
call this model \emph{local}: a cubic basis function is nonzero over
four cells of the lattice, so $\hat{\vbd}(\vbp)$ depends only on the
$4\times4$ control directions around $\vbp$, and a control direction
changes the back-projection only in the $4\times4$ cells around its
node. The model is given by its back-projection and not, as the
global models are, by its forward projection, which has no closed
form. Where a projection is needed, as for the reprojection errors of
\cref{sec:geometric-accuracy}, we invert
\cref{eq:supp-spline-lens} numerically by Levenberg--Marquardt
iterations; a second spline regressed to the inverse is an
alternative, and it is local as well.

The locality has three consequences for the calibration.
\emph{First}, the model can follow a distortion that no global model
describes, but a node is determined only by the photosites in its
own support. Nodes at the border of the lens coverage, which few
photosites see, are therefore not determined by the data, whereas
the parameters of a global model are determined by the whole image.
This is why the central spline, and no other model, is regularized
(\cref{sec:optimization}). \emph{Second}, the model has no
distinguished camera frame: rotating every control direction by a
rotation $\mathrm{R}$ and every pose by its inverse changes no ray
in the scene. A global model does not have this gauge freedom, since
its frame is fixed by its optical axis
(\cref{sec:supp-gauge-freedoms}). \emph{Third}, and for the same
reason, the model has no principal point. The intrinsics of
\cref{eq:supp-intrinsics} do not enter \cref{eq:supp-spline-lens},
which is defined over the sensor directly, so the point about which
the vignetting is symmetric has to be estimated
(\cref{sec:supp-vignetting-center}).

\subsection{The Center of the Vignetting}
\label{sec:supp-vignetting-center}
The vignetting of a lens is assumed to be symmetric about its optical axis.
Therefore, $V$ of \cref{sec:vignetting} is radially symmetric
about the principal point $\vbc$.

\paragraph{Global models.} The principal point of
\cref{eq:supp-intrinsics} is this point by definition: each of the
four models projects the direction of the optical axis,
$\vbd=(0,0,1)^\top$, to $(x_d,y_d)=(0,0)$ and thus to $\vbp=\vbc$.
We center the vignetting at the principal point of the geometric
start.

\paragraph{The local model.} For the central spline, we estimate the
optical axis as the axis about which the camera is closest to
radially symmetric. Let
\begin{equation}
\label{eq:supp-symmetric-camera}
    \vbpi_{\text{sym}}(\vbd)
    \coloneqq \vbc' + r(\theta)\begin{pmatrix} f_x\cos\varphi\\ f_y\sin\varphi\end{pmatrix},
    \quad
    r(\theta)=\theta+\sum_{k=1}^{4}\kappa_k\theta^{2k+1},
\end{equation}
be the projection of a radially symmetric camera, with $\theta$ and
$\varphi$ of \cref{eq:supp-angles}: the model of
\cref{eq:supp-kannala-brandt} with its own focal lengths $f_x,f_y$,
principal point $\vbc'$ and coefficients $\vbkappa$. We regress this
camera and a rotation $\vbR\in SO(3)$ to the rays of the
spline,
\begin{equation}
\label{eq:supp-axis-fit}
    \min_{\vbR,\,\vbc',\,f_x,\,f_y,\,\vbkappa}\;
    \sum_{\vbp\in\mathcal{R}}
    \bigl\lVert \vbp-\vbpi_{\text{sym}}\bigl(\vbR\,\hat{\vbd}(\vbp)\bigr)\bigr\rVert^{2},
\end{equation}
where $\mathcal{R}$ are the pixels in which the geometric start is
defined and $\hat{\vbd}$ is the spline back-projection of \cref{eq:supp-spline-lens}. We minimize it by Levenberg--Marquardt iterations, first
with the coefficient $\kappa_1$ alone and then with one more
coefficient at a time, each from the previous minimum, which keeps
the higher orders from folding the profile at the image border. The
symmetric camera is not used as a camera model. Its role is to
locate the axis: at the minimum, $\vbR^\top(0,0,1)^\top$ is the
axis about which the rays of the spline are closest to symmetric,
the optical axis of the lens, and $\vbc'$ is the sensor point that
images it. A lens that is not symmetric about any axis leaves a
residual in \cref{eq:supp-axis-fit}, but the axis remains defined as
its minimizer.

We then turn the camera frame onto this axis. Every control
direction is rotated, $\vbd_{ij}\leftarrow\vbR\,\vbd_{ij}$,
which rotates every ray of \cref{eq:supp-spline-lens} by
$\mathrm{R}$, since the normalization commutes with a rotation, and
every pose $(\vbR_k,\vbt_k)$ from the world to the camera
becomes $(\vbR\vbR_k,\vbR\vbt_k)$. This uses the
gauge freedom of the local model and changes no ray in the scene.
Afterwards the $z$-axis of the camera frame is the optical axis, and
$\vbc\coloneqq\vbc'$ is the principal point of the spline camera,
with the same meaning as that of a global model. The regression is
done once, on the geometric start. In our experiments it turns the
frame of the spline starts by $0.2^\circ$ to $1.6^\circ$, and the
principal point lies 8 to 25 pixels from the center of the sensor.

\section{Thin Lens Equivalence}
\label{sec:supp-thin-lens-equiv}
Let $\vbpi^{-1}$ be the perspective back-projection,
$\hat{\vbd}(\vbp)=\frac{\vbo-\vbp}{\lVert\vbo-\vbp\rVert}$, and let the
focus surface be the plane at the object distance $s$ from $\vbo$,
perpendicular to the optical axis. The ray of $\vbp$ meets this plane
at the distance $t(\vbp)=\frac{s}{\cos\theta(\vbp)}$ from $\vbo$, where
$\theta(\vbp)$ is the angle between $\hat{\vbd}(\vbp)$ and the optical
axis, so $\tau(\vbp)=\frac{\cos\theta(\vbp)}{s}$ and
$\vbx_f(\vbp)=\vbo+\frac{s}{\cos\theta(\vbp)}\hat{\vbd}(\vbp)$. In the
thin lens model, all rays from $\vbp$ through the aperture meet in the
point in which the ray through the center $\vbo$ of the lens, which is
not refracted, meets the plane of focus \cite{pharr2023physically}.
This point is $\vbx_f(\vbp)$, so that the ray through an aperture point
$\vba$ has the direction
$\frac{\vbx_f(\vbp)-\vba}{\lVert\vbx_f(\vbp)-\vba\rVert}=\vbpi^{-1}(\vbp,\vba)$
of \cref{eq:lifted-back-projection}: without distortion and with a
plane of focus, the radiometric camera is the thin lens.

\section{Inverse Focus Distance}
\label{sec:supp-inverse-depth}
The lifted back-projection \cref{eq:lifted-back-projection} depends
on the focus surface through the distance $t(\vbp)=1/\tau(\vbp)$
from $\vbo$ to the point of focus. We show that its derivative \wrt
the distance $t$ vanishes for a far focus, whereas its derivative
\wrt the inverse distance $\tau$ does not, which is why we
parameterize the focus surface by $\tau$
\cite{unified-inverse-depth,direct-sparse-odometry}. Let
$\vbo = \mathbf{0}$ without loss of generality. With the distance,
\begin{align*}
    &\frac{\partial}{\partial t} \frac{t(\vbp)\hat{\vbd}(\vbp)-\vba}{\lVert t(\vbp)\hat{\vbd}(\vbp)-\vba\rVert}
    = \frac{\hat{\vbd}(\vbp)\lVert t(\vbp)\hat{\vbd}(\vbp)-\vba\rVert}{\lVert t(\vbp)\hat{\vbd}(\vbp)-\vba\rVert^2}\\
    &\qquad-\frac{(t(\vbp)\hat{\vbd}(\vbp)-\vba)\frac{t(\vbp)\lVert \hat{\vbd}(\vbp) \rVert^2 - \hat{\vbd}(\vbp)\cdot\vba}{\lVert t(\vbp)\hat{\vbd}(\vbp)-\vba\rVert}}{\lVert t(\vbp)\hat{\vbd}(\vbp)-\vba\rVert^2}
\end{align*}
\begin{align*}
    &\lim_{t\to \infty}\frac{\partial}{\partial t} \frac{t(\vbp)\hat{\vbd}(\vbp)-\vba}{\lVert t(\vbp)\hat{\vbd}(\vbp)-\vba\rVert}\\
    &= \lim_{t\to \infty}\frac{1}{t(\vbp)}\Bigl(\frac{\hat{\vbd}(\vbp)\lVert \hat{\vbd}(\vbp)-\frac{\vba}{t(\vbp)}\rVert}{\lVert \hat{\vbd}(\vbp)-\frac{\vba}{t(\vbp)}\rVert^2}\\
    &\qquad\qquad-\frac{(\hat{\vbd}(\vbp)-\frac{\vba}{t(\vbp)})\frac{\lVert \hat{\vbd}(\vbp) \rVert^2 - \frac{\hat{\vbd}(\vbp)\cdot\vba}{t(\vbp)}}{\lVert \hat{\vbd}(\vbp)-\frac{\vba}{t(\vbp)}\rVert}}{\lVert \hat{\vbd}(\vbp)-\frac{\vba}{t(\vbp)}\rVert^2}
    \Bigr)
    = 0.
\end{align*}
A focus surface that is far away thus receives no gradient. With the
inverse distance $\tau=\frac{1}{t}$,
\begin{align*}
    &\frac{\partial}{\partial \tau}\frac{-\tau(\vbp)\vba + \hat{\vbd}(\vbp)}{\lVert-\tau(\vbp)\vba + \hat{\vbd}(\vbp)\rVert}\\
    = &\frac{-\vba\lVert \hat{\vbd}(\vbp)-\tau(\vbp)\vba\rVert + (\hat{\vbd}(\vbp)-\tau(\vbp)\vba)\frac{\hat{\vbd}(\vbp)\cdot\vba - \tau(\vbp)\lVert\vba\rVert^2}{\lVert \hat{\vbd}(\vbp)-\tau(\vbp)\vba\rVert}}{\lVert \hat{\vbd}(\vbp)-\tau(\vbp)\vba\rVert^2}
\end{align*}
\begin{align*}
    \lim_{\tau\to 0}\frac{\partial}{\partial \tau}\frac{-\tau(\vbp)\vba + \hat{\vbd}(\vbp)}{\lVert-\tau(\vbp)\vba + \hat{\vbd}(\vbp)\rVert}
    = \frac{-\vba\lVert\hat{\vbd}(\vbp)\rVert + \hat{\vbd}(\vbp)\frac{\hat{\vbd}(\vbp)\cdot\vba}{\lVert \hat{\vbd}(\vbp)\rVert}}{\lVert\hat{\vbd}(\vbp)\rVert^2}
\end{align*}
and since $\lVert\hat{\vbd}(\vbp)\rVert=1$, this limit is
$-\bigl(\vba-(\vba\cdot\hat{\vbd}(\vbp))\,\hat{\vbd}(\vbp)\bigr)$, the
negative of the component of $\vba$ perpendicular to the ray. It is
nonzero for every aperture point off the ray, so the focus surface
keeps a gradient however far it is.

\section{Vignetting of the Thin Lens}
\label{sec:supp-thin-lens-vignetting}
The radiant flux \cref{eq:per-channel-rad-flux} integrates over the
solid angle $\calOmega(\vbp)$ under which the sensor point $\vbp$ sees
the aperture. For a thin lens with the image distance $s'$, changing
the integration domain to the aperture area $\mathcal{A}$ by the
transformation
$\vbomega = (\vba - \vbp)/\lVert \vba - \vbp \rVert$~\cite{pharr2023physically}
gives
\begin{align}
    \Phi(\vbu)\label{eq:rad-flux-disent}
    = \int_{\mathcal{I}_{\vbu}}\int_{\mathcal{A}}&\frac{\cos^4\theta(\vba-\vbp)}{s'^2}\\
    &L\bigl(\vbx(\vbp,\vba),-\vbpi^{-1}(\vbp,\vba)\bigr)\odif{\vba}\odif{\vbp}\nonumber,
\end{align}
where $\theta(\vba-\vbp)$ is the angle between $\vba-\vbp$ and the
optical axis. This separates the falloff toward the sensor border from
the solid angle: the factor $\cos^4\theta/s'^2$ is a well-known
property of the thin lens and describes its
vignetting~\cite{Goossens_2019}. \Cref{fig:vignetting} shows that the
vignetting of real lenses differs much from it. In the radiometric
camera, we therefore replace this factor by the generic vignetting
function $V(\vbp)$, normalized by the aperture area $\pi r^2$:
\begin{align}
    \Phi(\vbu)
    \approx \int_{\mathcal{I}_{\vbu}}&\frac{V(\vbp)}{\pi r^2}
    \int_{\mathcal{A}}L\bigl(\vbx(\vbp,\vba),-\vbpi^{-1}(\vbp,\vba)\bigr)\nonumber\\
    &\cdot\cancel{\frac{\cos^4\theta(\vba-\vbp)}{s'^2}}\odif{\vba}\odif{\vbp},
\end{align}
where the back-projection of the thin lens is replaced by the lifted
back-projection $\vbpi^{-1}(\vbp,\vba)$ of \cref{eq:lifted-back-projection},
which is \cref{eq:rad-flux-with-vignetting} after the substitution
$\vba=\vbo+r\vbxi$ that maps the aperture to the unit disc
$\mathcal{D}$, with $\odif{\vba}=r^2\odif{\vbxi}$.

\section{Knots of the Vignetting Spline}
\label{sec:supp-vignetting-uniform-knots}
The vignetting spline is defined over the normalized distance
\begin{equation*}
    \rho(\vbp) \coloneqq \frac{\lVert \vbp-\vbc \rVert}{r_{\max} + \lVert \vbp-\vbc \rVert},
\end{equation*}
which maps the distances $[0,r_{\max}]$ of the illuminated pixels to
$[0,\frac{1}{2}]$ and every larger distance into $(\frac{1}{2},1)$.
The knots are placed at equal steps of the distance in the image and
not of $\rho$: with $m$ spans up to $r_{\max}$, they lie at the
distances $d_k = k\,r_{\max}/m$, $k=0,\dots,m$, and thus at
\begin{equation*}
    \rho_k = \frac{d_k}{r_{\max}+d_k} = \frac{k}{m+k}.
\end{equation*}
We use $m=13$. Two further knots follow beyond $r_{\max}$ at the same
spacing, a last span reaches $\rho=1$, and the knot vector is clamped
at both ends. The spline then has 19 control points, of which 16 act
on the illuminated pixels.

\section{Boundary Terms of the Radiant Flux}
\label{sec:supp-boundaries}
\Cref{sec:boundary-terms} removes the boundary terms that arise
in our formulation of the calibration. Here we list the boundary terms of the radiant
flux in a general scene and say how each is treated.

\paragraph{The problem.}
Differentiable ray tracing learns scene parameters by minimizing the
error between rendered and captured images
\cite{NimierDavidVicini2019Mitsuba2, Li:2018:DMC}. It has to
differentiate integrals whose domain, or whose integrand's
discontinuities, depend on the parameters
\cite{Zhang:2020:PSDR, Li:2018:DMC, mitsuba-projective}. By the
multidimensional Leibniz integral rule, such a derivative is the sum
of the integral of the derivative, the interior integral, and a
boundary integral over the moving boundaries
\cite{Casey2012JumpConditions, leibniz-multi-dim, Zhang:2020:PSDR}.
Automatic differentiation of a sampled estimate computes the
interior integral only \cite{Zhang:2020:PSDR, mitsuba-projective}.

\paragraph{The boundary terms of a general scene.}
Differentiating the radiant flux $\Phi$
(\cref{eq:rad-flux-with-vignetting}) \wrt the parameters
$\bar{\vbTheta}$ differentiates the integrals over the pixel
$\mathcal{I}_{\vbu}$ and the aperture $\mathcal{A}(r)$ and, for each
bounce of a ray, the integral over the hemisphere
$\mathcal{H}^2(\vbx)$ in the radiance $L$ (\cref{eq:scene-rad}). With
\begin{align*}
    \hat{E}(\vbp,\vba)\coloneqq L\bigl(\vbx(\vbp,\vba),-\vbpi^{-1}(\vbp,\vba)\bigr),
\end{align*}
we identify the following sources of boundary terms.\\
\textbf{I.} $\vba \mapsto \hat{E}(\vbp,\vba)$ has parameter-dependent
discontinuities over $\mathcal{A}(r)$, from surface silhouettes,
sharp shadows and a discontinuous surface texture, \eg the pattern
of a calibration target.\\
\textbf{II.} $\vbp \mapsto \hat{E}(\vbp,\vba)$ has the same
discontinuities over $\mathcal{I}_{\vbu}$.\\
\textbf{III.} The aperture $\mathcal{A}(r)$ depends on the radius
$r\in\bar{\vbTheta}$.\\
\textbf{IV.} By the chain rule, every bounce requires
$\frac{\partial}{\partial \vbx}L(\vbx,\vbomega_o)$, the derivative of
the integral over $\mathcal{H}^2(\vbx)$ in \cref{eq:scene-rad}. Its
integrand has discontinuities from the silhouettes, the shadows and
the discontinuous textures seen from $\vbx$, which move with $\vbx$,
and its domain $\mathcal{H}^2(\vbx)$ depends on $\vbx$. Light sources
with a boundary, such as area lights, add a further discontinuity of
the incident radiance. This repeats at every further bounce.

\paragraph{The terms of the calibration.}
A few of these sources remain in a calibration scene, and we remove
all of them by a reformulation (\cref{sec:boundary-terms},
\cref{fig:boundary}). For \textbf{III}, we transform the aperture to
the fixed unit disc $\mathcal{D}\subseteq\mathbb{R}^2$ with
$\vba=\vbo+r\vbxi$, $\vbxi\in\mathcal{D}$, which gives
\cref{eq:rad-flux-with-vignetting},
\begin{align*}
    \Phi(\vbu)
    =\frac{1}{\pi}\int_{\mathcal{I}_{\vbu}}V(\vbp)
    \int_{\mathcal{D}} \hat{E}(\vbp,\vbo+r\vbxi)\odif{\vbxi}\odif{\vbp},
\end{align*}
whose domain does not depend on $r$. For the edges of the pattern in
\textbf{I} and \textbf{II}, we use that a print is never perfectly
sharp. We model an edge of the print by the convolution of the print
design with a Gaussian of the standard deviation
$\sigma_w = w/1.805951$, where $w=50\,\mathrm{\mu m}$ is the
$10$--$70\,\%$ edge width of a realistic print
\cite{rosalen2019cardboards}. The albedo is then the cumulative
Gaussian of the signed distance to the edges of the design, which we
compute from the print file, and is smooth.
Moreover, to exclude surface silhouettes (\textbf{I}, \textbf{II}), we use a pixelmask for the single calibration board.

\paragraph{The terms that do not arise.}
The remaining sources are absent from a calibration scene. The
silhouettes in \textbf{I} and \textbf{II} are not seen: we render
only the pixels of a mask $\mathcal{M}\subseteq\mathcal{I}$, with a
padding, whose rays all meet the target. For \textbf{IV}, the scene
contains the target alone, and the light scattered by its
environment is represented by a smooth lighting
(\cref{sec:lighting}), so that no silhouette, shadow or boundary of
a light is seen from $\vbx$. The dependence of the hemisphere
$\mathcal{H}^2(\vbx)$ on $\vbx$ and the silhouette of the target seen
from a point of the target contribute nothing, since the cosine in
\cref{eq:scene-rad} vanishes at the horizon, $\cos(\pi/2)=0$. A
general scene requires these terms, which differentiable renderers
estimate by sampling the boundaries
\cite{Li:2018:DMC, Zhang:2020:PSDR, mitsuba-projective}; we leave
them to future work.

\section{Lighting and Reflectance}
\label{sec:supp-reflectance}
\paragraph{Lighting.}
The radiance of the spherical-harmonic lighting of degree
$l_{\max}$ seen from the direction $\vbomega$ is
\begin{align*}
    L^{\text{SH}}(\vbomega;\vbTheta_l)
    \coloneqq \sum_{l=0}^{l_{\max}} \sum_{m=-l}^{l} a_{l}^{m}\,y_l^m(\theta(\vbomega), \phi(\vbomega)),
\end{align*}
with the basis functions $y_l^m$, the coefficients
$\vbTheta_l=\{a_{l}^m\}\subseteq\mathbb{R}$ per color channel and the
spherical coordinates $(\theta(\vbomega), \phi(\vbomega))$ of
$\vbomega$ \cite{sphericalHarmonics, peter-pikesloanand2002precomputed, sh-legendre, ramamoorthi-2001}.
We use $l_{\max}=2$ \cite{ramamoorthi-2001}. \Cref{sec:lighting}
places it on a sphere of finite radius.

\paragraph{Reflectance.}
We compute \cref{eq:scene-rad} by Monte Carlo estimation
\cite{pharr2023physically}. $N$ microfacet normals \cite{microfacet}
$\vbn_k$ are sampled from the GGX distribution of visible normals,
$\vbn_k \sim D^{\text{v}}(\cdot, \vbomega_o; \alpha)$, by sampling a
spherical cap \cite{sphericalCaps}, and give the directions
$\vbomega_{i,k}$ by reflection, with
$\text{PDF}(\vbn_k, \vbomega_o; \alpha)= D^{\text{v}}(\vbn_k, \vbomega_o;\alpha)/(4 (\vbomega_o \cdot \vbn_k))$,
so that
\begin{align*}
    &L(\vbx,\vbomega_o;\vbTheta_r)
    \approx \frac{1}{N} \sum_{k=1}^N \frac{f^{\text{CT}}_k\,L_k\cos\theta_{i,k}}{\text{PDF}(\vbn_k, \vbomega_o; \alpha)},
\end{align*}
with $L_k\coloneqq L(\vbx'(\vbx,\vbomega_{i,k}),-\vbomega_{i,k};\vbTheta_r)$
and the Cook--Torrance reflectance \cite{cookTorrance, sphericalCaps}
\begin{align*}
    &f^{\text{CT}}_k
    \coloneqq f^{\text{CT}}(\vbx,\vbomega_o,\vbomega_{i,k}; \vbTheta_m)
    =\mu\frac{\rho(\vbx)}{\pi}\\
    &\qquad\quad+ (1-\mu)\frac{F(\vbomega_o, \vbomega_h; \eta)\, D(\vbomega_h; \alpha)\, G_2(\vbomega_o, \vbomega_{i,k})}{4\cos\theta_o\cos\theta_{i,k}}.
\end{align*}
Here $\vbomega_h$ is the half-vector between $\vbomega_o$ and
$\vbomega_{i,k}$, $\theta_o$ and $\theta_{i,k}$ are the angles
between the directions and the surface normal, $F$ is the exact
unpolarized Fresnel term of a dielectric with the index of
refraction $\eta$ per color channel \cite{pharr2023physically}, $D$
the isotropic GGX distribution, $\alpha_x=\alpha_y=\alpha$
\cite{Heitz2014Microfacet}, and $G_2$ the height-correlated masking
and shadowing function \cite{sphericalCaps, Heitz2014Microfacet}.
$\vbTheta_r\coloneqq\vbTheta_l\cup\vbTheta_m$ are the parameters of
the lighting and the material. The roughness and the index of
refraction are blended between their values for ink and for the
blank print by the albedo $\rho(\vbx)$.

\section{Gauge Freedoms}
\label{sec:supp-gauge-freedoms}
The parameters $\bar{\vbTheta}$ contain three families of
transformations that leave every rendered photosite value $\Phi$
unchanged. We fix each of them exactly by the parameterization, \ie by
removing the corresponding direction from the unknowns, and not by a
penalty or a prior, so that no weight has to be chosen and the
remaining parameters keep their meaning.

\paragraph{Vignetting and lighting.} $\Phi$ is the product of the
vignetting $V$ and a radiance that is linear in the lighting
$\vbTheta_l$. Scaling $V$ by $\sigma>0$ and the lighting by
$1/\sigma$ therefore leaves $\Phi$ unchanged. We hold $V=1$ at the
centre of the vignetting, the principal point: the control points are scaled once so that
the centre value is one, and the direction of the control points that
changes the centre value is eliminated from the unknowns, so that
every update keeps it exactly. The overall scale is then carried by
the lighting alone.

\paragraph{Rotation of the spline camera and the poses.} A parametric
model fixes its camera frame by its optical axis. The directions of
the generic spline model, in contrast, can all be rotated by the same
rotation, and the poses of all views by its inverse, without changing
any ray in the scene. We hold three coordinates of the directions that
determine this rotation: both tangent coordinates of the node whose
direction is closest to the optical axis, and, at the node farthest
from it in angle, the tangent coordinate most aligned with the roll
about that direction. These three coordinates are a nonsingular
function of the rotation, so holding them removes the gauge and no
degree of freedom of the lens. The poses remain free.

\paragraph{Board warping and the poses.} A height field over the board
that is affine in the board coordinates, \ie an offset along the
normal and two tilts, is a rigid motion of the board and is already
described by the pose of each view. The nodal heights of the
thin-plate spline are therefore restricted to those whose interpolant
has no affine term, which leaves pure bending. Likewise, the skew of
the pixel grid is held for every model, since it is interchangeable
with the in-plane shear of the printed target.

\section{Linear Parameters of the Radiant Flux}
\label{sec:supp-linear}
The start of the calibration (\cref{sec:optimization}) uses that the
radiant flux \cref{eq:rad-flux-with-vignetting} is linear in two of
its parameter blocks and affine in a third.

\paragraph{Vignetting.} The vignetting
$V(\vbp)=\sum_{i} N_{i,3}(\cdot)\,P_i$ is linear in its control points
$\vbTheta_v=\{P_i\}$ and multiplies the integral over the aperture,
so $\Phi$ is linear in $\vbTheta_v$.

\paragraph{Lighting.} The lighting $L^{\text{SH}}$ is linear in its
coefficients $\vbTheta_l=\{a_l^m\}$
(\cref{sec:supp-reflectance}), the radiance that leaves the target
(\cref{eq:scene-rad}) is linear in the incident radiance, and the
integrals over the hemisphere, the aperture and the pixel are linear,
so $\Phi$ is linear in $\vbTheta_l$.

\paragraph{Diffuse weight.} The Cook--Torrance reflectance is
$f=\mu f_{\text{d}}+(1-\mu)f_{\text{s}}$ with its diffuse part
$f_{\text{d}}$ and its specular part $f_{\text{s}}$, so $\Phi$ is
affine in $\mu$.

With the other parameters held, each of the three blocks is therefore
the solution of a linear least-squares problem in the residuals of
\cref{eq:objective}, which we solve once per block at the start: the
lighting with $V\equiv1$, then the vignetting, then $\mu$. The flux
is bilinear in the vignetting and the lighting, which is the scale
gauge of \cref{sec:supp-gauge-freedoms}.

\section{Data Set}
\label{sec:supp-dataset}
\begin{table}[t]
	\centering
	\footnotesize\setlength{\tabcolsep}{2.5pt}
	\begin{tabular*}{\linewidth}{@{\extracolsep{\fill}}llccc@{}}
		\toprule
		\textbf{Lens} & \textbf{Type} & \textbf{Geom.} & \textbf{Vign.} & \textbf{Defocus} \\
		\midrule
		Tamron 20\,mm F/2.8 Di III & wide angle & \checkmark & \checkmark & \checkmark \\
		Sigma 14\,mm F2.8 EX HSM & rectilinear & \checkmark & \checkmark & \checkmark \\
		Sigma 8\,mm F3.5 EX DG & circular fisheye & \checkmark & \checkmark & \checkmark \\
		24--70\,mm zoom at 48\,mm & normal & & & \checkmark \\
		\bottomrule
	\end{tabular*}
	\caption{The lenses of the data set and the experiments they take part in: geometric accuracy (\cref{sec:geometric-accuracy}), vignetting (\cref{sec:experiment-vignetting}) and defocus (\cref{sec:experiment-defocus}). The camera is a Sony ILME-FX3 throughout.}
	\label{tab:dataset-lenses}
\end{table}
\Cref{tab:supp-dataset} lists every capture of the data set (\cref{sec:dataset}), grouped by experiment. All images are Sony ILME-FX3 RAW files ($4256\times2848$ decoded pixels) at ISO\,100. The board distance is not measured: it is the median perpendicular distance from the camera centre to the target plane over the estimated poses of a set. \emph{Captured} counts the images of a set and \emph{used} those in the experiment's configuration: a training image is left out when the geometric pipeline that starts our refinement does not localize the board in it, and a test image when its dots are not registered. Each Thorlabs test set is captured together with its training set, in the same session and with the same focus and aperture.
\begin{table*}[t]
  \centering
  \small
  \setlength{\tabcolsep}{4pt}
  \begin{tabular}{@{}lllrrlrr@{}}
    \toprule
    lens & target & role & distance [mm] & $f/$ & exposure & captured & used \\
    \midrule
    \multicolumn{8}{@{}l}{\textbf{Geometric accuracy} (\cref{sec:geometric-accuracy})} \\
    Sigma 14\,mm & Siemens stars & train & 295 & 22 & 1.6\,s & 168 & 132 \\
    Sigma 14\,mm & AprilGrid & train & -- & 22 & 2\,s & 408 & 399 \\
    Sigma 14\,mm & Thorlabs & test & 300 & 22 & 1/10\,s & 142 & 141 \\
    Sigma 8\,mm fisheye & AprilGrid & train & 186 & 22 & 2\,s & 407 & 249 \\
    Sigma 8\,mm fisheye & Thorlabs & test & 121 & 22 & 1/13\,s & 154 & 154 \\
    Sigma 8\,mm fisheye & Siemens stars & train & -- & 32 & 3.2\,s & 301 & 162 \\
    Sigma 8\,mm fisheye & Thorlabs & test & -- & 32 & 1/10\,s & 254 & 82 \\
    Tamron 20\,mm & Siemens stars & train & 189 & 22 & 1.3\,s & 264 & 209 \\
    Tamron 20\,mm & Thorlabs & test & 197 & 22 & 1/60\,s & 112 & 112 \\
    Tamron 20\,mm & AprilGrid & train & -- & 22 & 2.5\,s & 411 & 404 \\
    Tamron 20\,mm & Thorlabs & test & 448 & 22 & 1/60\,s & 142 & 142 \\
    \addlinespace
    \multicolumn{8}{@{}l}{\textbf{Vignetting} (\cref{sec:experiment-vignetting})} \\
    Sigma 14\,mm & Siemens stars & train & 762 & 4 & 1/8\,s & 174 & 166 \\
    Sigma 14\,mm & flat field & reference & -- & 4 & 1/640\,s & 1 & 1 \\
    Sigma 8\,mm fisheye & Siemens stars & train & 693 & 4 & 1/10\,s & 160 & 63 \\
    Sigma 8\,mm fisheye & flat field & reference & -- & 4 & 1/800\,s & 1 & 1 \\
    Tamron 20\,mm & Siemens stars & train & -- & 2.8 & 1/50\,s & 163 & 138 \\
    Tamron 20\,mm & flat field & reference & -- & 2.8 & 1/2000\,s & 1 & 1 \\
    \addlinespace
    \multicolumn{8}{@{}l}{\textbf{Defocus} (\cref{sec:experiment-defocus})} \\
    Tamron 20\,mm & Siemens stars & train & 440 & 8 & 1/4\,s & 340 & 332 \\
    Tamron 20\,mm & Thorlabs & test & -- & 8 & 1/800\,s & 190 & 190 \\
    zoom at 48\,mm & Siemens stars & train & -- & 10 & 0.8\,s & 171 & 159 \\
    zoom at 48\,mm & Thorlabs & test & -- & 10 & 1/500\,s & 54 & 54 \\
    Sigma 14\,mm & Siemens stars & train & 931 & 2.8 & 1/15\,s & 479 & 475 \\
    Sigma 14\,mm & text board & test & -- & 2.8 & 1/15\,s & 106 & 106 \\
    \bottomrule
  \end{tabular}
  \caption{The captures of the data set, grouped by experiment, each training set followed by the test set captured with it. \emph{Role} is the training/test split: training images calibrate the camera, test images (Thorlabs) only score it. \emph{Captured}: the images of the set; \emph{used}: those that enter the experiment, \ie for a training set the images in which the geometric pipeline that starts our refinement localizes the board, and for a test set those whose dots are registered. The \emph{used} counts are those of the starts before the geometric image changed to the full-resolution green image (\cref{sec:supp-thorlabs-protocol}) and are to be regenerated. Distance: the median camera-to-target-plane distance inferred from the estimated poses (--: not yet computed). Of the 106 images of the text board, 63 fit the texture, 22 choose its regularization and 21 test it. \emph{Flat field}: one image per lens through a diffuser (a white acrylic plate over the lens, hood on, lit by diffuse daylight), captured in the same session as the lens's vignetting set (2026-10-28); it is the reference the learned vignetting is compared with, not used in any calibration. The fisheye Siemens-star set of 2026-10-27 (150 images, 118 Thorlabs) is excluded: the start pipeline localized too few of its images.}
  \label{tab:supp-dataset}
\end{table*}

The runtime of a calibration is given in \cref{sec:supp-runtime}.

\section{Geometric Starts}
\label{sec:supp-starts}
Every radiometric refinement starts from a geometric calibration of
the same model on the same images. All of them are calibrations from
correspondences between detected image points and the known points of
the target, which differ in the detector and in the tool that
estimates the camera.

\paragraph{AprilGrid.} The corners are detected by OpenCV's ArUco
detector \cite{garrido2014aruco} with sub-pixel refinement. On these
corners, Brown--Conrady and OpenCV-14 are calibrated by OpenCV's
calibration \cite{opencv_library,zhang2000flexible}, Kannala--Brandt
by its fisheye calibration, and the central spline by the calibration
of Schöps \etal \cite{10000params}.
Double Sphere is calibrated by Kalibr~\cite{furgale2013unified}, which
detects the corners with its own AprilGrid detector.

\paragraph{Siemens stars.} The features are detected by the detector
of Schöps \etal \cite{10000params}, whose calibration gives the
central spline. OpenCV and Kalibr cannot detect this target.
Brown--Conrady and OpenCV-14 are calibrated by OpenCV's calibration on
the correspondences of Schöps's detector, and Kannala--Brandt by its
fisheye calibration on them. Our own estimator replaces them for all
three models on the Sigma 14\,mm and for Kannala--Brandt on the
fisheye. Double Sphere, which OpenCV does not provide, is always
calibrated by our own estimator: a robust pose of the target per
image, followed by a joint refinement of the camera and all poses
that minimizes the reprojection error.

\section{Thorlabs Evaluation}
\label{sec:supp-chrome-evaluation}
The geometric accuracy of every camera model in \cref{sec:geometric-accuracy} is measured on images of a high-precision distortion target that none of the models was calibrated on. \Cref{sec:supp-thorlabs-target} defines the target and \cref{sec:supp-thorlabs-protocol} the evaluation protocol.

\subsection{The Thorlabs Distortion Target}
\label{sec:supp-thorlabs-target}
\begin{figure}[t]
  \centering
  \includegraphics[width=\linewidth]{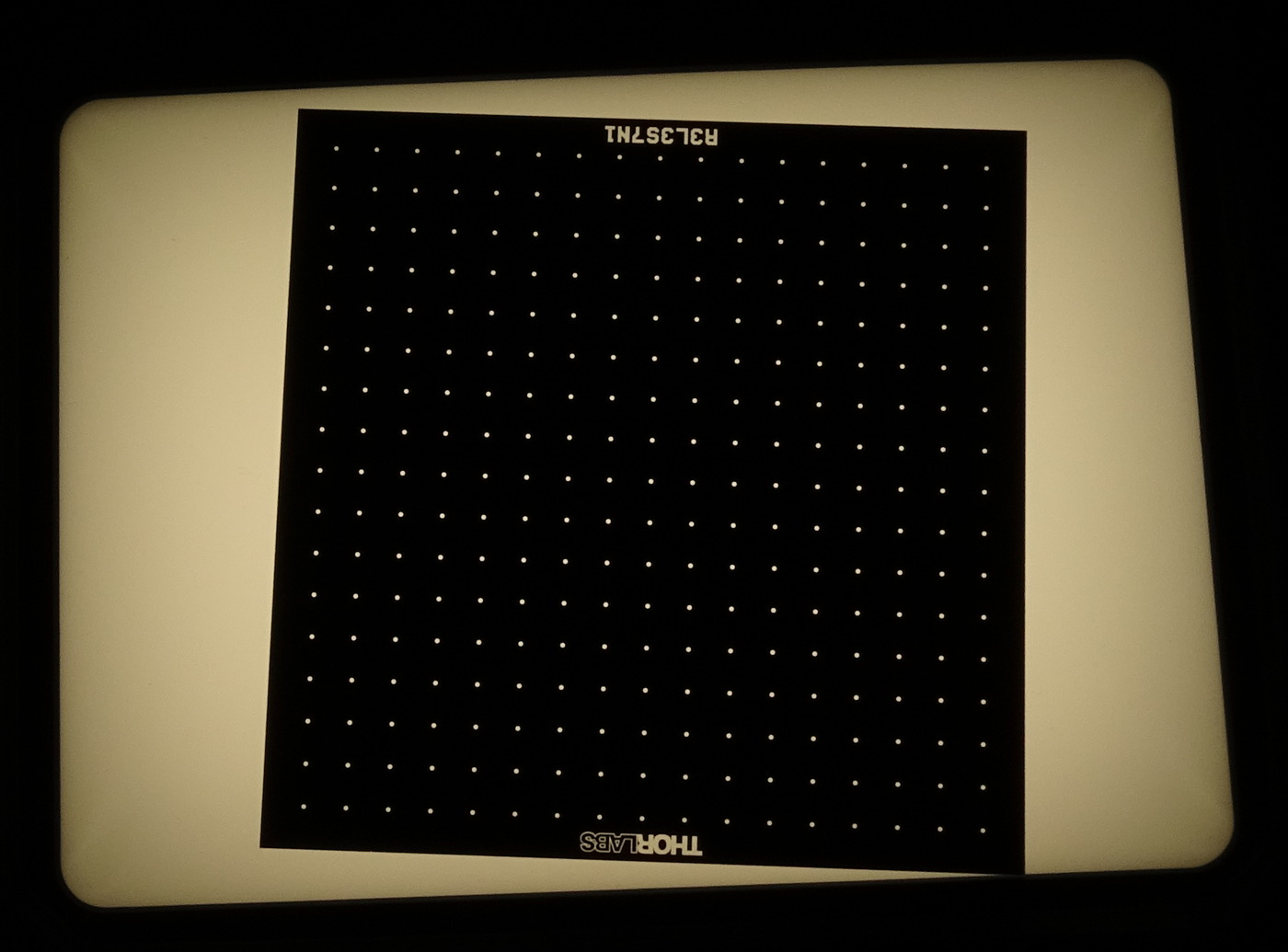}
  \caption{The Thorlabs R3L3S7N1 target on its LED backlight, from our capture.}
  \label{fig:supp-thorlabs-target}
\end{figure}

The test target is the Thorlabs R3L3S7N1 negative grid distortion target \cite{thorlabsR3L3S7N1} (\cref{fig:supp-thorlabs-target}): $17\times17$ clear dots of $0.5\,\mathrm{mm}$ diameter at $5\,\mathrm{mm}$ spacing in a chrome coating on a $90\,\mathrm{mm}\times90\,\mathrm{mm}$ soda-lime glass plate, with the dot diameter, the spacing and the corner-to-corner distance within $\pm5\,\mathrm{\mu m}$ (\cref{tab:supp-thorlabs-target}). Its dimensions, from the manufacturer's drawing, define the target points $\vbX$; no part of the target is estimated. A tolerance of $\pm5\,\mathrm{\mu m}$ is far below the size of one pixel's footprint on the target at our capture distances, so the target geometry can be treated as exact.
We captured the images in a dark room with the target backlit by an LED panel, producing bright dots on a dark background (\cref{fig:supp-thorlabs-target}).
\begin{table}[t]
  \centering
  \small
  \begin{tabular}{@{}ll@{}}
    \toprule
    grid & $17\times17$ dots, $80\times80\,\mathrm{mm}$ \\
    dot spacing & $5.000\,\mathrm{mm}$ \\
    dot diameter & $0.500\,\mathrm{mm}$ \\
    tolerance$^{*}$ & $\pm5\,\mathrm{\mu m}$ \\
    substrate & soda-lime glass, $90\times90\,\mathrm{mm}$ \\
    coating & OD\,3 chrome, $0.12\,\mathrm{\mu m}$ \\
    polarity & clear dots, opaque field \\
    \bottomrule
  \end{tabular}
  \caption{Specifications of the Thorlabs R3L3S7N1 target (manufacturer's drawing, 15 Nov 2024). $^{*}$Of the dot diameter, the dot spacing and the corner-to-corner distance.}
  \label{tab:supp-thorlabs-target}
\end{table}

\subsection{Evaluation Protocol}
\label{sec:supp-thorlabs-protocol}
\paragraph{Capture.}
The target is backlit by an LED panel in a dark room, so the dots are the only light in the image. The target is kept fixed, and the camera, mounted on a ball head on a pivot arm, is rotated between the images so that the dots cover the whole image, the rim included. The distance to the target is in the same range as for the Siemens-star and AprilGrid captures of the same lens, and the aperture is the same ($f/22$; $f/32$ for the fisheye at $30\,\mathrm{cm}$), so the defocus is comparable.

\paragraph{Detection.}
We detect the dots with a Förstner blob detector on the full-resolution green image, in which each red and blue photosite is interpolated bilinearly from its green neighbours (the image every geometric detector of our pipeline uses, the Siemens-star front end included), \cite{forstner1987fast} and take the blob centroid as the image of the dot centre. The imaged dots cover a median of 12 to 34\,px$^2$ per set (diameters of 4 to 7\,px, full-resolution pixels) across the image set, so we assume that identifying the blob centroid with the dot centre introduces negligible bias. The imaged dots cover a median of 12 to 34\,px$^2$ per set (diameters of 4 to 7\,px), so identifying the blob centroid with the dot centre introduces negligible bias.

\paragraph{Correspondences.}
In each image, the detected dot centres are assigned to the $17\times17$ grid nodes once, before any model is evaluated: an LO-RANSAC\ifarxiv~\cite{ransac, lo-ransac, rsdr, lodr}\fi{} pose of the target with a one-to-one (Hungarian) assignment\ifarxiv~\cite{hungarian, munkres}\fi, in which a detection farther than half the grid spacing ($2.5\,\mathrm{mm}$ on the target) from every node stays unassigned.
These correspondences are the same for every model, and no model rejects dots afterwards.

\paragraph{Motion.}
The camera model under test is held fixed. Since the camera rotates about one unknown pivot, a target point $\vbX$ is in view $k$ at $\vbR_k(\vbX-\vbc)+\vbd$, with a rotation $\vbR_k$ per view and the centre $\vbc$ and the offset $\vbd$ shared by all views ($3N+6$ parameters for $N$ views, instead of $6N$ for free poses). They are fitted by minimizing the robust reprojection error. Constraining the motion in this way keeps the unknown poses from absorbing reprojection error at test time.

\paragraph{Train and test split.}
The grid nodes $(i,j)$ are split by the parity of $i+j$ into two interleaved halves, like the colours of a checkerboard. The motion is fitted on one half and the reprojection error is measured on the other; the halves are then swapped and the errors of both are pooled. Every reported residual is therefore of a dot that did not enter the fit, and both halves cover the whole image, the rim included.

\paragraph{Statistics.}
We report the median, mean, $95$th percentile and RMS of the reprojection error in pixels. A model is evaluated on a dot only if it back-projects the dot and projects its fitted target point; the models of a set are compared on the dots that every one of them evaluates, so all rows of a set rest on the same dots.
A model that is not defined over the whole image, \eg a calibration restricted to the area its detections covered or a parametric distortion that does not reach the rim, leaves out the dots it cannot evaluate, which lie at the image border and are the hardest ones; the common dots then favour such a model. We therefore also computed each model's coverage, the share of the registered dots it evaluates, and its error on all of them. Every model of \cref{tab:refinement-chrome} evaluates at least 99.4\,\% of the registered dots, except the spline start on the fisheye's AprilGrid set (95.0\,\%), whose refinement, estimated over the whole lens coverage, evaluates 99.4\,\%. The median error of a model on all of its dots differs from that on the common dots by at most 0.002\,px, and by at most 6\,\% on the fisheye's AprilGrid set, where the common dots leave out the border.

\paragraph{Confidence intervals.}
We compare two models of a set, \eg a camera after the radiometric refinement and its geometric start, by the difference of a statistic of the reprojection error (median, mean, $95$th percentile or RMS), each computed over the dots that every model of the set evaluates, pooled over all images. The interval of this difference is a paired, nonparametric bootstrap over images. The dots of one image are correlated through its pose, so the image is the resampling unit and not the dot. In each of $10{,}000$ draws, as many images as the set contains are drawn with replacement; the statistic of both models is recomputed on the dots of the drawn images, an image drawn several times counting as often, and their difference is taken. Both models are thus always evaluated on the same images and dots, which removes the variation between images from the comparison. From the $10{,}000$ differences we take the $95\,\%$ bias-corrected and accelerated (BCa) interval \cite{efron1987better}, whose bias correction is the share of draws below the observed difference and whose acceleration is estimated by a jackknife that leaves out one image at a time. The four statistics share the same draws, and the random seed is fixed, so the intervals are reproducible. A difference whose interval excludes zero is significant at the $5\,\%$ level. The intervals are computed over the whole image and, separately, over its centre and its edge (radii below $0.4$ and above $0.6$ of the largest radius).

\paragraph{Results.}
\Cref{tab:supp-refinement-a,tab:supp-refinement-b} give the four statistics of \cref{tab:refinement-chrome} for every camera model, lens and calibration target, with the difference between the radiometric refinement and its geometric start and the 95\,\% interval of this difference.

\section{Runtime}
\label{sec:supp-runtime}
The radiometric refinement of the Sigma 14\,mm Siemens-star set (\timingPhotographs{} photographs of \timingWidth{}\,$\times$\,\timingHeight{}\,px, \timingSamples{} samples each, \timingObservations{} observations per iteration; \timingIterations{} LM iterations over \timingParametersTotal{} free parameters: \timingParametersCamera{} of the camera, \timingParametersPoses{} of the poses, \timingParametersTarget{} of the target and \timingParametersLighting{} of the lighting) takes \timingMeanS{}\,$\pm$\,\timingStdS{}\,s (mean $\pm$ std over \timingRuns{} runs, \timingPerIterationS{}\,s per iteration, writing the calibrated camera included) on one NVIDIA \timingGpu{} with \timingThreads{} CPU threads, from its loaded start; the decoding and the geometric start are excluded. \Cref{tab:supp-runtime} lists the size of the problem.
\begin{table}[h!]
	\centering
	\renewcommand{\thetable}{A6}\renewcommand{\theHtable}{A6}%
	\resizebox{\linewidth}{!}{\begin{minipage}{1.07\linewidth}\footnotesize
\setlength{\tabcolsep}{1.5pt}
\begin{tabular*}{\linewidth}{@{\extracolsep{\fill}}lr@{}}
\toprule
\multicolumn{2}{@{}l}{\underline{\textit{Sigma 14\,mm, Siemens stars, central spline}}} \\
Photographs & 132 at $4256\times2848$\,px \\
Samples per photograph & 16,384 \\
Observations per iteration & 2,162,688 \\
Free parameters (camera / poses / target / lighting) & 1,443 / 792 / 29 / 9 \\
LM iterations & 25 \\
Time per iteration [s] & 26.1 \\
Refinement time [s] (mean $\pm$ std, min--max) & 703.5 $\pm$ 3.7 (700.3--710.5) \\
Runs & 8 \\
GPU, CPU threads & NVIDIA B200, 8 \\
\bottomrule
\end{tabular*}
\end{minipage}}
	\caption{Runtime of one radiometric calibration: the refinement of the central spline of the Sigma 14\,mm lens on the Siemens-star set, from its loaded geometric start to the written calibrated camera, over \timingRuns{} runs, each on its own GPU after an untimed warm-up run. The decoding of the photographs and the geometric start are not included.}
	\label{tab:supp-runtime}
\end{table}
\addtocounter{table}{-1}

\begin{table*}[t]
  \centering
  \scriptsize
\setlength{\tabcolsep}{3pt}
\begin{tabular*}{\linewidth}{@{\extracolsep{\fill}}lrrlrrl@{}}
\toprule
 & \multicolumn{3}{c}{\textbf{Median [px]}} & \multicolumn{3}{c}{\textbf{95th percentile [px]}} \\
\cmidrule(lr){2-4}\cmidrule(lr){5-7}
\textbf{Model} & Geom. & Radiom. & $\Delta$ [95\,\% interval] & Geom. & Radiom. & $\Delta$ [95\,\% interval] \\
\midrule
\multicolumn{7}{@{}l}{\underline{\textit{Tamron 20\,mm, AprilGrid}} (142 test images, 40,999 dots)} \\
Brown--Conrady & 0.119 & \textbf{0.067} & $-0.052$ [$-0.062$, $-0.043$] & 0.356 & \textbf{0.173} & $-0.183$ [$-0.208$, $-0.161$] \\
OpenCV-14 & 0.107 & \textbf{0.060} & $-0.048$ [$-0.059$, $-0.037$] & 0.373 & \textbf{0.155} & $-0.217$ [$-0.270$, $-0.180$] \\
Kannala--Brandt & 0.117 & \textbf{0.106} & $-0.011$ [$-0.020$, $-0.001$] & 0.446 & \textbf{0.285} & $-0.162$ [$-0.220$, $-0.102$] \\
Double Sphere & 0.261 & \textit{0.286} & $+0.025$ [$+0.000$, $+0.046$] & 1.70 & \textbf{1.16} & $-0.545$ [$-0.773$, $-0.294$] \\
Central spline & 0.040 & \textit{0.048} & $+0.008$ [$+0.006$, $+0.010$] & 0.131 & \textbf{0.119} & $-0.012$ [$-0.027$, $-0.002$] \\
\midrule
\multicolumn{7}{@{}l}{\underline{\textit{Tamron 20\,mm, Siemens stars}} (112 test images, 32,170 dots)} \\
Brown--Conrady & 0.288 & \textbf{0.135} & $-0.154$ [$-0.179$, $-0.132$] & 0.855 & \textbf{0.321} & $-0.534$ [$-0.589$, $-0.480$] \\
OpenCV-14 & 0.308 & \textbf{0.131} & $-0.177$ [$-0.203$, $-0.154$] & 0.969 & \textbf{0.337} & $-0.633$ [$-0.704$, $-0.566$] \\
Kannala--Brandt & 0.183 & 0.176 & $-0.007$ [$-0.022$, $+0.005$] & 0.730 & \textbf{0.515} & $-0.215$ [$-0.264$, $-0.164$] \\
Double Sphere & 0.219 & \textit{0.268} & $+0.049$ [$+0.036$, $+0.062$] & 0.717 & \textbf{0.625} & $-0.092$ [$-0.141$, $-0.040$] \\
Central spline & 0.161 & \textbf{0.136} & $-0.026$ [$-0.034$, $-0.017$] & 0.436 & \textbf{0.331} & $-0.105$ [$-0.128$, $-0.084$] \\
\midrule
\multicolumn{7}{@{}l}{\underline{\textit{Sigma 14\,mm, AprilGrid}} (141 test images, 40,744 dots)} \\
Brown--Conrady & 0.499 & \textbf{0.377} & $-0.122$ [$-0.144$, $-0.099$] & 2.07 & \textbf{1.78} & $-0.289$ [$-0.360$, $-0.211$] \\
OpenCV-14 & 0.476 & \textbf{0.359} & $-0.117$ [$-0.146$, $-0.092$] & 2.16 & \textbf{1.74} & $-0.419$ [$-0.543$, $-0.290$] \\
Kannala--Brandt & 0.456 & \textbf{0.351} & $-0.105$ [$-0.125$, $-0.084$] & 1.99 & \textbf{1.79} & $-0.202$ [$-0.308$, $-0.122$] \\
Double Sphere & 1.42 & 1.34 & $-0.081$ [$-0.170$, $+0.000$] & 5.13 & \textbf{4.42} & $-0.707$ [$-1.171$, $-0.334$] \\
Central spline & 0.479 & \textbf{0.331} & $-0.148$ [$-0.189$, $-0.114$] & 2.45 & \textbf{1.76} & $-0.691$ [$-0.895$, $-0.539$] \\
\midrule
\multicolumn{7}{@{}l}{\underline{\textit{Sigma 14\,mm, Siemens stars}} (141 test images, 40,744 dots)} \\
Brown--Conrady & 0.687 & \textbf{0.359} & $-0.328$ [$-0.375$, $-0.284$] & 2.73 & \textbf{1.72} & $-1.009$ [$-1.253$, $-0.790$] \\
OpenCV-14 & 0.600 & \textbf{0.384} & $-0.217$ [$-0.270$, $-0.170$] & 2.87 & \textbf{1.93} & $-0.942$ [$-1.200$, $-0.703$] \\
Kannala--Brandt & 0.567 & \textbf{0.329} & $-0.238$ [$-0.277$, $-0.199$] & 2.37 & \textbf{1.75} & $-0.620$ [$-0.813$, $-0.392$] \\
Double Sphere & 0.734 & \textbf{0.461} & $-0.273$ [$-0.315$, $-0.232$] & 2.42 & \textbf{1.94} & $-0.484$ [$-0.708$, $-0.323$] \\
Central spline & 0.504 & \textbf{0.366} & $-0.137$ [$-0.174$, $-0.104$] & 2.35 & \textbf{1.97} & $-0.375$ [$-0.560$, $-0.229$] \\
\midrule
\multicolumn{7}{@{}l}{\underline{\textit{Sigma 8\,mm fisheye, AprilGrid}} (153 test images, 40,928 dots)} \\
OpenCV-14 & 1.98 & \textbf{1.70} & $-0.285$ [$-0.406$, $-0.177$] & 9.15 & \textbf{5.68} & $-3.472$ [$-3.825$, $-3.029$] \\
Kannala--Brandt & 2.00 & \textbf{1.52} & $-0.479$ [$-0.560$, $-0.395$] & 6.80 & \textbf{5.43} & $-1.371$ [$-1.564$, $-1.207$] \\
Double Sphere & 2.09 & \textbf{1.49} & $-0.600$ [$-0.671$, $-0.538$] & 7.41 & \textbf{5.45} & $-1.965$ [$-2.126$, $-1.842$] \\
Central spline & 1.99 & \textbf{1.45} & $-0.543$ [$-0.654$, $-0.433$] & 7.58 & \textbf{5.51} & $-2.076$ [$-2.491$, $-1.702$] \\
\midrule
\multicolumn{7}{@{}l}{\underline{\textit{Sigma 8\,mm fisheye, Siemens stars}} (82 test images, 22,973 dots)} \\
OpenCV-14 & 0.123 & \textbf{0.079} & $-0.044$ [$-0.065$, $-0.025$] & 0.818 & \textbf{0.440} & $-0.378$ [$-0.584$, $-0.224$] \\
Kannala--Brandt & 0.113 & \textbf{0.063} & $-0.050$ [$-0.066$, $-0.037$] & 0.590 & \textbf{0.254} & $-0.336$ [$-0.437$, $-0.237$] \\
Double Sphere & 0.109 & \textbf{0.067} & $-0.042$ [$-0.059$, $-0.029$] & 0.591 & \textbf{0.289} & $-0.302$ [$-0.391$, $-0.238$] \\
Central spline & 0.121 & \textbf{0.100} & $-0.020$ [$-0.039$, $-0.002$] & 0.996 & \textbf{0.538} & $-0.458$ [$-0.717$, $-0.309$] \\
\bottomrule
\end{tabular*}

  \caption{Median and $95$th percentile of the reprojection error on the held-out target at the geometric start and after the radiometric refinement, with their difference $\Delta$ and its 95\,\% bootstrap interval. Bold: significantly lower than the start; italic: significantly higher.}
  \label{tab:supp-refinement-a}
\end{table*}
\begin{table*}[t]
  \centering
  \scriptsize
\setlength{\tabcolsep}{3pt}
\begin{tabular*}{\linewidth}{@{\extracolsep{\fill}}lrrlrrl@{}}
\toprule
 & \multicolumn{3}{c}{\textbf{Mean [px]}} & \multicolumn{3}{c}{\textbf{RMS [px]}} \\
\cmidrule(lr){2-4}\cmidrule(lr){5-7}
\textbf{Model} & Geom. & Radiom. & $\Delta$ [95\,\% interval] & Geom. & Radiom. & $\Delta$ [95\,\% interval] \\
\midrule
\multicolumn{7}{@{}l}{\underline{\textit{Tamron 20\,mm, AprilGrid}} (142 test images, 40,999 dots)} \\
Brown--Conrady & 0.147 & \textbf{0.077} & $-0.069$ [$-0.080$, $-0.059$] & 0.185 & \textbf{0.092} & $-0.093$ [$-0.110$, $-0.081$] \\
OpenCV-14 & 0.144 & \textbf{0.070} & $-0.075$ [$-0.090$, $-0.063$] & 0.201 & \textbf{0.083} & $-0.118$ [$-0.145$, $-0.098$] \\
Kannala--Brandt & 0.159 & \textbf{0.124} & $-0.034$ [$-0.052$, $-0.020$] & 0.214 & \textbf{0.149} & $-0.065$ [$-0.092$, $-0.043$] \\
Double Sphere & 0.473 & \textbf{0.429} & $-0.044$ [$-0.085$, $-0.007$] & 0.875 & \textbf{0.696} & $-0.179$ [$-0.228$, $-0.128$] \\
Central spline & 0.051 & \textit{0.055} & $+0.003$ [$+0.001$, $+0.006$] & 0.067 & 0.065 & $-0.002$ [$-0.007$, $+0.001$] \\
\midrule
\multicolumn{7}{@{}l}{\underline{\textit{Tamron 20\,mm, Siemens stars}} (112 test images, 32,170 dots)} \\
Brown--Conrady & 0.350 & \textbf{0.151} & $-0.199$ [$-0.225$, $-0.177$] & 0.430 & \textbf{0.176} & $-0.253$ [$-0.282$, $-0.228$] \\
OpenCV-14 & 0.386 & \textbf{0.152} & $-0.234$ [$-0.261$, $-0.209$] & 0.483 & \textbf{0.181} & $-0.302$ [$-0.333$, $-0.273$] \\
Kannala--Brandt & 0.257 & \textbf{0.213} & $-0.044$ [$-0.061$, $-0.028$] & 0.342 & \textbf{0.261} & $-0.081$ [$-0.103$, $-0.060$] \\
Double Sphere & 0.279 & 0.296 & $+0.017$ [$-0.001$, $+0.033$] & 0.358 & 0.344 & $-0.013$ [$-0.036$, $+0.008$] \\
Central spline & 0.190 & \textbf{0.153} & $-0.037$ [$-0.045$, $-0.029$] & 0.230 & \textbf{0.180} & $-0.050$ [$-0.060$, $-0.041$] \\
\midrule
\multicolumn{7}{@{}l}{\underline{\textit{Sigma 14\,mm, AprilGrid}} (141 test images, 40,744 dots)} \\
Brown--Conrady & 0.716 & \textbf{0.568} & $-0.148$ [$-0.168$, $-0.128$] & 0.997 & \textbf{0.825} & $-0.172$ [$-0.201$, $-0.146$] \\
OpenCV-14 & 0.712 & \textbf{0.544} & $-0.168$ [$-0.198$, $-0.142$] & 1.02 & \textbf{0.791} & $-0.233$ [$-0.284$, $-0.191$] \\
Kannala--Brandt & 0.671 & \textbf{0.542} & $-0.128$ [$-0.147$, $-0.110$] & 0.956 & \textbf{0.802} & $-0.154$ [$-0.185$, $-0.126$] \\
Double Sphere & 1.95 & \textbf{1.80} & $-0.145$ [$-0.237$, $-0.061$] & 2.88 & \textbf{2.59} & $-0.298$ [$-0.387$, $-0.210$] \\
Central spline & 0.770 & \textbf{0.530} & $-0.240$ [$-0.288$, $-0.199$] & 1.22 & \textbf{0.797} & $-0.428$ [$-0.525$, $-0.343$] \\
\midrule
\multicolumn{7}{@{}l}{\underline{\textit{Sigma 14\,mm, Siemens stars}} (141 test images, 40,744 dots)} \\
Brown--Conrady & 0.968 & \textbf{0.546} & $-0.422$ [$-0.485$, $-0.367$] & 1.42 & \textbf{0.796} & $-0.625$ [$-0.761$, $-0.512$] \\
OpenCV-14 & 0.942 & \textbf{0.598} & $-0.344$ [$-0.405$, $-0.288$] & 1.50 & \textbf{0.884} & $-0.611$ [$-0.736$, $-0.503$] \\
Kannala--Brandt & 0.825 & \textbf{0.515} & $-0.310$ [$-0.360$, $-0.263$] & 1.23 & \textbf{0.767} & $-0.459$ [$-0.570$, $-0.361$] \\
Double Sphere & 0.926 & \textbf{0.650} & $-0.276$ [$-0.324$, $-0.235$] & 1.17 & \textbf{0.905} & $-0.262$ [$-0.333$, $-0.206$] \\
Central spline & 0.780 & \textbf{0.594} & $-0.186$ [$-0.235$, $-0.148$] & 1.20 & \textbf{0.901} & $-0.301$ [$-0.403$, $-0.226$] \\
\midrule
\multicolumn{7}{@{}l}{\underline{\textit{Sigma 8\,mm fisheye, AprilGrid}} (153 test images, 40,928 dots)} \\
OpenCV-14 & 2.98 & \textbf{2.18} & $-0.801$ [$-0.963$, $-0.651$] & 4.14 & \textbf{2.80} & $-1.336$ [$-1.536$, $-1.144$] \\
Kannala--Brandt & 2.59 & \textbf{2.01} & $-0.578$ [$-0.649$, $-0.506$] & 3.33 & \textbf{2.65} & $-0.682$ [$-0.750$, $-0.614$] \\
Double Sphere & 2.75 & \textbf{1.99} & $-0.763$ [$-0.830$, $-0.698$] & 3.59 & \textbf{2.64} & $-0.949$ [$-1.023$, $-0.878$] \\
Central spline & 2.70 & \textbf{1.96} & $-0.746$ [$-0.868$, $-0.638$] & 3.69 & \textbf{2.63} & $-1.059$ [$-1.306$, $-0.897$] \\
\midrule
\multicolumn{7}{@{}l}{\underline{\textit{Sigma 8\,mm fisheye, Siemens stars}} (82 test images, 22,973 dots)} \\
OpenCV-14 & 0.280 & \textbf{0.164} & $-0.115$ [$-0.188$, $-0.077$] & 2.62 & \textbf{1.09} & $-1.533$ [$-2.953$, $-0.132$] \\
Kannala--Brandt & 0.182 & \textbf{0.089} & $-0.093$ [$-0.122$, $-0.070$] & 0.265 & \textbf{0.123} & $-0.141$ [$-0.181$, $-0.107$] \\
Double Sphere & 0.181 & \textbf{0.098} & $-0.083$ [$-0.109$, $-0.064$] & 0.265 & \textbf{0.151} & $-0.114$ [$-0.148$, $-0.086$] \\
Central spline & 0.256 & \textbf{0.182} & $-0.074$ [$-0.113$, $-0.042$] & 0.472 & 0.502 & $+0.029$ [$-0.153$, $+0.342$] \\
\bottomrule
\end{tabular*}

  \caption{Mean and RMS of the reprojection error, as \cref{tab:supp-refinement-a}.}
  \label{tab:supp-refinement-b}
\end{table*}

\section{Vignetting Correction of Panoramas}
\label{sec:supp-panoramas}
Following Kim and Pollefeys \cite{Kim-PAMI-2008}, we use that a scene
point seen in two images of a camera that rotates about its center of
projection, under unchanged lighting and exposure, has the same
radiance in both images, so that its two measurements differ only by
the vignetting.

\begin{figure*}[t]
        \centering
        \providecommand{\rcfigdir}{fig}%
\begingroup
\ifdefined\rcpanelwd\else\newlength\rcpanelwd\fi
\footnotesize\setlength{\tabcolsep}{0pt}\renewcommand{\arraystretch}{1}%
\ifdim\linewidth<4.5in\relax
\setlength{\rcpanelwd}{\dimexpr(\linewidth-4pt)/2\relax}\def\rcpanelwidth{\rcpanelwd}%
\begin{tabular}{@{}c@{\hspace{4pt}}c@{}}
  {\scriptsize no vignetting correction} & {\scriptsize our vignetting correction} \\\noalign{\vskip 2pt}
  \multicolumn{2}{@{}c@{}}{{\scriptsize Sigma 14\,mm lens}} \\\noalign{\vskip 2pt}
  \providecommand{\rcfigdir}{fig}%
\providecommand{\rcpanelwidth}{\linewidth}%
\begin{tikzpicture}[baseline=(current axis.center)]
\begin{axis}[
  scale only axis, width=\rcpanelwidth, height=0.27214*\rcpanelwidth,
  grid=none, log ticks with fixed point, tick label style={font=\scriptsize}, label style={font=\small}, legend cell align=left, legend style={font=\scriptsize, draw=none, fill=none},
  xmin=0, xmax=1400, ymin=0, ymax=381, enlargelimits=false,
  xtick=\empty, ytick=\empty, axis on top, axis line style={draw=none}]
\addplot graphics[xmin=0, xmax=1400, ymin=0, ymax=381] {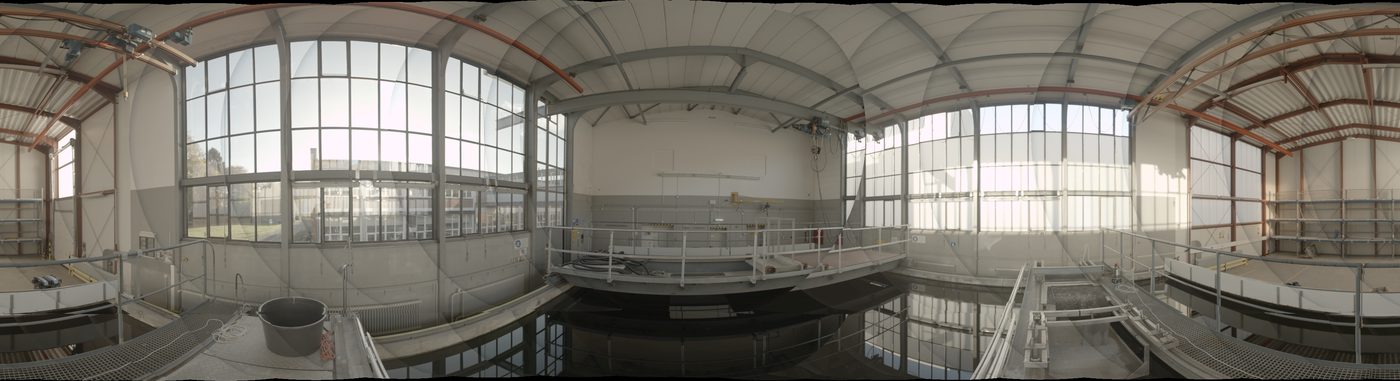};
\end{axis}
\end{tikzpicture}%
 & \providecommand{\rcfigdir}{fig}%
\providecommand{\rcpanelwidth}{\linewidth}%
\begin{tikzpicture}[baseline=(current axis.center)]
\begin{axis}[
  scale only axis, width=\rcpanelwidth, height=0.27214*\rcpanelwidth,
  grid=none, log ticks with fixed point, tick label style={font=\scriptsize}, label style={font=\small}, legend cell align=left, legend style={font=\scriptsize, draw=none, fill=none},
  xmin=0, xmax=1400, ymin=0, ymax=381, enlargelimits=false,
  xtick=\empty, ytick=\empty, axis on top, axis line style={draw=none}]
\addplot graphics[xmin=0, xmax=1400, ymin=0, ymax=381] {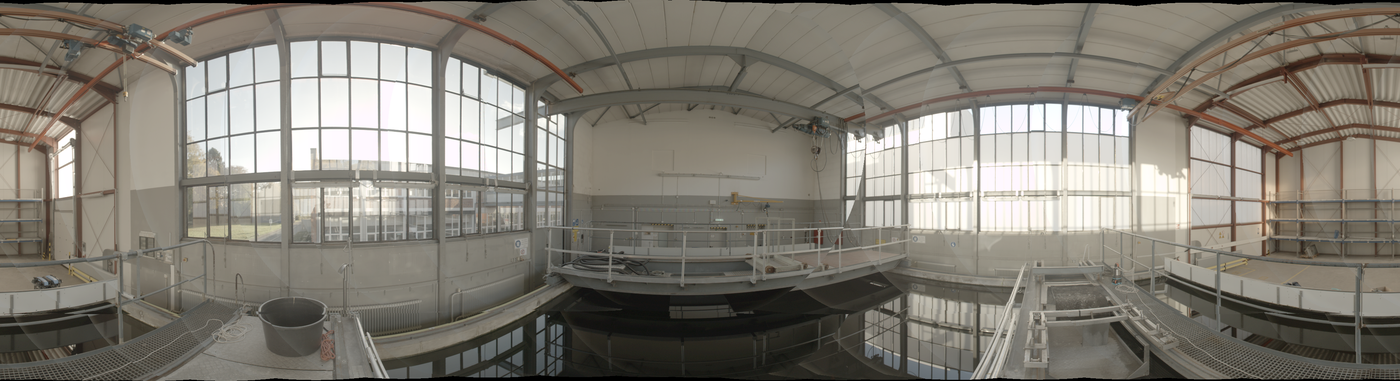};
\end{axis}
\end{tikzpicture}%
 \\\noalign{\vskip 3pt}
  \multicolumn{2}{@{}c@{}}{{\scriptsize fisheye lens}} \\\noalign{\vskip 2pt}
  \providecommand{\rcfigdir}{fig}%
\providecommand{\rcpanelwidth}{\linewidth}%
\begin{tikzpicture}[baseline=(current axis.center)]
\begin{axis}[
  scale only axis, width=\rcpanelwidth, height=0.43214*\rcpanelwidth,
  grid=none, log ticks with fixed point, tick label style={font=\scriptsize}, label style={font=\small}, legend cell align=left, legend style={font=\scriptsize, draw=none, fill=none},
  xmin=0, xmax=1400, ymin=0, ymax=605, enlargelimits=false,
  xtick=\empty, ytick=\empty, axis on top, axis line style={draw=none}]
\addplot graphics[xmin=0, xmax=1400, ymin=0, ymax=605] {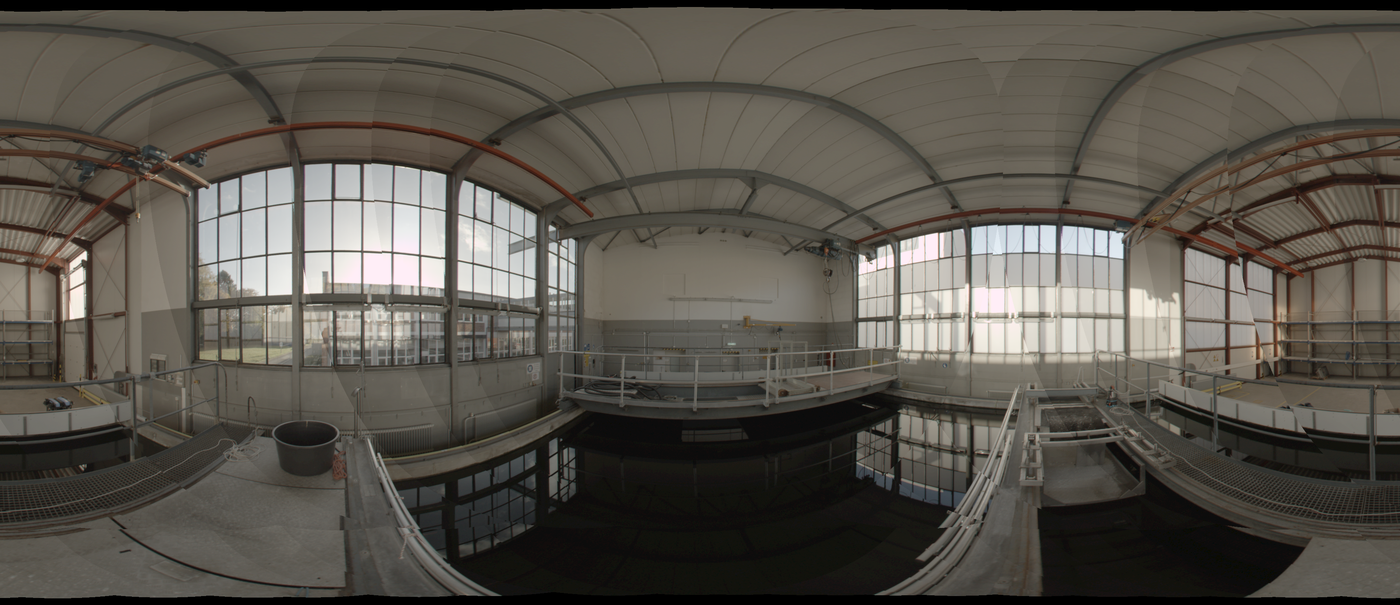};
\end{axis}
\end{tikzpicture}%
 & \providecommand{\rcfigdir}{fig}%
\providecommand{\rcpanelwidth}{\linewidth}%
\begin{tikzpicture}[baseline=(current axis.center)]
\begin{axis}[
  scale only axis, width=\rcpanelwidth, height=0.43214*\rcpanelwidth,
  grid=none, log ticks with fixed point, tick label style={font=\scriptsize}, label style={font=\small}, legend cell align=left, legend style={font=\scriptsize, draw=none, fill=none},
  xmin=0, xmax=1400, ymin=0, ymax=605, enlargelimits=false,
  xtick=\empty, ytick=\empty, axis on top, axis line style={draw=none}]
\addplot graphics[xmin=0, xmax=1400, ymin=0, ymax=605] {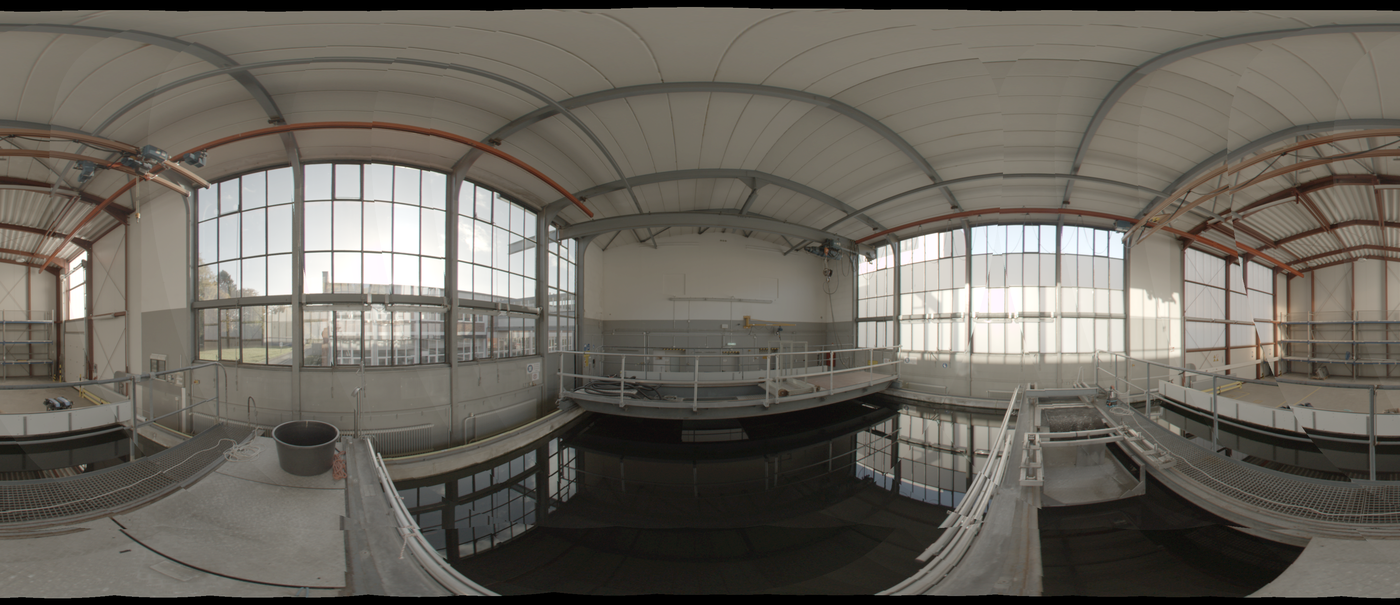};
\end{axis}
\end{tikzpicture}%
 \\\noalign{\vskip 3pt}
  \multicolumn{2}{@{}c@{}}{{\scriptsize Tamron 20\,mm lens}} \\\noalign{\vskip 2pt}
  \providecommand{\rcfigdir}{fig}%
\providecommand{\rcpanelwidth}{\linewidth}%
\begin{tikzpicture}[baseline=(current axis.center)]
\begin{axis}[
  scale only axis, width=\rcpanelwidth, height=0.20143*\rcpanelwidth,
  grid=none, log ticks with fixed point, tick label style={font=\scriptsize}, label style={font=\small}, legend cell align=left, legend style={font=\scriptsize, draw=none, fill=none},
  xmin=0, xmax=1400, ymin=0, ymax=282, enlargelimits=false,
  xtick=\empty, ytick=\empty, axis on top, axis line style={draw=none}]
\addplot graphics[xmin=0, xmax=1400, ymin=0, ymax=282] {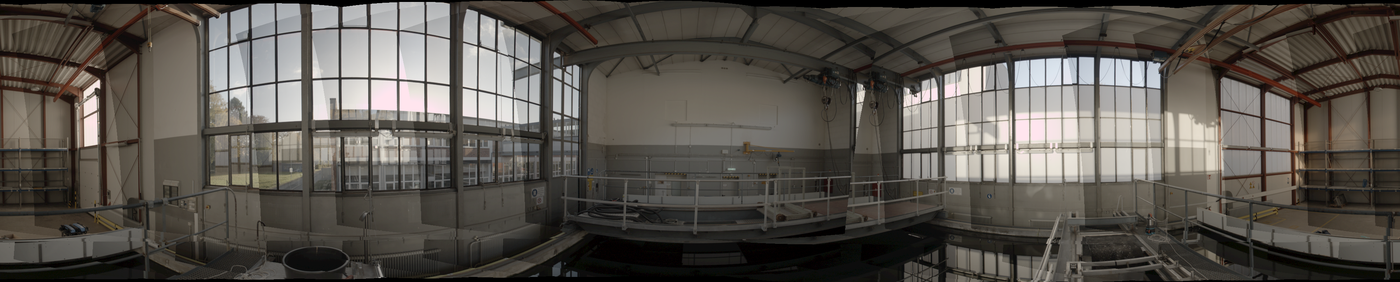};
\end{axis}
\end{tikzpicture}%
 & \providecommand{\rcfigdir}{fig}%
\providecommand{\rcpanelwidth}{\linewidth}%
\begin{tikzpicture}[baseline=(current axis.center)]
\begin{axis}[
  scale only axis, width=\rcpanelwidth, height=0.20143*\rcpanelwidth,
  grid=none, log ticks with fixed point, tick label style={font=\scriptsize}, label style={font=\small}, legend cell align=left, legend style={font=\scriptsize, draw=none, fill=none},
  xmin=0, xmax=1400, ymin=0, ymax=282, enlargelimits=false,
  xtick=\empty, ytick=\empty, axis on top, axis line style={draw=none}]
\addplot graphics[xmin=0, xmax=1400, ymin=0, ymax=282] {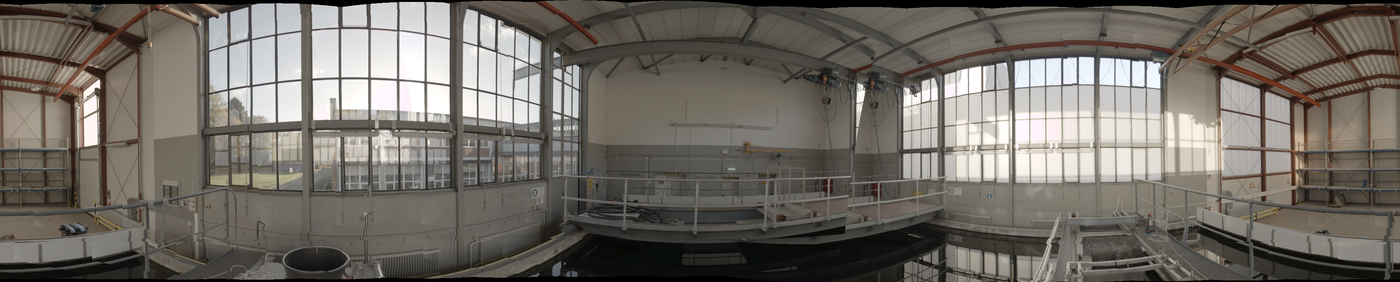};
\end{axis}
\end{tikzpicture}%
 \\
\end{tabular}%
\else
\setlength{\rcpanelwd}{\dimexpr(\linewidth-17pt-4pt)/2\relax}\def\rcpanelwidth{\rcpanelwd}%
\begin{tabular}{@{}l@{}c@{\hspace{4pt}}c@{}}
  & {\scriptsize no vignetting correction} & {\scriptsize our vignetting correction} \\\noalign{\vskip 2pt}
  \makebox[17pt][l]{\tikz[baseline=(n.center)]\node[rotate=90,inner sep=0pt,align=center,font=\scriptsize](n){Sigma\\14\,mm lens};} & \providecommand{\rcfigdir}{fig}%
\providecommand{\rcpanelwidth}{\linewidth}%
\begin{tikzpicture}[baseline=(current axis.center)]
\begin{axis}[
  scale only axis, width=\rcpanelwidth, height=0.27214*\rcpanelwidth,
  grid=none, log ticks with fixed point, tick label style={font=\scriptsize}, label style={font=\small}, legend cell align=left, legend style={font=\scriptsize, draw=none, fill=none},
  xmin=0, xmax=1400, ymin=0, ymax=381, enlargelimits=false,
  xtick=\empty, ytick=\empty, axis on top, axis line style={draw=none}]
\addplot graphics[xmin=0, xmax=1400, ymin=0, ymax=381] {\rcfigdir/panoramas-james/sigma14-no-correction.png};
\end{axis}
\end{tikzpicture}%
 & \providecommand{\rcfigdir}{fig}%
\providecommand{\rcpanelwidth}{\linewidth}%
\begin{tikzpicture}[baseline=(current axis.center)]
\begin{axis}[
  scale only axis, width=\rcpanelwidth, height=0.27214*\rcpanelwidth,
  grid=none, log ticks with fixed point, tick label style={font=\scriptsize}, label style={font=\small}, legend cell align=left, legend style={font=\scriptsize, draw=none, fill=none},
  xmin=0, xmax=1400, ymin=0, ymax=381, enlargelimits=false,
  xtick=\empty, ytick=\empty, axis on top, axis line style={draw=none}]
\addplot graphics[xmin=0, xmax=1400, ymin=0, ymax=381] {\rcfigdir/panoramas-james/sigma14-our-correction.png};
\end{axis}
\end{tikzpicture}%
 \\\noalign{\vskip 3pt}
  \makebox[17pt][l]{\tikz[baseline=(n.center)]\node[rotate=90,inner sep=0pt,align=center,font=\scriptsize](n){fisheye\\lens};} & \providecommand{\rcfigdir}{fig}%
\providecommand{\rcpanelwidth}{\linewidth}%
\begin{tikzpicture}[baseline=(current axis.center)]
\begin{axis}[
  scale only axis, width=\rcpanelwidth, height=0.43214*\rcpanelwidth,
  grid=none, log ticks with fixed point, tick label style={font=\scriptsize}, label style={font=\small}, legend cell align=left, legend style={font=\scriptsize, draw=none, fill=none},
  xmin=0, xmax=1400, ymin=0, ymax=605, enlargelimits=false,
  xtick=\empty, ytick=\empty, axis on top, axis line style={draw=none}]
\addplot graphics[xmin=0, xmax=1400, ymin=0, ymax=605] {\rcfigdir/panoramas-james/fisheye-no-correction.png};
\end{axis}
\end{tikzpicture}%
 & \providecommand{\rcfigdir}{fig}%
\providecommand{\rcpanelwidth}{\linewidth}%
\begin{tikzpicture}[baseline=(current axis.center)]
\begin{axis}[
  scale only axis, width=\rcpanelwidth, height=0.43214*\rcpanelwidth,
  grid=none, log ticks with fixed point, tick label style={font=\scriptsize}, label style={font=\small}, legend cell align=left, legend style={font=\scriptsize, draw=none, fill=none},
  xmin=0, xmax=1400, ymin=0, ymax=605, enlargelimits=false,
  xtick=\empty, ytick=\empty, axis on top, axis line style={draw=none}]
\addplot graphics[xmin=0, xmax=1400, ymin=0, ymax=605] {\rcfigdir/panoramas-james/fisheye-our-correction.png};
\end{axis}
\end{tikzpicture}%
 \\\noalign{\vskip 3pt}
  \makebox[17pt][l]{\tikz[baseline=(n.center)]\node[rotate=90,inner sep=0pt,align=center,font=\scriptsize](n){Tamron\\20\,mm lens};} & \providecommand{\rcfigdir}{fig}%
\providecommand{\rcpanelwidth}{\linewidth}%
\begin{tikzpicture}[baseline=(current axis.center)]
\begin{axis}[
  scale only axis, width=\rcpanelwidth, height=0.20143*\rcpanelwidth,
  grid=none, log ticks with fixed point, tick label style={font=\scriptsize}, label style={font=\small}, legend cell align=left, legend style={font=\scriptsize, draw=none, fill=none},
  xmin=0, xmax=1400, ymin=0, ymax=282, enlargelimits=false,
  xtick=\empty, ytick=\empty, axis on top, axis line style={draw=none}]
\addplot graphics[xmin=0, xmax=1400, ymin=0, ymax=282] {\rcfigdir/panoramas-james/tamron20-no-correction.png};
\end{axis}
\end{tikzpicture}%
 & \providecommand{\rcfigdir}{fig}%
\providecommand{\rcpanelwidth}{\linewidth}%
\begin{tikzpicture}[baseline=(current axis.center)]
\begin{axis}[
  scale only axis, width=\rcpanelwidth, height=0.20143*\rcpanelwidth,
  grid=none, log ticks with fixed point, tick label style={font=\scriptsize}, label style={font=\small}, legend cell align=left, legend style={font=\scriptsize, draw=none, fill=none},
  xmin=0, xmax=1400, ymin=0, ymax=282, enlargelimits=false,
  xtick=\empty, ytick=\empty, axis on top, axis line style={draw=none}]
\addplot graphics[xmin=0, xmax=1400, ymin=0, ymax=282] {\rcfigdir/panoramas-james/tamron20-our-correction.png};
\end{axis}
\end{tikzpicture}%
 \\
\end{tabular}%
\fi
\endgroup

        \caption{\protect\input{fig/panoramas-james-caption}}
        \label{fig:panoramas-james}
\end{figure*}

Let $\vbu$ and $\vbu'$ be two pixels in two different images,
corresponding to the same scene point $\vbx$, and let $\vbu$ and
$\vbu'$ denote the pixel centers as well. Let
\begin{equation*}
    \bar{L}(\vbp) \coloneqq \frac{1}{\pi}\int_{\mathcal{D}}L\bigl(\vbx(\vbp,\vba),-\vbpi^{-1}(\vbp,\vba)\bigr)\odif{\vbxi}
\end{equation*}
be the radiance that arrives at the sensor point $\vbp$, averaged over
the aperture, so that \cref{eq:rad-flux-with-vignetting} reads
$\Phi(\vbu)=\int_{\mathcal{I}_{\vbu}}V(\vbp)\bar{L}(\vbp)\odif{\vbp}$.
We assume that the radiance is constant over the pixel area and the
aperture, so that both pixels collect the same radiance,
\begin{equation*}
    \int_{\mathcal{I}_{\vbu}}\bar{L}(\vbp)\odif{\vbp}
    \approx \int_{\mathcal{I}_{\vbu'}}\bar{L}(\vbp')\odif{\vbp'},
\end{equation*}
where each side is evaluated in its own image. We also assume that
the vignetting is constant over a pixel area, \ie
\begin{equation*}
    V(\vbp)\approx V(\vbu), \qquad V(\vbp')\approx V(\vbu')
\end{equation*}
for all $\vbp\in\mathcal{I}_{\vbu}$ and $\vbp'\in\mathcal{I}_{\vbu'}$.
With the measured values $\phi(\vbu)$ and $\phi(\vbu')$ of the two
pixels, it follows that
\begin{align*}
    \frac{\phi(\vbu)}{V(\vbu)}
    &\approx \frac{\Phi(\vbu)}{V(\vbu)}\\
    &= \frac{1}{V(\vbu)}\int_{\mathcal{I}_{\vbu}}V(\vbp)\bar{L}(\vbp)\odif{\vbp}\\
    &\approx \int_{\mathcal{I}_{\vbu}}\bar{L}(\vbp)\odif{\vbp}\\
    &\approx \int_{\mathcal{I}_{\vbu'}}\bar{L}(\vbp')\odif{\vbp'}\\
    &\approx \frac{1}{V(\vbu')}\int_{\mathcal{I}_{\vbu'}}V(\vbp')\bar{L}(\vbp')\odif{\vbp'}\\
    &= \frac{\Phi(\vbu')}{V(\vbu')}
    \approx \frac{\phi(\vbu')}{V(\vbu')}.
\end{align*}
The panoramas of \cref{fig:panoramas-james} test this relation. Every
image is divided by the learned vignetting, and the images are
composed without blending and without exposure compensation per
image, so that each pixel of the panorama shows the value of a single
image. Where two images meet, neighboring pixels of the panorama
therefore show $\phi(\vbu)/V(\vbu)$ from one image and
$\phi(\vbu')/V(\vbu')$ from the other for nearly the same scene
point. Without the correction, these values differ by the factor
$V(\vbu)/V(\vbu')$, which is visible as a seam. With the correct
vignetting they agree, and the seam disappears. For display, all
images of a lens share one exposure, one white balance and one tone
curve, in the uncorrected and in the corrected panorama alike, so
that the two differ by the vignetting correction alone.

\end{document}